\pdfoutput=1
\documentclass[drones,review,accept,pdftex,moreauthors]{Definitions/mdpi}

\firstpage{1} 

\pubvolume{10}
\issuenum{6}
\articlenumber{412}
\pubyear{2026}
\copyrightyear{2026}
\externaleditor{Peihu Duan}
\datereceived{9 April 2026} 
\daterevised{13 May 2026}
\dateaccepted{21 May 2026} 
\datepublished{26 May 2026}

\usepackage{amsfonts}
\usepackage{algorithm, algorithmic}
\usepackage{bm}
\usepackage{pgfplots}
\usetikzlibrary{shapes.geometric, arrows.meta, positioning, fit, calc, backgrounds}
\pgfplotsset{compat=1.16}

\definecolor{cBlue}{HTML}{1971C2}
\definecolor{cBlueFill}{HTML}{D0EBFF}
\definecolor{cTeal}{HTML}{0C8599}
\definecolor{cTealFill}{HTML}{C3FAE8}
\definecolor{cTealPale}{HTML}{E6FCF5}
\definecolor{cGreen}{HTML}{2F9E44}
\definecolor{cGreenFill}{HTML}{D3F9D8}
\definecolor{cOrange}{HTML}{E8590C}
\definecolor{cOrangeFill}{HTML}{FFD8A8}
\definecolor{cOrangePale}{HTML}{FFF4E6}
\definecolor{cRed}{HTML}{C92A2A}
\definecolor{cRedFill}{HTML}{FFE3E3}
\definecolor{cViolet}{HTML}{7048E8}
\definecolor{cVioletFill}{HTML}{E5DBFF}
\definecolor{cVioletPale}{HTML}{F3F0FF}
\definecolor{cPink}{HTML}{C2255C}
\definecolor{cPinkFill}{HTML}{FCC2D7}
\definecolor{cPinkPale}{HTML}{FFF0F6}
\definecolor{cGray}{HTML}{495057}
\definecolor{cGrayFill}{HTML}{F1F3F5}

\newcommand{\ie}{{i.e.}}
\newcommand{\eg}{{e.g.}}
\newcommand{\etal}{{et al.}}

\newif\ifshowreviewnotes
\showreviewnotesfalse         % <- set to false for final submission
\usepackage{CJKutf8}         % Korean text support
\usepackage{tcolorbox}
\tcbuselibrary{breakable,skins}
\definecolor{rnR1}{HTML}{D63384}   % R1 pink
\definecolor{rnR2}{HTML}{0D6EFD}   % R2 blue
\definecolor{rnR3}{HTML}{198754}   % R3 green
\definecolor{rnR4}{HTML}{FD7E14}   % R4 orange
\newcommand{\rncolor}[1]{%
  \ifnum\pdfstrcmp{#1}{R1}=0 rnR1\else
  \ifnum\pdfstrcmp{#1}{R2}=0 rnR2\else
  \ifnum\pdfstrcmp{#1}{R4}=0 rnR4\else
                              rnR3\fi\fi\fi
}

\Title{Vision--Language--Action (VLA) Models for Unmanned Aerial Robotics and Bimanual Manipulation: A Review}

\Author{Inkyu Sa %
 $^{1,}$*\orcidA{}, Chanoh Park $^{2}$, Hea-Min Lee $^{3}$\orcidC{}, Donghee Noh $^{3}$\orcidD{} and Ho Seok Ahn $^{4}$\orcidE{}}

\AuthorNames{Inkyu Sa, Chanoh Park, Hea-Min Lee, Donghee Noh, and Ho Seok Ahn}

\address{%
$^{1}$ \quad Chef Robotics, San Francisco, CA 94103, USA %
 \\
$^{2}$ \quad RovifyLab, Gyeonggi 13840, Republic of Korea;
chanoh.park@rovifylab.com\\
$^{3}$ \quad IT Application Research Center, Jeonbuk Regional Branch, Korea Electronics Technology Institute (KETI), Jeonju %
 54853, Republic of Korea; lee10849@keti.re.kr (H.-M.L.); dhee.noh@keti.re.kr (D.N.)\\
$^{4}$ \quad Department of Electrical, Computer and Software Engineering,
 University of Auckland, \mbox{Auckland 1010, New~Zealand} %
; hs.ahn@auckland.ac.nz
}

\corres{Correspondence: inkyu@chefrobotics.ai}

\abstract{Vision--Language--Action (VLA) models unify visual perception, natural-language understanding, and action generation within a single foundation model, allowing a robot to follow instructions such as ``fold the towel'' or ``fly to the red building'' directly from camera images. Because VLAs inherit world knowledge from internet-scale pre-training, they have become the dominant framework for learning-based manipulation, with bimanual coordination serving as the most demanding testbed: two arms with $7{+}$ degrees of freedom each must move in concert to fold, assemble, and reorient objects. Unmanned aerial robotics faces a structurally similar challenge: a drone must coordinate thrust, attitude, and increasingly gripper commands from visual observations under strict latency and payload constraints. This review covers 183 contributions spanning 2017--2026 and organized along seven dimensions: VLA architectures, training recipes, action representations, bimanual coordination (2022--2026), unmanned aerial vehicle (UAV) navigation and control (2017--2026), language grounding, and cross-cutting concerns including memory and world models. We show that the coordination strategies, training recipes, and action representations developed for bimanual VLAs transfer to unmanned aerial systems and identify fourteen research directions across both domains.}


\addhighlights{yes}
\renewcommand{\addhighlights}{%

\noindent\textbf{What are the main findings?}
\begin{itemize}[labelsep=2.5mm,topsep=-3pt]
  \item The survey finds that VLA research is converging on continuous, chunked action generation---especially flow-matching and hybrid designs---because they avoid the quantization bottleneck of autoregressive action tokens and the latency burden of multi-step diffusion, making them better suited to tightly coordinated bimanual control and transferable to aerial systems.
  \item It also finds that progress is driven as much by training strategy as by model architecture: cross-embodiment data diversity and co-training improve downstream generalization more reliably than raw dataset scale alone, while reinforcement learning from autonomous practice is emerging as the key mechanism for surpassing demonstration-limited performance---that is, achieving higher task success rates, faster execution, and broader generalization than what teleoperated demonstrations alone can support.
\end{itemize}\vspace{3pt}

\noindent\textbf{What are the implications of the main findings?}
\begin{itemize}[labelsep=2.5mm,topsep=-3pt]
  \item These trends suggest that the most promising path for real-world deployment is not a monolithic end-to-end model but a dual-system design that combines a slower reasoning module with a faster action module, enabling both semantic understanding and high-frequency control in manipulation and aerial robotics.
  \item Looking forward, the field is likely to expand toward production-grade embodied autonomy through end-to-end drone VLAs, aerial manipulation, memory and world-model integration, standardized bimanual benchmarks, safety certification, and continuous self-improvement pipelines that close the gap between benchmark performance and industrial reliability.
\end{itemize}
}

\begin{document}

\vspace{2em}
\noindent\textbf{Version of Record:} This is the author's accepted manuscript. The definitive Version of Record was published in \textit{Drones} \textbf{2026}, \textit{10}(6), 412. \url{https://doi.org/10.3390/drones10060412}. This work is licensed under a Creative Commons Attribution 4.0 International License (CC BY 4.0).
\vspace{2em}

% ============================================================
% 1. INTRODUCTION
% ============================================================
\section{Introduction}
\label{sec:introduction}

Vision--Language--Action %
 (VLA) models use a single foundation model to map camera images and language instructions to robot actions. A VLA processes visual and language inputs through a Vision--Language Model (VLM) pre-trained on internet-scale data then generates motor commands through a learned action head. Because the architecture makes no assumptions about the specific robot, VLAs can control manipulators, mobile robots, and drones with the same model family, enabling robots to assist in homes, factories, and disaster-response scenarios.

To date, the vast majority of VLA research has focused on manipulation, and bimanual coordination in particular. Bimanual tasks (folding laundry, assembling boxes, clearing tables) require two $7{+}$-degree-of-freedom arms to move in concert under partial observability, making them among the most challenging testbeds for VLA models. This concentration of research effort means that bimanual manipulation is where VLA architectures, training recipes, and action representations are best understood. We therefore devote the first application section of this review to a detailed analysis of VLAs for bimanual manipulation.

We then extend the analysis to unmanned aerial robotics, where the same VLA ideas are now being adopted. The connection between the two domains is not merely conceptual. Coordinating two arms and coordinating a drone fleet both require generating coupled multi-agent actions from shared observations. The action chunking methods that produce smooth bimanual trajectories also produce smooth flight paths. Drones with grippers or robotic arms face both challenges at once, stabilizing flight while manipulating objects. The training recipes (pre-training on diverse data, sim-to-real transfer, reinforcement learning from practice) are shared. Language grounding is also unified: the same VLM mechanisms that interpret ``fold the shirt neatly'' for a manipulator interpret ``fly to the red building and inspect the roof'' for a drone. Reviewing bimanual VLAs first provides the vocabulary and analytical framework that makes the aerial discussion concrete.

\textbf{Progress in manipulation VLAs.} %
 The field has moved quickly. \textbf{RT-2}~(2023)%
~\cite{brohan2023rt2} first showed that a VLM can be fine-tuned to output robot actions. \textbf{$\pi_0$} (2024)~\cite{black2024pi0} introduced flow matching, a method that learns to transform random noise into robot actions. It reached state-of-the-art bimanual performance on tasks like laundry folding and box assembly. \textbf{$\pi_{0.5}$} (2025)~\cite{black2025pi05} deployed VLAs in real homes with high success rates, and \textbf{$\pi_0^*$}~\cite{amin2025pi06} enabled VLAs to improve from their own practice via reinforcement learning. Open-source systems (\textbf{OpenVLA}~\cite{kim2024openvla}, \textbf{Octo}~\cite{octo2024}) and efficient architectures~\cite{wen2024tinyvla,pertsch2025fast} have made the technology broadly accessible.

\textbf{Emergence of unmanned aerial VLAs.} In parallel, the unmanned aerial systems community has begun adopting VLA ideas. \textbf{CognitiveDrone}~\cite{cognitivedrone2025} generates real-time flight commands from camera images and text instructions. \textbf{DroneVLA}~\cite{dronevla2026} performs language-commanded aerial grasping, while \textbf{AIR-VLA}~\cite{airvla2026} benchmarks aerial manipulation VLAs. \textbf{Flying Hand}~\cite{flyinghand2025} uses the same action chunking method developed for bimanual manipulation (ACT) on a hexarotor with a robotic arm. These systems confirm that the VLA framework transfers across embodiments.

\textbf{Gap in existing surveys.} Surveys on foundation models for robotics~\cite{firoozi2023foundation} address high-level planning but not low-level motor control. Reviews of diffusion-based imitation learning~\cite{wolf2025diffusionsurvey} focus on policy generation but do not cover VLA architectures. Surveys on multi-arm systems~\cite{abbas2023dualarm} cover classical methods, not learned policies. Aerial surveys have examined perception and detection but not end-to-end VLA-based drone control. No existing review treats bimanual manipulation and unmanned aerial robotics as two instances of the same VLA problem.

This review fills that gap by treating VLAs as a single framework applied to two embodiment families. We first review the shared VLA machinery (architectures, training recipes, action representations, language grounding) and then apply it to bimanual manipulation and unmanned aerial robotics, in turn, drawing explicit parallels throughout. The main contributions are:
\begin{itemize}
    \item A unified taxonomy of VLA models covering architectures, training, action representations, bimanual manipulation, and unmanned aerial robotics, with comparison tables spanning 30+ methods.
    \item The first cross-domain analysis connecting bimanual coordination strategies to multi-drone and aerial manipulation systems, showing how insights transfer between embodiments.
    \item Fourteen research directions identifying open challenges across both domains, from real-time control and safety certification to end-to-end drone VLAs and bridging the research-to-production gap.
\end{itemize}

This paper is structured to build from shared foundations to domain-specific applications. Sections~\ref{sec:problem}--\ref{sec:action} cover the common VLA stack: problem formulation, background, benchmarks, architectures, training, and action representations. Section~\ref{sec:bimanual} then applies this stack to bimanual manipulation, where VLAs are most mature. Section~\ref{sec:aerial} applies it to unmanned aerial robotics, drawing on the bimanual analysis to highlight what transfers and what differs. Section~\ref{sec:language} examines language grounding across both domains. Section~\ref{sec:crosscutting} addresses cross-cutting concerns (visual representations, world models, memory, safety, sim-to-real). Section~\ref{sec:discussion} synthesizes findings and identifies research directions that span both embodiment families.

% ============================================================
% 2. PROBLEM DEFINITION AND SCOPE
% ============================================================
\section{Problem Definition and Scope}
\label{sec:problem}

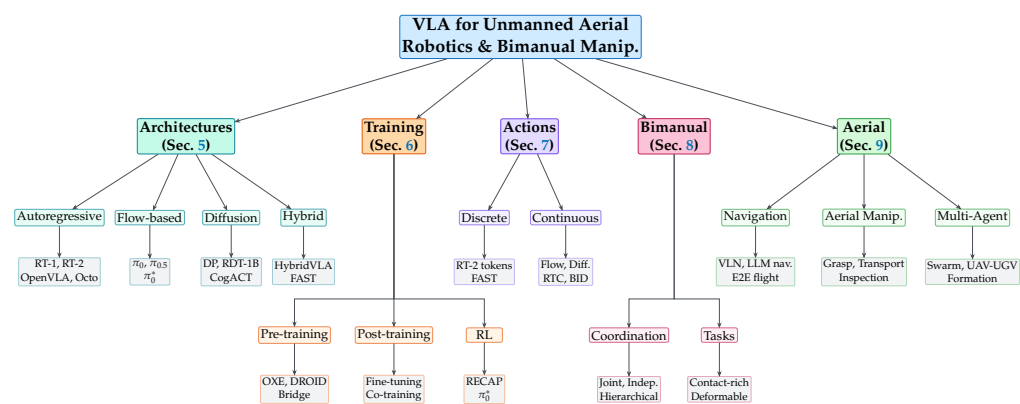
\begin{figure}[H]
%\centering
\resizebox{\textwidth}{!}{%
\begin{tikzpicture}[
    every node/.style={align=center},
    root/.style={rectangle, rounded corners=4pt, draw=cBlue, fill=cBlueFill, line width=0.8pt,
                 font=\fontsize{28}{34}\selectfont\bfseries, minimum width=55mm, minimum height=15mm},
    lvl1/.style={rectangle, rounded corners=3pt, line width=0.7pt,
                 font=\fontsize{24}{29}\selectfont\bfseries, minimum width=28mm, minimum height=9mm},
    lvl2/.style={rectangle, rounded corners=3pt, line width=0.5pt,
                 font=\fontsize{20}{24}\selectfont, minimum width=20mm, minimum height=8mm},
    lvl3/.style={rectangle, rounded corners=2pt, line width=0.4pt,
                 font=\fontsize{17}{21}\selectfont, minimum width=16mm, minimum height=6.5mm},
    arr/.style={draw=cGray, -{Stealth[length=2.5mm]}, thick},
    bus/.style={draw=cGray, thick}
]
% === Root ===
\node[root] (root) at (25, 2) {VLA for Unmanned Aerial\\Robotics \& Bimanual Manip.};

% === Level 1: five branches ===
\node[lvl1, draw=cTeal, fill=cTealFill]    (arch)   at (6.5,  -3.5) {Architectures\\(Sec.~\ref{sec:architectures})};
\node[lvl1, draw=cOrange, fill=cOrangeFill] (train)  at (18,   -3.5) {Training\\(Sec.~\ref{sec:training})};
\node[lvl1, draw=cViolet, fill=cVioletFill] (act)    at (25.5, -3.5) {Actions\\(Sec.~\ref{sec:action})};
\node[lvl1, draw=cPink, fill=cPinkFill]    (biman)  at (33.5, -3.5) {Bimanual\\(Sec.~\ref{sec:bimanual})};
\node[lvl1, draw=cGreen, fill=cGreenFill]  (aerial) at (44,   -3.5) {Aerial\\(Sec.~\ref{sec:aerial})};

% === Tier 1 L2 (y=-8): Arch, Actions, Aerial children ===
\node[lvl2, draw=cTeal!70, fill=cTealPale] (autoreg) at (-0.5,  -8) {Autoregressive};
\node[lvl2, draw=cTeal!70, fill=cTealPale] (flow)    at (4.5,   -8) {Flow-based};
\node[lvl2, draw=cTeal!70, fill=cTealPale] (diff)    at (9,     -8) {Diffusion};
\node[lvl2, draw=cTeal!70, fill=cTealPale] (hybrid)  at (13,    -8) {Hybrid};
%                                      gap x=13..23 — Training arrows route here
\node[lvl2, draw=cViolet!70, fill=cVioletPale] (discrete)   at (23,   -8) {Discrete};
\node[lvl2, draw=cViolet!70, fill=cVioletPale] (continuous) at (27.5,  -8) {Continuous};
%                                      gap x=27.5..38 — Bimanual arrows route here
\node[lvl2, draw=cGreen!70, fill=cGreenFill!30] (anav)   at (38,    -8) {Navigation};
\node[lvl2, draw=cGreen!70, fill=cGreenFill!30] (amanip) at (44,    -8) {Aerial Manip.};
\node[lvl2, draw=cGreen!70, fill=cGreenFill!30] (amulti) at (50,    -8) {Multi-Agent};

% === Tier 1 L3 (y=-11): Arch, Actions, Aerial leaves ===
\node[lvl3, draw=cTeal!40, fill=cGrayFill] (l-autoreg) at (-0.5,  -11) {RT-1, RT-2\\OpenVLA, Octo};
\node[lvl3, draw=cTeal!40, fill=cGrayFill] (l-flow)    at (4.5,   -11) {$\pi_0$, $\pi_{0.5}$\\$\pi_0^*$};
\node[lvl3, draw=cTeal!40, fill=cGrayFill] (l-diff)    at (9,     -11) {DP, RDT-1B\\CogACT};
\node[lvl3, draw=cTeal!40, fill=cGrayFill] (l-hybrid)  at (13,    -11) {HybridVLA\\FAST};

\node[lvl3, draw=cViolet!40, fill=cGrayFill] (l-discrete)   at (23,   -11) {RT-2 tokens\\FAST};
\node[lvl3, draw=cViolet!40, fill=cGrayFill] (l-continuous) at (27.5,  -11) {Flow, Diff.\\RTC, BID};

\node[lvl3, draw=cGreen!40, fill=cGrayFill] (l-anav)   at (38,    -11) {VLN, LLM nav.\\E2E flight};
\node[lvl3, draw=cGreen!40, fill=cGrayFill] (l-amanip) at (44,    -11) {Grasp, Transport\\Inspection};
\node[lvl3, draw=cGreen!40, fill=cGrayFill] (l-amulti) at (50,    -11) {Swarm, UAV-UGV\\Formation};

% === Tier 2 L2 (y=-14.5): Training, Bimanual children ===
\node[lvl2, draw=cOrange!70, fill=cOrangePale] (pretrain)  at (12.5,  -14.5) {Pre-training};
\node[lvl2, draw=cOrange!70, fill=cOrangePale] (posttrain) at (18,    -14.5) {Post-training};
\node[lvl2, draw=cOrange!70, fill=cOrangePale] (rl)        at (23,    -14.5) {RL};

\node[lvl2, draw=cPink!70, fill=cPinkPale] (coord) at (31,    -14.5) {Coordination};
\node[lvl2, draw=cPink!70, fill=cPinkPale] (tasks) at (36,    -14.5) {Tasks};

% === Tier 2 L3 (y=-17.5): Training, Bimanual leaves ===
\node[lvl3, draw=cOrange!40, fill=cGrayFill] (l-pretrain)  at (12.5,  -17.5) {OXE, DROID\\Bridge};
\node[lvl3, draw=cOrange!40, fill=cGrayFill] (l-posttrain) at (18,    -17.5) {Fine-tuning\\Co-training};
\node[lvl3, draw=cOrange!40, fill=cGrayFill] (l-rl)        at (23,    -17.5) {RECAP\\$\pi_0^*$};

\node[lvl3, draw=cPink!40, fill=cGrayFill] (l-coord) at (31,    -17.5) {Joint, Indep.\\Hierarchical};
\node[lvl3, draw=cPink!40, fill=cGrayFill] (l-tasks) at (36,    -17.5) {Contact-rich\\Deformable};

% === Edges: Root -> Level 1 ===
\draw[arr] (root) -- (arch);
\draw[arr] (root) -- (train);
\draw[arr] (root) -- (act);
\draw[arr] (root) -- (biman);
\draw[arr] (root) -- (aerial);

% === Edges: Arch/Actions/Aerial L1 -> L2 (direct) ===
\foreach \parent/\child in {arch/autoreg, arch/flow, arch/diff, arch/hybrid,
                            act/discrete, act/continuous,
                            aerial/anav, aerial/amanip, aerial/amulti} {
    \draw[arr] (\parent) -- (\child);
}

% === Edges: Training L1 -> L2 (orthogonal routing below Tier 1) ===
\draw[arr] (train.south) -- (18, -12.5) -- (12.5, -12.5) -- (pretrain.north);
\draw[arr] (train.south) -- (18, -12.5) -- (posttrain.north);
\draw[arr] (train.south) -- (18, -12.5) -- (23, -12.5) -- (rl.north);

% === Edges: Bimanual L1 -> L2 (orthogonal routing below Tier 1) ===
\draw[arr] (biman.south) -- (33.5, -12.5) -- (31, -12.5) -- (coord.north);
\draw[arr] (biman.south) -- (33.5, -12.5) -- (36, -12.5) -- (tasks.north);

% === Edges: Level 2 -> Level 3 ===
\foreach \parent/\child in {autoreg/l-autoreg, flow/l-flow, diff/l-diff, hybrid/l-hybrid,
                            pretrain/l-pretrain, posttrain/l-posttrain, rl/l-rl,
                            discrete/l-discrete, continuous/l-continuous,
                            coord/l-coord, tasks/l-tasks,
                            anav/l-anav, amanip/l-amanip, amulti/l-amulti} {
    \draw[arr] (\parent) -- (\child);
}

% Bounding box
\path (-4, -19.5) rectangle (54, 4);
\end{tikzpicture}%
}% end resizebox
\caption{Taxonomy %
 of VLA models for bimanual manipulation and unmanned aerial robotics. This review is organized along five major dimensions: architectural foundations (autoregressive, flow-based, diffusion-based, hybrid), training recipes (pre-training, post-training, reinforcement learning), action representations (discrete tokenization, continuous generation), bimanual-specific concerns (coordination strategies, task types), and unmanned aerial robotics (navigation, aerial manipulation, multi-agent unmanned systems). Each branch is covered in a dedicated section.}
\label{fig:taxonomy}
\end{figure}

We begin by formalizing the core concepts that underpin the review: the VLA policy, action chunking, flow matching for action generation, and bimanual coordination. The notation introduced here is used consistently in subsequent sections; Table~\ref{tab:notation} provides a summary. Figure~\ref{fig:taxonomy} presents the taxonomy that organizes this review.

% ============================================================
% TABLE: NOTATION
% ============================================================
\begin{table}[H]
%\centering
\caption{Summary %
 of notation used in this review.}
\label{tab:notation}
\begin{tabularx}{\textwidth}{XX}
\toprule
\textbf{Symbol} & \textbf{Description} \\
\midrule
$\pi_\theta$ & VLA policy parameterized by $\theta$ \\
$\mathbf{o}_t$ & Visual %
 observation at time $t$ \\
$\ell$ & Natural-language instruction \\
$\mathbf{q}_t$ & Proprioceptive state (joint positions) \\
$\mathbf{a}_t$ & Single-step action \\
$\mathbf{A}_t$ & Action chunk of horizon $H$ \\

\bottomrule
\end{tabularx}
\end{table}

\begin{table}[H]\ContinuedFloat

\caption{{\em Cont.}}

%AUTHOR FIX: duplicate \label{tab:notation} removed (continuation table inherits the label from the first part above).
\begin{tabularx}{\textwidth}{XX}
\toprule
\textbf{Symbol} & \textbf{Description} \\
\midrule

$H$ & Action chunk horizon (number of steps) \\
$T$ & Terminal (final) time step of an episode \\

$\boldsymbol{\tau}$ & Trajectory $(\mathbf{a}_0, \ldots, \mathbf{a}_{T-1})$ \\

$\mathbf{g}$ & Goal state \\
$\mathcal{S}(\ell, \mathbf{o}_T)$ & Task completion indicator \\

$d_a$ & Action dimensionality \\
$K$ & Number of denoising/flow steps \\
$\mathbf{v}_\theta$ & Learned velocity field (flow matching) \\
$\mathbf{a}_t^L, \mathbf{a}_t^R$ & Left and right arm actions \\
$f_{\text{vis}}, f_{\text{VLM}}, f_{\text{act}}$ & Visual encoder, VLM backbone, action head \\
$\alpha, \gamma, \sigma_k$ & Diffusion schedule coefficients (Equation~(\ref{eq:diffusion})) \\
$\mathbf{z}$ & Gaussian noise, $\mathbf{z} \sim \mathcal{N}(\mathbf{0}, \mathbf{I})$ \\
$\epsilon_\theta$ & Noise prediction network (diffusion) \\
\bottomrule
\end{tabularx}
\end{table}

\vspace{-14pt}

\subsection{VLA Policy Formulation}
\label{sec:problem:vla}

\textbf{Policy definition.} A Vision--Language--Action model defines a policy $\pi_\theta$ parameterized by $\theta$ that maps a visual observation $\mathbf{o}_t \in \mathcal{O}$, a language instruction $\ell \in \mathcal{L}$, and optionally a proprioceptive state $\mathbf{q}_t \in \mathcal{Q}$ to an action %
 $\mathbf{a}_t \in \mathcal{A}$:
\begin{linenomath}
\begin{equation}
    \pi_\theta : \mathcal{O} \times \mathcal{L} \times \mathcal{Q} \rightarrow \mathcal{A}.
    \label{eq:vla_policy}
\end{equation}
\end{linenomath}
The %
 observation space $\mathcal{O}$ typically consists of one or more camera images $\mathbf{I}_t \in \mathbb{R}^{H_{\text{img}} \times W_{\text{img}} \times 3}$. The language instruction $\ell$ is a natural-language string tokenized and embedded by the VLM backbone. The action space $\mathcal{A}$ varies by embodiment; for a single $n$-DOF arm with a gripper, $\mathbf{a}_t \in \mathbb{R}^{n+1}$, encoding either joint velocities or end-effector displacements plus a gripper command.

\textbf{Architecture.} The VLA framework distinguishes itself from prior vision-based control policies by sharing a backbone with a pre-trained VLM. Concretely, a VLA typically consists of three components: (1)~a visual encoder $f_{\text{vis}}$ that produces image tokens, (2)~a vision--language backbone $f_{\text{VLM}}$ that jointly reasons over image and text tokens, and (3)~an action head $f_{\text{act}}$ that decodes actions from the VLM's hidden representations:\vspace{-6pt}
\begin{linenomath}
\begin{equation}
    \mathbf{a}_t = f_{\text{act}}\!\left(f_{\text{VLM}}\!\left(f_{\text{vis}}(\mathbf{I}_t),\; \text{Tok}(\ell),\; \mathbf{q}_t\right)\right),
    \label{eq:vla_components}
\end{equation}
\end{linenomath}
where $\text{Tok}(\ell)$ denotes the tokenized language instruction. The proprioceptive state $\mathbf{q}_t$ is likewise tokenized and fed into $f_{\text{VLM}}$ alongside the visual and language tokens.

\textbf{Goal formalization.} A VLA policy is trained to reach a \emph{goal state} $\mathbf{g} \in \mathcal{G}$ specified implicitly by the language instruction $\ell$: the task is complete when the world state matches the intent of $\ell$. We formalize this with a task completion indicator:
\begin{linenomath}
\begin{equation}
    \mathcal{S}(\ell, \mathbf{o}_T) = \begin{cases} 1 & \text{if } \mathbf{o}_T \text{ satisfies } \ell, \\ 0 & \text{otherwise}, \end{cases}
    \label{eq:goal}
\end{equation}
\end{linenomath}
where $\mathbf{o}_T$ is the observation at the terminal step $T$ of an episode of length $T$. In practice, $\mathcal{S}$ is evaluated by a human judge, a VLM-based evaluator, or structured predicates. %

\textbf{Trajectory planning.} Over the same episode of $T$ steps, the VLA produces a trajectory $\boldsymbol{\tau} = (\mathbf{a}_0, \mathbf{a}_1, \ldots, \mathbf{a}_{T-1})$ by iteratively predicting and executing action chunks. No explicit trajectory optimization is performed; the trajectory emerges from successive chunk predictions, each conditioned on the latest observation:
\begin{linenomath}
\begin{equation}
    \boldsymbol{\tau} = \bigoplus_{k=0}^{\lceil T/H \rceil - 1} \pi_\theta(\mathbf{o}_{kH}, \ell, \mathbf{q}_{kH}),
    \label{eq:trajectory}
\end{equation}
\end{linenomath}
where $\boldsymbol{\tau}$ is the full trajectory over $T$ time steps, $k$ indexes successive chunk queries, $H$ is the chunk horizon, and $\bigoplus$ denotes concatenation. In hierarchical VLAs such as $\pi_{0.5}$, a high-level planner additionally decomposes $\ell$ into subgoal instructions $(\ell_1, \ell_2, \ldots)$ that each produce a trajectory segment.

\textbf{Language applicability.} Because VLM backbones inherit tokenizers trained on multi-lingual web corpora, VLA policies can in principle accept instructions in any language the tokenizer supports. In practice, however, all current VLA systems are trained and evaluated exclusively in English, and multi-lingual generalization has not been tested. Whether cross-lingual transfer degrades action prediction quality remains an open question.

\textbf{Illustrative examples.} To ground the formulation above, consider two representative tasks.
\emph{(i)~Bimanual manipulation:} given a camera image $\mathbf{I}_t$ of a towel on a table and the instruction $\ell =$ ``fold the towel,'' $f_{\text{vis}}$ extracts image tokens, $f_{\text{VLM}}$ fuses them with $\text{Tok}(\ell)$ to form a latent plan, and $f_{\text{act}}$ decodes an action chunk $\mathbf{A}_t$ that moves both arms to grasp opposite edges and bring them together. The goal indicator $\mathcal{S}$ returns 1 when the towel is folded at step $T$.
\emph{(ii)~Aerial navigation:} given a forward-facing camera image and $\ell =$ ``fly to the red building,'' the same pipeline produces a trajectory $\boldsymbol{\tau}$ of 3D waypoints; here, the action chunk encodes position commands rather than joint angles, and $\mathcal{S}$ returns 1 when the drone reaches the target.

\subsection{Action Chunking}
\label{sec:problem:chunking}

Rather than predicting a single action $\mathbf{a}_t$, modern VLA policies predict an \emph{action chunk}, a sequence of $H$ future actions, in a single forward pass:
\begin{linenomath}
\begin{equation}
    \mathbf{A}_t = (\mathbf{a}_t, \mathbf{a}_{t+1}, \ldots, \mathbf{a}_{t+H-1}) \in \mathbb{R}^{H \times d_a},
    \label{eq:action_chunk}
\end{equation}
\end{linenomath}
where $H$ is the \emph{chunk horizon} and $d_a$ is the action dimension. Action chunking, introduced in the context of \textbf{ACT}~\cite{zhao2023aloha}, offers two key advantages. First, it amortizes the cost of a single VLM forward pass over multiple control steps, which allows high-frequency control despite the latency of large models. Second, it captures temporal correlations between successive actions, producing smoother trajectories than single-step prediction. The chunk is typically executed open-loop or with temporal ensembling, where overlapping chunks are averaged to reduce jitter.

\subsection{Flow Matching for Action Generation}
\label{sec:problem:flow}

\emph{Flow matching}~\cite{lipman2022flow} provides a framework for learning continuous normalizing flows by regressing a vector field that transports samples from a simple prior $p_0$ (\eg, a %
 standard Gaussian) to the data distribution $p_1$. Given a time-dependent vector field $\mathbf{v}_\theta(\mathbf{x}, t)$ for $t \in [0, 1]$, the flow $\phi_t(\mathbf{x})$ satisfies (here, $t$ denotes the continuous flow time parameter, distinct from the discrete control step index used elsewhere):
\begin{linenomath}
\begin{equation}
    \frac{d}{dt}\phi_t(\mathbf{x}) = \mathbf{v}_\theta(\phi_t(\mathbf{x}), t), \quad \phi_0(\mathbf{x}) = \mathbf{x}, \quad \mathbf{x} \sim p_0.
    \label{eq:flow_ode}
\end{equation}
\end{linenomath}
where $\phi_t$ is the flow map at time $t$, transporting a sample from $p_0$ toward $p_1$. The training objective minimizes the conditional flow matching loss:
\begin{linenomath}
\begin{equation}
    \mathcal{L}_{\text{FM}} = \mathbb{E}_{t, \mathbf{x}_0, \mathbf{x}_1}\!\left[\left\| \mathbf{v}_\theta(\mathbf{x}_t, t) - (\mathbf{x}_1 - \mathbf{x}_0) \right\|^2\right],
    \label{eq:flow_loss}
\end{equation}
\end{linenomath}
where $\mathbf{x}_t = (1 - t)\mathbf{x}_0 + t\mathbf{x}_1$ is a linear interpolation. In the VLA context, $\mathbf{x}_1$ is the ground-truth action chunk $\mathbf{A}_t$ and $\mathbf{x}_0$ is Gaussian noise. \textbf{$\pi_0$}~\cite{black2024pi0} applies this formulation with a VLM backbone: the VLM hidden states condition the flow, and the action head iteratively denoises a noisy action chunk over $K$ steps during inference.

\subsection{Bimanual Coordination}
\label{sec:problem:bimanual}

For a bimanual system with a left arm and a right arm, the joint action space is:
\begin{linenomath}
\begin{equation}
    \mathbf{a}_t^{\text{bi}} = [\mathbf{a}_t^L \;;\; \mathbf{a}_t^R] \in \mathbb{R}^{d_L + d_R},
    \label{eq:bimanual_action}
\end{equation}
\end{linenomath}
where $\mathbf{a}_t^L \in \mathbb{R}^{d_L}$ and $\mathbf{a}_t^R \in \mathbb{R}^{d_R}$ are the actions for the left and right arms, respectively. For typical 7-DOF arms with grippers, $d_L = d_R = 8$ (7 joint positions or velocities + 1 gripper command), yielding $d_a = 16$. With action chunking of horizon $H$, the full bimanual action chunk has dimensionality $H \times (d_L + d_R)$, which for typical settings ($H = 50$, $d_a = 16$) reaches 800 dimensions.

Bimanual coordination can be categorized into three modes:
\begin{enumerate}
    \item \textbf{Independent}: each arm executes its own subtask without coupling (\eg, one arm picks an object while the other holds a container).
    \item \textbf{Loosely coupled}: arms must coordinate timing but not forces (\eg, handover tasks where one arm releases as the other grasps).
    \item \textbf{Tightly coupled}: arms must coordinate both motion and forces simultaneously (\eg, folding fabric where both arms must apply tension).
\end{enumerate}

\subsection{Scope of This Review}
\label{sec:problem:scope}

This review covers VLA models that integrate a pre-trained vision--language backbone with an action generation mechanism, with emphasis on their application to bimanual manipulation and unmanned aerial robotics. We include autoregressive, flow-based, diffusion-based, and hybrid architectures published through early 2026. We focus on learning-based approaches trained from demonstrations or reinforcement learning; classical motion planning, optimization-based bimanual coordination, and traditional PID-based drone controllers are outside our scope. For bimanual motion planning, we refer readers to Abbas~\etal~\cite{abbas2023dualarm}; for classical aerial control, we refer to standard flight dynamics references~\cite{mahony2012multirotor}. The inclusion of both manipulation and aerial domains reflects the cross-embodiment nature of VLAs: the same architectures and training recipes power policies across diverse robot morphologies.

With these definitions established, Section~\ref{sec:background} reviews the prerequisite concepts.

% ============================================================
% FIGURE 1: TAXONOMY
% ============================================================

% ============================================================
% 3. BACKGROUND
% ============================================================
\section{Background}
\label{sec:background}
%1. Please state the name of the manufacturer, city, state abbreviation (for USA and Canada only) and country from where the equipment was sourced. This applies to all first instances of each mentioned piece of equipment in the manuscript. E.g., Bravo Liquid Handler (Agilent Technologies Inc., Santa Clara, CA, USA)
%2. Please state the version number of software used in the study. If a version is not available, please provide a valid link to the software's website instead and state the access date (day, month and year). This applies to the first instance of each mentioned software in the manuscript. E.g., MATLAB (v2024)

Before surveying specific VLA methods, we review the foundational concepts they build upon: vision--language models, imitation learning, generative modeling for action generation, bimanual robotic systems, and aerial robotic systems.

\subsection{Vision--Language Models}
\label{sec:background:vlm}

Vision--Language Models (VLMs) jointly process visual and textual inputs, built upon the Transformer architecture~\cite{vaswani2017attention} and trained on internet-scale image--text datasets. The Vision Transformer (ViT)~\cite{dosovitskiy2021vit} extended self-attention to image patches, while \textbf{CLIP}~\cite{radford2021clip} established contrastive pre-training for aligned visual--textual representations. The pre-train-then-fine-tune recipe, scaled by GPT-3~\cite{brown2020gpt3} and refined via instruction-tuning~\cite{ouyang2022training}, is the foundation-model methodology that VLAs inherit.

Key VLMs relevant to this review include \textbf{PaLM-E}~\cite{driess2023palme}, which demonstrated embodied reasoning in a 562B-parameter model; \textbf{PaLIGemma}~\cite{beyer2024paligemma} and \textbf{Gemma}~\cite{team2024gemma}, which provide efficient open-weight backbones used by several VLA systems; and open-source models (\textbf{LLaMA}~\cite{touvron2023llama}, \textbf{LLaVA}~\cite{liu2024llava}) that democratized access. VLMs are attractive for robotics because they recognize objects, understand spatial relationships, and interpret instructions without robotics-specific training.

The transition from VLM to VLA requires adding an action output modality. This can be achieved by (1)~tokenizing actions as text tokens and fine-tuning the VLM's language head~\cite{brohan2023rt2}, (2)~attaching a separate action head that reads from the VLM's hidden states~\cite{black2024pi0}, or (3)~using the VLM as a high-level planner that conditions a low-level policy~\cite{shi2025hirobot}. Each approach trades off between exploiting pre-trained knowledge and accommodating the continuous, high-frequency nature of robot control. A limitation for robotics is that VLMs lack grounding in physical interaction dynamics; they recognize objects but cannot predict contact forces or material deformation.

\subsection{Imitation Learning}
\label{sec:background:il}

Imitation learning (IL) trains a policy $\pi_\theta$ to mimic expert demonstrations $\mathcal{D} = \{(\mathbf{o}_i, \ell_i, \mathbf{a}_i)\}_{i=1}^N$, dating back to \textbf{ALVINN}~\cite{pomerleau1989alvinn}. The simplest form, \emph{behavioral cloning} (BC), minimizes a supervised loss:
\begin{linenomath}
\begin{equation}
    \mathcal{L}_{\text{BC}} = \mathbb{E}_{(\mathbf{o}, \ell, \mathbf{a}) \sim \mathcal{D}}\!\left[\|\pi_\theta(\mathbf{o}, \ell) - \mathbf{a}\|^2\right].
    \label{eq:bc_loss}
\end{equation}
\end{linenomath}
BC suffers from compounding errors due to distribution shift~\cite{ross2011dagger}: at test time, the policy visits states not seen in training. Action chunking~\cite{zhao2023aloha} mitigates this by reducing decision points. A second challenge is \emph{multi-modality}: for a given observation, multiple valid action sequences may exist. Mean-squared-error regression averages over modes, producing invalid intermediate actions. Bimanual tasks amplify both problems because the state space is higher-dimensional and errors propagate across both arms. This motivates expressive generative models (diffusion, flow matching, autoregressive sampling) as action decoders. \textbf{Language-conditioned IL}~\cite{stepputtis2020lcil} extends BC by conditioning on language instructions; VLAs take this further by using pre-trained VLMs for rich semantic grounding.

\subsection{Generative Modeling for Actions}
\label{sec:background:generative}

Three families of generative models underpin VLA action generation. Early approaches used VAEs~\cite{kingma2014vae} and GANs~\cite{goodfellow2014gan} for latent action representations. DDPMs~\cite{ho2020ddpm} and score-based models~\cite{song2021score} provided higher-fidelity generation at the cost of iterative sampling, with \textbf{Latent Diffusion Models}~\cite{rombach2022ldm} reducing this cost via learned latent spaces. The \textbf{Decision Transformer}~\cite{chen2021decisiontransformer} reframed RL as sequence modeling, foreshadowing the autoregressive approach that VLAs later adopted, and \textbf{Gato}~\cite{reed2022generalist} extended this to a generalist agent handling text, images, and robotic actions in one Transformer.

\subsubsection{Autoregressive Models}
Autoregressive models factorize the action distribution as a product of conditionals:
\begin{linenomath}
\begin{equation}
    p(\mathbf{A}_t | \mathbf{o}_t, \ell) = \prod_{h=0}^{H-1} p(\mathbf{a}_{t+h} | \mathbf{a}_{t:t+h-1}, \mathbf{o}_t, \ell).
    \label{eq:autoregressive}
\end{equation}
\end{linenomath}
\textbf{RT-2}~\cite{brohan2023rt2} discretizes continuous actions into 256 bins per dimension and generates action tokens left-to-right using the VLM's language modeling head. This approach naturally exploits VLM pre-training but introduces quantization error and sequential latency that scales with action dimensionality.

\subsubsection{Diffusion Models}
\textbf{Diffusion Policy}~\cite{chi2023diffusion} generates actions by iteratively denoising a Gaussian sample through a learned reverse diffusion process:
\begin{linenomath}
\begin{equation}
    \mathbf{A}_t^{(k-1)} = \alpha\!\left(\mathbf{A}_t^{(k)} - \gamma \epsilon_\theta(\mathbf{A}_t^{(k)}, k, \mathbf{o}_t)\right) + \sigma_k \mathbf{z},
    \label{eq:diffusion}
\end{equation}
\end{linenomath}
where $\epsilon_\theta$ is the noise prediction network, $k$ indexes the denoising step, $\mathbf{z} \sim \mathcal{N}(\mathbf{0}, \mathbf{I})$ is standard Gaussian noise, and $\alpha, \gamma, \sigma_k$ are schedule-dependent coefficients. Diffusion models excel at capturing distributions over multiple valid actions and produce smooth trajectories but require multiple denoising steps ($K = 10$--$100$), increasing inference latency.

\subsubsection{Flow Matching}
Flow matching~\cite{lipman2022flow}, formalized in Equation~(\ref{eq:flow_loss}), offers a simpler training objective and often requires fewer steps than diffusion. \textbf{Rectified Flow}~\cite{liu2022rectified} straightens transport paths to reduce integration steps. \textbf{$\pi_0$}~\cite{black2024pi0} demonstrated that flow matching with $K = 10$ steps produces high-quality action chunks at $50\,\text{Hz}$ for bimanual systems.

\subsection{Bimanual Robotic Systems}
\label{sec:background:bimanual}

Three hardware platforms have transformed bimanual VLA research (see Section~\ref{sec:bimanual} %
 for specifications). \textbf{ALOHA}~\cite{zhao2023aloha} provides low-cost bilateral teleoperation for two 6-DOF arms, paired with the ACT policy (Action Chunking with Transformers) that predicts action chunks at $50\,\text{Hz}$. \textbf{Mobile ALOHA}~\cite{fu2024mobilealoha} extends this to a mobile base and demonstrated co-training (mixing target-task data with diverse data), which directly influenced VLA training recipes (Section~\ref{sec:training:posttrain}). \textbf{UMI}~\cite{chi2024umi} decouples data collection from the robot via hand-held grippers with visual--inertial tracking, allowing demonstrations in diverse environments without teleoperation hardware. The standardization of action spaces across these platforms has facilitated cross-system transfer; data collection strategies are detailed in Section~\ref{sec:training:data}.

Two practical concerns affect bimanual VLA deployment. \emph{Calibration}: even small errors ($\sim$1\,cm position, $\sim$2$^\circ$ orientation) between arms can cause policies to fail; \textbf{UMI}~\cite{chi2024umi} addresses this via visual--inertial tracking that decouples data collection from arm calibrations. \emph{Action space choice}: joint-space actions (\textbf{ACT}~\cite{zhao2023aloha}, \textbf{RDT-1B}~\cite{liu2024rdt1b}) provide direct control but are embodiment-specific, while end-effector actions (\textbf{$\pi_0$}~\cite{black2024pi0}) facilitate cross-embodiment transfer at the cost of inverse kinematics errors.

\subsection{Aerial Robotic Systems}
\label{sec:background:aerial}

A quadrotor is a 6-DOF rigid body (3 translational, 3 rotational) controlled through differential thrust of four rotors, making it underactuated (4 inputs for 6 DOF). This underactuation creates inherent coupling between translational and rotational motion that complicates learned control policies. Quadrotors are the dominant platform for learning-based unmanned aerial robotics due to their mechanical simplicity, hovering capability, and commercial availability.

Traditional drone control employs cascaded PID loops operating at ${\geq}250\,\text{Hz}$, with an inner attitude loop and an outer position loop. Learning-based approaches replace part or all of this pipeline with neural network policies. The action space varies from high-level waypoints (suitable for navigation VLAs operating at $5$--$10\,\text{Hz}$) to low-level motor commands (required for agile flight at ${\geq}100\,\text{Hz}$). This range of control frequencies and abstraction levels parallels the hierarchy observed in manipulation VLAs, from high-level subgoal generation ($\pi_{0.5}$) to low-level continuous action chunks ($\pi_0$).

High-fidelity simulators play an outsized role in aerial VLA development. \textbf{AirSim}~\cite{shah2018airsim} provides photorealistic rendering via Unreal Engine with accurate quadrotor dynamics. \textbf{Flightmare}~\cite{song2021flightmare} decouples rendering from physics, allowing massively parallel RL training at $200\times$ real-time. These simulators are to aerial VLAs what LIBERO~\cite{liu2024libero} (a suite of language-conditioned manipulation tasks) and SIMPLER~\cite{li2024simpler} (a simulation-to-real evaluation framework) are to manipulation VLAs: essential infrastructure for training and evaluation at scale. Both benchmarks are described in detail in Section~\ref{sec:benchmarks:sim}.

The datasets and benchmarks that drive VLA development and evaluation are reviewed in the next section.

% ============================================================
% 4. DATASETS, BENCHMARKS, AND EVALUATION
% ============================================================
\section{Datasets, Benchmarks, and Evaluation}
\label{sec:benchmarks}

Large-scale datasets and standardized benchmarks form the infrastructure that drives VLA research. We review the major datasets used for pre-training and evaluation, simulation benchmarks, and the metrics employed to assess bimanual manipulation and aerial navigation performance.

\subsection{Pre-Training Datasets}
\label{sec:benchmarks:datasets}

VLA training relies on large-scale robot demonstration data for pre-training. Table~\ref{tab:datasets} compares the major datasets. Three have proved most influential.

\textbf{Open X-Embodiment (OXE)}~\cite{oxe2024} is the largest open robot dataset, aggregating demonstrations from over 20 institutions across 22 robot embodiments. It contains more than 1 million episodes spanning single-arm, bimanual, and mobile manipulation tasks. OXE's diversity in embodiments, viewpoints, and environments makes it the standard pre-training corpus for cross-embodiment VLAs. \textbf{OpenVLA}~\cite{kim2024openvla}, \textbf{Octo}~\cite{octo2024}, and \textbf{$\pi_0$}~\cite{black2024pi0} all use OXE (or subsets thereof) for pre-training.

\textbf{DROID}~\cite{khazatsky2024droid} provides 76,000 trajectories collected across 564 scenes and 86 tasks using Franka Emika arms. Unlike OXE, DROID emphasizes diversity within a single embodiment: 50 operators collected data across varied environments, capturing natural scene diversity. DROID has been shown to improve generalization when included in VLA pre-training mixtures.

\textbf{BridgeData V2}~\cite{walke2023bridge}, building on the original BridgeData~\cite{ebert2021bridge} that first demonstrated cross-domain dataset boosting, contains 60,096 trajectories from a WidowX robot performing tabletop manipulation tasks across 24 environments. Its relatively uniform setup and reliable labeling make it a standard evaluation dataset. Many VLA papers report results on Bridge tasks.

\textbf{GigaBrain-0.5M}~\cite{gigabrain2025} is a recent large-scale dataset containing 500,000 episodes collected via a combination of teleoperation and autonomous data collection. It includes bimanual manipulation episodes and was designed to support VLA training with world-model-based reinforcement learning.

\begin{table}[H]
%\centering
\caption{Comparison %
 of major robot datasets used for VLA pre-training and evaluation. Columns include the number of episodes, robot embodiments covered, task count, scene diversity (distinct environments), bimanual support, language annotation availability, action space type, and primary manipulation domain. In this and all subsequent comparison tables, \checkmark indicates that the feature/capability is supported, while ``--'' or a blank entry indicates that the feature is not supported, not applicable, or not reported in the original publication.}
 
\begin{adjustwidth}{-\extralength}{0cm}
%\centering %% If there is a figure in wide page, please release command \centering

\fontsize{6.5}{6.5}\selectfont

\label{tab:datasets}
%\resizebox{\fulllength}{!}{%
%\begin{tabular}{}
%\toprule
\begin{tabularx}{\fulllength}{lcccccccccc}
\toprule
\textbf{Dataset} & \textbf{Episodes} & \textbf{Embodiments} & \textbf{Tasks} & \textbf{Scenes} & \textbf{Bimanual} & \textbf{Language} & \textbf{Action Space} & \textbf{Domain} & \textbf{Open Access} & \textbf{Year} \\
\midrule
\textbf{OXE %
}~\cite{oxe2024} & $>$1M %
 & 22 & 527 & 20+ inst. & \checkmark & \checkmark & Mixed (joint/EE) & Tabletop + mobile & \checkmark & 2024 \\
\textbf{DROID}~\cite{khazatsky2024droid} & 76K & 1 (Franka) & 86 & 564 & -- & \checkmark & EE delta & Tabletop (in-the-wild) & \checkmark & 2024 \\
\textbf{BridgeData V2}~\cite{walke2023bridge} & 60K & 1 (WidowX) & 13 & 24 & -- & \checkmark & EE delta & Tabletop & \checkmark & 2023 \\
\textbf{ALOHA}~\cite{zhao2023aloha} & $\sim$1K & 1 (ALOHA) & 6 & 1 (lab) & \checkmark & -- & Joint pos. & Bimanual fine-grained & \checkmark & 2023 \\
\textbf{GigaBrain-0.5M}~\cite{gigabrain2025} & 500K & Multiple & 200+ & Diverse & \checkmark & \checkmark & Mixed & Household + manip. & \checkmark & 2025 \\
\bottomrule
\end{tabularx}\end{adjustwidth}
\end{table}

\subsection{Simulation Benchmarks}
\label{sec:benchmarks:sim}

Simulation benchmarks support reproducible evaluation at scale without the expense and variability of real-world experiments.

\textbf{LIBERO}~\cite{liu2024libero} is a benchmark suite of 130 language-conditioned manipulation tasks organized into five suites: LIBERO-Spatial (spatial relationship understanding), LIBERO-Object (novel object generalization), LIBERO-Goal (goal specification comprehension), LIBERO-Long (multi-step task execution), and LIBERO-100 (a larger training set). The four evaluation suites each contain 10 tasks with 50 demonstrations per task. LIBERO has become the primary simulation benchmark for VLA evaluation because it tests multiple generalization axes independently, allowing researchers to diagnose specific weaknesses.

\textbf{SIMPLER}~\cite{li2024simpler} provides simulated counterparts to real-world evaluation setups, which allows VLA evaluation without physical hardware. It includes tasks from the Bridge and Google Robot environments, with visual fidelity and physics parameters calibrated to correlate with real-world performance. SIMPLER's key contribution is demonstrating that simulation success rates predict real-world success rates with $r > 0.8$ correlation for most task categories. This validates simulation as a low-cost proxy for real evaluation.

Other simulation platforms include \textbf{RLBench}~\cite{james2020rlbench} (100 procedurally generated tasks), \textbf{Meta-World}~\cite{yu2020metaworld} (50 parametric tasks), \textbf{RoboSuite}~\cite{zhu2020robosuite}, \textbf{ManiSkill2}~\cite{gu2023maniskill2} (soft-body tasks), and \textbf{BEHAVIOR-1K}~\cite{lee2024behavior} (1000 %
 household activities). Li~\etal~\cite{li2024simpler} found that both physics fidelity and visual realism matter for reliable simulation-to-real prediction.

Neither LIBERO nor SIMPLER includes bimanual tasks, a significant limitation. Bimanual simulation requires dual-arm physics and contact-rich interaction modeling that no standard benchmark provides.

\subsection{Evaluation Metrics}
\label{sec:benchmarks:metrics}

VLA evaluation relies on several complementary metrics:

\textbf{Task success rate} is the primary metric, defined as the fraction of $N$ evaluation episodes in which the robot completes the specified task:
\begin{linenomath}
\begin{equation}
    \text{SR} = \frac{1}{N}\sum_{i=1}^{N} \mathbb{1}[\text{task}_i \text{ completed}],
    \label{eq:success_rate}
\end{equation}
\end{linenomath}
where $\mathbb{1}[\cdot]$ is the indicator function.

\textbf{Normalized score} averages success rates across multiple task suites, which allows comparison across benchmarks with different numbers of tasks.

\textbf{Inference latency} measures the wall-clock time for a single action chunk prediction, critical for real-time control. In dual-arm setups running at $50\,\text{Hz}$, the action generation must complete within $20\,\text{ms}$ per step (or within $H \times 20\,\text{ms}$ per chunk).

\textbf{Data efficiency} tracks how many demonstrations are required to reach a target success rate, relevant for bimanual tasks where data collection is expensive.

% ============================================================
% TABLE 1: DATASETS
% ============================================================

Bimanual evaluation protocols remain less standardized. Most papers define their own task suites (\eg, \textbf{$\pi_0$}~\cite{black2024pi0} evaluates on laundry folding, box assembly, and table busing), making cross-method comparison difficult. Task completion criteria also vary: some papers use binary success, others use partial completion scores, and temporal efficiency is rarely reported.

\textbf{Generalization metrics} assess whether the VLA transfers to novel settings:
\begin{linenomath}
\begin{equation}
    \text{Gen}(\pi_\theta) = \frac{\text{SR}_{\text{novel}}}{\text{SR}_{\text{train}}},
    \label{eq:generalization}
\end{equation}
\end{linenomath}
where $\text{SR}_{\text{novel}}$ and $\text{SR}_{\text{train}}$ are success rates on novel and training environments, respectively. A generalization ratio near 1.0 indicates robust transfer. \textbf{RT-2}~\cite{brohan2023rt2} reported a generalization ratio of $\sim$0.76 for novel objects, while \textbf{$\pi_{0.5}$}~\cite{black2025pi05} achieved $\sim$0.95 for novel homes, indicating strong environment generalization.

Bimanual-specific benchmarks remain limited: most simulation suites focus on single-arm tasks, and real-world bimanual evaluation varies across papers. This gap motivates the need for standardized bimanual benchmarks (Section~\ref{sec:discussion}).

% ============================================================
% FIGURE 2: TIMELINE
% ============================================================

% ============================================================
% 5. VLA ARCHITECTURES AND FOUNDATIONS
% ============================================================
\section{VLA Architectures and Foundations}
\label{sec:architectures}

\begin{figure}[H]
\centering
\resizebox{\textwidth}{!}{%
\begin{tikzpicture}[
    every node/.style={font=\scriptsize},
    ar/.style={rectangle, rounded corners=3pt, draw=cBlue, fill=cBlueFill, minimum height=6mm, font=\scriptsize\bfseries, inner sep=2pt, line width=0.6pt},
    fl/.style={rectangle, rounded corners=3pt, draw=cRed, fill=cRedFill, minimum height=6mm, font=\scriptsize\bfseries, inner sep=2pt, line width=0.6pt},
    di/.style={rectangle, rounded corners=3pt, draw=cGreen, fill=cGreenFill, minimum height=6mm, font=\scriptsize\bfseries, inner sep=2pt, line width=0.6pt},
    hw/.style={rectangle, rounded corners=3pt, draw=cOrange, fill=cOrangeFill, minimum height=6mm, font=\scriptsize\bfseries, inner sep=2pt, line width=0.6pt},
    hy/.style={rectangle, rounded corners=3pt, draw=cViolet, fill=cVioletFill, minimum height=6mm, font=\scriptsize\bfseries, inner sep=2pt, line width=0.6pt},
]
% Timeline axis
\draw[-{Stealth[length=3mm]}, thick] (0,0) -- (19,0);
\foreach \x/\yr in {1/2022, 4.5/2023, 8.5/2024, 13/2025, 16.5/2026} {
    \draw[thick] (\x, -0.15) -- (\x, 0.15);
    \node[below] at (\x, -0.2) {\yr};
}

% 2022
\node[ar] at (1.8, 1.2) {RT-1};
\draw[->, thin] (1.8, 0.9) -- (1.8, 0.15);
\node[di] at (2.8, 2.2) {Diff.~Policy};
\draw[->, thin] (2.8, 1.9) -- (2.8, 0.15);

% 2023
\node[ar] at (4.2, 1.2) {RT-2};
\draw[->, thin] (4.2, 0.9) -- (4.2, 0.15);
\node[hw] at (5.5, 2.2) {ALOHA};
\draw[->, thin] (5.5, 1.9) -- (5.5, 0.15);
\node[ar] at (6.5, 1.2) {PaLM-E};
\draw[->, thin] (6.5, 0.9) -- (6.5, 0.15);

% 2024
\node[ar] at (7.8, 1.2) {OpenVLA};
\draw[->, thin] (7.8, 0.9) -- (7.8, 0.15);
\node[ar] at (8.5, 2.2) {Octo};
\draw[->, thin] (8.5, 1.9) -- (8.5, 0.15);
\node[fl] at (9.5, 1.2) {$\pi_0$};
\draw[->, thin] (9.5, 0.9) -- (9.5, 0.15);
\node[di] at (10.3, 2.2) {RDT-1B};
\draw[->, thin] (10.3, 1.9) -- (10.3, 0.15);
\node[hw] at (10.3, 3.2) {UMI};
\draw[->, thin] (10.3, 2.9) -- (10.3, 2.55);

% 2025
\node[fl] at (11.8, 1.2) {$\pi_{0.5}$};
\draw[->, thin] (11.8, 0.9) -- (11.8, 0.15);
\node[fl] at (12.8, 2.2) {$\pi_0^*$};
\draw[->, thin] (12.8, 1.9) -- (12.8, 0.15);
\node[hy] at (13.8, 1.2) {HybridVLA};
\draw[->, thin] (13.8, 0.9) -- (13.8, 0.15);
\node[hy] at (14.8, 2.2) {Hi Robot};
\draw[->, thin] (14.8, 1.9) -- (14.8, 0.15);
\node[hy] at (15.5, 1.2) {FAST};
\draw[->, thin] (15.5, 0.9) -- (15.5, 0.15);
% 2026
\node[fl] at (17, 1.2) {MEM};
\draw[->, thin] (17, 0.9) -- (17, 0.15);

% Legend background box
\fill[cGrayFill, rounded corners=4pt] (-0.2, -1.7) rectangle (11.6, -0.65);
\draw[cGray!40, rounded corners=4pt, line width=0.4pt] (-0.2, -1.7) rectangle (11.6, -0.65);
\node[font=\scriptsize\bfseries, text=cGray] at (-0.2, -0.85) [right] {Legend:};
% Legend items
\node[ar, minimum width=14mm] at (2.2, -1.2) {AR};
\node[fl, minimum width=14mm] at (4.2, -1.2) {Flow};
\node[di, minimum width=14mm] at (6.2, -1.2) {Diffusion};
\node[hw, minimum width=14mm] at (8.2, -1.2) {Hardware};
\node[hy, minimum width=14mm] at (10.4, -1.2) {Hybrid};
% Bounding box
\path (-0.5, -2) rectangle (19.5, 3.8);
\end{tikzpicture}%
}% end resizebox
\caption{Timeline %
 of key VLA and bimanual manipulation milestones (2022--2026). Colors indicate the architectural family: autoregressive (blue), flow-based (red), diffusion-based (green), hardware platforms (orange), and hybrid/efficient methods (purple). The field has accelerated rapidly, with the majority of VLA contributions appearing in 2024--2026.}
\label{fig:timeline}
\end{figure}
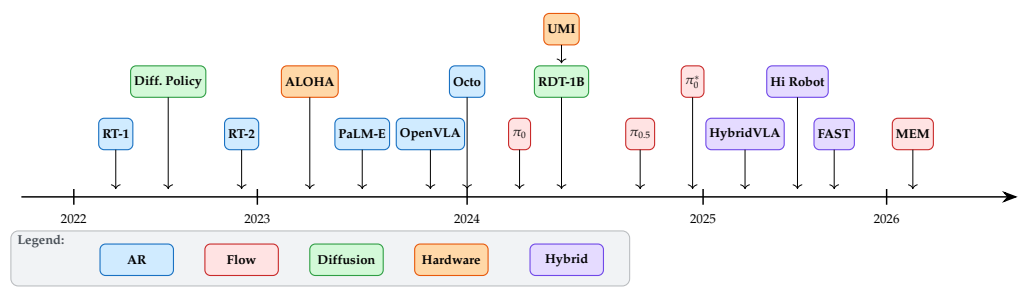

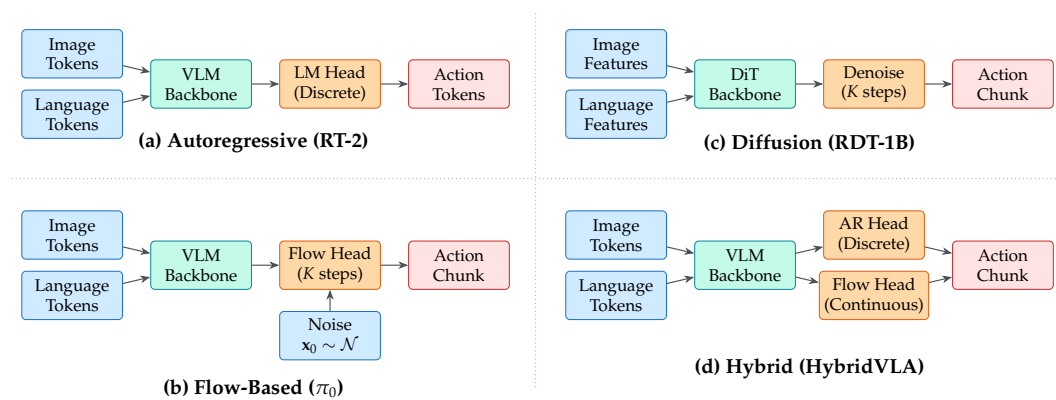
\begin{figure}[H]

\resizebox{\textwidth}{!}{%
\begin{tikzpicture}[
    every node/.style={font=\large, align=center},
    block/.style={rectangle, draw, rounded corners=3pt, minimum width=25mm, minimum height=12mm, font=\large, line width=0.6pt},
    input/.style={block, draw=cBlue, fill=cBlueFill},
    vlm/.style={block, draw=cTeal, fill=cTealFill},
    head/.style={block, draw=cOrange, fill=cOrangeFill},
    output/.style={block, draw=cRed, fill=cRedFill},
    arr/.style={-{Stealth[length=2.5mm]}, thick, draw=cGray},
]

% Autoregressive
\node[font=\Large\bfseries] at (4.5, 0.8) {(a) Autoregressive (RT-2)};
\node[input] (a_img) at (0, 3) {Image\\Tokens};
\node[input] (a_lang) at (0, 1.5) {Language\\Tokens};
\node[vlm] (a_vlm) at (3.2, 2.25) {VLM\\Backbone};
\node[head] (a_head) at (6.4, 2.25) {LM Head\\(Discrete)};
\node[output] (a_out) at (9.6, 2.25) {Action\\Tokens};
\draw[arr] (a_img) -- (a_vlm);
\draw[arr] (a_lang) -- (a_vlm);
\draw[arr] (a_vlm) -- (a_head);
\draw[arr] (a_head) -- (a_out);

% Flow-Based
\node[font=\Large\bfseries] at (4.5, -5.3) {(b) Flow-Based ($\pi_0$)};
\node[input] (b_img) at (0, -1.5) {Image\\Tokens};
\node[input] (b_lang) at (0, -3) {Language\\Tokens};
\node[vlm] (b_vlm) at (3.2, -2.25) {VLM\\Backbone};
\node[head] (b_head) at (6.4, -2.25) {Flow Head\\($K$ steps)};
\node[output] (b_out) at (9.6, -2.25) {Action\\Chunk};
\node[input] (b_noise) at (6.4, -4) {Noise\\$\mathbf{x}_0 \sim \mathcal{N}$};
\draw[arr] (b_img) -- (b_vlm);
\draw[arr] (b_lang) -- (b_vlm);
\draw[arr] (b_vlm) -- (b_head);
\draw[arr] (b_head) -- (b_out);
\draw[arr] (b_noise) -- (b_head);

% Diffusion
\node[font=\Large\bfseries] at (18.3, 0.8) {(c) Diffusion (RDT-1B)};
\node[input] (c_img) at (13.5, 3) {Image\\Features};
\node[input] (c_lang) at (13.5, 1.5) {Language\\Features};
\node[vlm] (c_vlm) at (16.7, 2.25) {DiT\\Backbone};
\node[head] (c_head) at (19.9, 2.25) {Denoise\\($K$ steps)};
\node[output] (c_out) at (23.1, 2.25) {Action\\Chunk};
\draw[arr] (c_img) -- (c_vlm);
\draw[arr] (c_lang) -- (c_vlm);
\draw[arr] (c_vlm) -- (c_head);
\draw[arr] (c_head) -- (c_out);

% Hybrid
\node[font=\Large\bfseries] at (18.3, -4.8) {(d) Hybrid (HybridVLA)};
\node[input] (d_img) at (13.5, -1.5) {Image\\Tokens};
\node[input] (d_lang) at (13.5, -3) {Language\\Tokens};
\node[vlm] (d_vlm) at (16.7, -2.25) {VLM\\Backbone};
\node[head] (d_head1) at (19.9, -1.5) {AR Head\\(Discrete)};
\node[head] (d_head2) at (19.9, -3) {Flow Head\\(Continuous)};
\node[output] (d_out) at (23.1, -2.25) {Action\\Chunk};
\draw[arr] (d_img) -- (d_vlm);
\draw[arr] (d_lang) -- (d_vlm);
\draw[arr] (d_vlm) -- (d_head1);
\draw[arr] (d_vlm) -- (d_head2);
\draw[arr] (d_head1) -- (d_out);
\draw[arr] (d_head2) -- (d_out);

% Separation lines between (a)-(d)
\draw[dotted, thick, cGray!40] (11.5, 4) -- (11.5, -5.3);  % vertical
\draw[dotted, thick, cGray!40] (-1.5, -0.1) -- (24.5, -0.1); % horizontal
\end{tikzpicture}%
}% end resizebox
\caption{Architectural %
 comparison of the four VLA families. (\textbf{a})~Autoregressive VLAs (RT-2, OpenVLA) discretize actions and generate them as language tokens. (\textbf{b})~Flow-based VLAs ($\pi_0$) use a flow-matching head that iteratively denoises a noise sample conditioned on VLM features. (\textbf{c})~Diffusion VLAs (RDT-1B) use a Diffusion Transformer to denoise action chunks. (\textbf{d})~Hybrid VLAs (HybridVLA) combine autoregressive and flow heads for discrete and continuous action components, respectively.}
\label{fig:architecture_comparison}
\end{figure}

This section presents VLA architectures organized by their action generation mechanism. We identify four families: autoregressive, flow-based, diffusion-based, and hybrid. Figure~\ref{fig:timeline} traces the chronological development of these methods from 2022 to 2025. Figure~\ref{fig:taxonomy} provides an overview of the families and their constituent methods, while Figure~\ref{fig:architecture_comparison} contrasts the four architectural paradigms side by side. Table~\ref{tab:architectures} compares representative methods across key architectural dimensions.
\vspace{-6pt}

\subsection{Autoregressive VLAs}
\label{sec:architectures:ar}

Autoregressive VLAs generate actions by extending the VLM's language modeling capability to action tokens. This approach directly builds on the pre-trained language model's sequential generation ability.

\subsubsection{RT-1 and RT-2}

\textbf{RT-1}~\cite{brohan2022rt1} was among the first Transformer-based robot policies trained on large-scale real-world data. It processes image histories through a FiLM-EfficientNet encoder and generates discretized actions via per-dimension classification heads. Trained on 130,000 demonstrations from Google's mobile manipulator fleet, RT-1 achieved 97\% success on seen tasks and 76\% on unseen tasks, establishing that data diversity produces meaningful generalization. RT-1 does not address bimanual manipulation, but its lessons directly informed subsequent VLA designs.

\textbf{RT-2}~\cite{brohan2023rt2} took the decisive step of unifying vision--language understanding and action generation within a single VLM. By fine-tuning a PaLI-X (55B parameters) or PaLM-E (12B parameters) model to output discretized actions as text tokens, RT-2 demonstrated that VLM pre-training on web data transfers to robotic manipulation. It exhibited capabilities absent from RT-1: reasoning about object categories, interpreting novel instructions, and performing rudimentary chain-of-thought planning for multi-step actions. Semantic knowledge encoded in VLM weights (object affordances, spatial relationships, physical intuition) directly benefits action generation. RT-2 remains closed-source, however, and its 55B-parameter scale demands TPU-class compute, putting real-time control out of reach for most research groups and limiting reproducibility.

\subsubsection{OpenVLA}

Reproducibility was a major barrier for VLA research until \textbf{OpenVLA}~\cite{kim2024openvla}, the first fully open-source 7B-parameter VLA. Based on the Prismatic VLM architecture, it tokenizes actions into 256 discrete bins per dimension following RT-2's approach and trains on the OXE dataset. Despite its open weights, OpenVLA matches \textbf{RT-2-X}~\cite{oxe2024} (the cross-embodiment variant of RT-2 trained on OXE alongside \textbf{RT-1-X}~\cite{oxe2024}). Its release catalyzed community research and revealed important scaling behaviors: performance improves consistently with data diversity, and fine-tuning on small task-specific datasets yields substantial gains over the pre-trained checkpoint. A subsequent study, \textbf{OpenVLA-OFT}~\cite{zheng2024openvla2}, introduced improved fine-tuning recipes optimizing both speed and success rate, narrowing the gap to proprietary systems.

\subsubsection{Octo}

Although Octo uses a diffusion action head, we discuss it here because it was designed as a generalist initialization for downstream fine-tuning, a role shared with autoregressive VLAs. Whereas OpenVLA adapts an existing VLM, \textbf{Octo}~\cite{octo2024} takes a different path: a purpose-built 93M-parameter Transformer with a diffusion action head, bridging the autoregressive and diffusion approaches. Prompted with language instructions or goal images, Octo supports flexible action spaces and was trained on 800,000 episodes from OXE. It serves as a versatile initialization for fine-tuning on downstream tasks, including bimanual manipulation with ALOHA. Because it uses a diffusion action head rather than autoregressive token prediction, Octo straddles both categories in our taxonomy (Figure~\ref{fig:taxonomy}). Its modest parameter count, however, restricts capacity for complex reasoning compared to VLM-based approaches.

Several other autoregressive VLAs have explored complementary design choices: \textbf{GR-1}~\cite{wu2023gr1} (video prediction as an implicit world model), \textbf{HAMSTER}~\cite{li2025hamster} (hierarchical vision--language--action prediction), \textbf{SpatialVLA}~\cite{chen2024spatialvla} (explicit spatial representations), \textbf{BAKU}~\cite{haldar2024baku} (efficient multi-task architecture), \textbf{KAT}~\cite{kim2024kat} (in-context imitation via keypoint-action tokens), and \textbf{SimpleVLA}~\cite{zhen2024simplevla} (minimal design matching elaborate systems with RL fine-tuning).

\subsection{Flow-Based VLAs}
\label{sec:architectures:flow}

Flow-based VLAs use flow matching (Section~\ref{sec:problem:flow}) to generate continuous action chunks, avoiding the discretization bottleneck of autoregressive approaches. As shown in Figure~\ref{fig:architecture_comparison}b, the flow head iteratively denoises a noise sample conditioned on VLM features.

\subsubsection{$\pi_0$}

The discretization bottleneck of autoregressive VLAs motivated a structurally different action head. \textbf{$\pi_0$}~\cite{black2024pi0} addresses it with the first flow-matching action head for VLAs. Built on a 3B-parameter PaLIGemma~\cite{beyer2024paligemma} backbone, the model processes image and language tokens through the VLM then uses the resulting hidden states to condition a flow-matching network that generates action chunks. The action head consists of Transformer layers that jointly attend to VLM features and noisy action tokens, iteratively denoising over $K = 10$ flow steps.

$\pi_0$ achieved state-of-the-art results on bimanual dexterous tasks including laundry folding (80\% success), box assembly, and table bussing. The flow-matching formulation is critical for bimanual tasks: it generates smooth, coherent 16-dimensional action chunks (8~per arm) without the quantization artifacts of autoregressive methods or the slow sampling of diffusion models. Pre-training on OXE followed by task-specific fine-tuning proved essential; the pre-trained model provides a strong initialization that allows learning from relatively few bimanual demonstrations. A significant limitation is that $\pi_0$'s strongest results depend on proprietary multi-task data collected across Physical Intelligence's robot fleet; reproducing these results with publicly available data alone has not been demonstrated, raising questions about how much of the performance stems from architecture versus data advantage.

\subsubsection{$\pi_{0.5}$}

\textbf{$\pi_{0.5}$}~\cite{black2025pi05} extends $\pi_0$ to open-world deployment by introducing a hierarchical architecture. A high-level VLM generates subgoal language commands, while the low-level flow-matching policy executes motor actions. This decomposition allows $\pi_{0.5}$ to handle complex, multi-step household tasks such as ``clean the kitchen'' that require planning over minutes rather than seconds.

$\pi_{0.5}$ was deployed on a fleet of mobile manipulators in real homes, achieving high success rates at following verbal instructions across a range of household tasks, a notable demonstration of VLA generalization beyond the lab, though the results are self-reported under conditions chosen by the authors and have not been independently reproduced. The hierarchical architecture proves well suited to bimanual tasks, where the high-level model can decompose complex instructions into single-step bimanual primitives.

\subsubsection{$\pi_0^*$ and RECAP}

\textbf{$\pi_0^*$} (also referred to as $\pi_{0.6}^*$ in the original publication)~\cite{amin2025pi06} addresses a fundamental limitation of imitation learning: performance is bounded by the quality of demonstrations. $\pi_0^*$ introduces RECAP (Reinforcement Learning from Autonomous CAPability), a training approach in which the VLA collects experience autonomously and then improves via reinforcement learning. A VLM-based evaluator provides success/failure labels, eliminating hand-designed reward functions. Starting from a $\pi_0$ checkpoint, RECAP alternates between autonomous data collection and policy optimization, progressively improving beyond the demonstration distribution. On bimanual tasks, $\pi_0^*$ achieved 10--40\% absolute improvement over the demonstration-only baseline, demonstrating that VLAs can bootstrap their own improvement.

\subsection{Diffusion-Based VLAs}
\label{sec:architectures:diff}

Diffusion-based approaches apply denoising diffusion probabilistic models to action generation. The key tradeoff is generation quality versus inference latency.

\subsubsection{Diffusion Policy}

The idea of treating action generation as a denoising process originated with \textbf{Diffusion Policy}~\cite{chi2023diffusion}, which established that generative models outperform deterministic behavioral cloning for tasks with multi-modal demonstrations. The principal drawback is computational cost: $K = 50$--$100$ DDPM steps push latency to $\sim$300\,ms per chunk, making real-time bimanual control impractical without acceleration (DDIM, consistency distillation). While Diffusion Policy predates VLAs (it uses task-specific encoders), its innovations (denoising action chunks, classifier-free guidance, temporal ensembling) are reused across subsequent VLA architectures.

\subsubsection{RDT-1B}

A natural question is whether scaling alone can close the gap between diffusion and flow-based VLAs. \textbf{RDT-1B}~\cite{liu2024rdt1b} tests this hypothesis by pushing diffusion-based action generation to 1.2 billion parameters, creating a ``diffusion foundation model'' for bimanual manipulation. The model uses a Transformer backbone (inspired by DiT) to denoise action chunks, conditioned on visual features from a pre-trained vision encoder and language features from a pre-trained text encoder. Pre-trained on multi-robot datasets and fine-tuned on ALOHA bimanual tasks, RDT-1B confirms that scale benefits diffusion-based robot policies just as it benefits language models.

RDT-1B's large parameter count allows it to capture the complex coordination patterns required for bimanual tasks. On ALOHA benchmarks, it outperformed ACT and Diffusion Policy on tasks requiring tight bimanual coordination such as handovers and collaborative assembly. The model also exhibited improved robustness to visual distractors and perturbations compared to smaller diffusion policies.

RDT-1B uses separate vision and language encoders (SigLIP and T5) rather than a unified VLM, reaching competitive performance through scale rather than joint pre-training. The DiT backbone handles long bimanual sequences (64 timesteps $\times$ 16 dimensions = 1024 elements) efficiently.

\subsubsection{CogACT}

A key tension in diffusion VLAs is the conflict between language generation and action denoising losses. \textbf{CogACT}~\cite{li2025cogact} resolves this by introducing learned ``cognitive action tokens'' that bridge VLM semantic representations and the diffusion action head. This abstraction layer isolates the two objectives, reducing their interference during training.

Related approaches include \textbf{PerAct}~\cite{shridhar2023perceiver} (3D voxel-based manipulation), \textbf{RVT}~\cite{goyal2023rvt} (efficient multi-view 3D manipulation), \textbf{3D Diffusion Policy}~\cite{ze20243d} (diffusion over point clouds), and \textbf{Transfusion}~\cite{zhou2024transfusion} (unified next-token prediction with diffusion generation).

\subsection{Hybrid and Efficient VLAs}
\label{sec:architectures:hybrid}

Several recent architectures combine multiple action generation approaches or focus on computational efficiency.

\subsubsection{HybridVLA}

Autoregressive and diffusion-based approaches have complementary strengths: autoregressive models excel at discrete decisions (\eg, grasp vs. release) while diffusion models excel at continuous trajectories. \textbf{HybridVLA}~\cite{chen2025hybridvla} exploits this complementarity by integrating both within a unified architecture, routing discrete action components through an autoregressive head and continuous components through a diffusion head. This hybrid design outperforms either approach alone on tasks with mixed discrete--continuous action~spaces.

The hybrid architecture suits bimanual manipulation well: gripper commands (open/close) are inherently discrete while arm motions are continuous. HybridVLA routes each component to the appropriate head, with both conditioning on the same VLM hidden states. A limitation is added training complexity: balancing the two heads requires careful loss weighting, and no published ablation isolates each head's independent contribution.

\subsubsection{TinyVLA}

Inference latency remains the primary obstacle to deploying large VLAs on real-time bimanual systems. \textbf{TinyVLA}~\cite{wen2024tinyvla} tackles this by pairing a compact 1B-parameter VLM backbone with a diffusion policy action head, achieving a small-footprint architecture that avoids the overhead of larger VLM-based systems. The result is real-time control at $50\,\text{Hz}$ on consumer GPUs, making VLA deployment practical for bimanual systems with tight latency requirements. The compact architecture introduces a capacity ceiling; however, the smaller VLM backbone limits semantic reasoning compared to larger VLAs, and the fidelity of action generation may degrade for high-dimensional bimanual action distributions.

\subsubsection{MiniVLA}

A contrasting strategy to distillation is to design for efficiency from the outset. \textbf{MiniVLA}~\cite{belkhale2024minivla} pairs a small VLM backbone with a lightweight action head, attaining competitive performance on standard benchmarks at a fraction of the compute. This ground-up efficiency could benefit bimanual systems, though MiniVLA has not been evaluated on such tasks. MiniVLA's reduced capacity limits its ability to handle complex multi-step reasoning.

\subsubsection{FAST}

Efficient autoregressive action generation requires a better tokenization scheme than naive per-dimension binning. \textbf{FAST}~\cite{pertsch2025fast} provides one with a learned action tokenizer that compresses continuous action chunks into discrete tokens. A VQ-VAE first learns to tokenize action chunks, and then a VLM is fine-tuned to predict these tokens. The tokenizer captures robot-specific action structure (temporal correlations, joint coupling), producing a compact discrete representation that avoids the quantization artifacts of naive binning. FAST achieves state-of-the-art results among autoregressive VLAs while maintaining the simplicity of text-token generation.

Applied to dual-arm control, the FAST tokenizer learns coordinated patterns as single codebook entries, compressing a 50-step chunk ($50 \times 16 = 800$ values) into 16--32 tokens \mbox{(25--50$\times$ compression),} which reduces latency and improves long-range dependency modeling. A limitation is that the VQ-VAE codebook is fixed after training, so novel action patterns outside the codebook's coverage may be poorly represented.

\textbf{Hi Robot}~\cite{shi2025hirobot} takes the hierarchical route: a high-level VLM selects subtasks while a low-level flow policy executes them, enabling open-ended instruction following on bimanual platforms.
% ============================================================
% TABLE 2: VLA ARCHITECTURES
% ============================================================
\begin{table}[H]
\centering
\caption{Comparison %
 of VLA architectures. Action type indicates the generation mechanism: AR = autoregressive, FM = flow matching, Diff = diffusion, Hybrid = combined. Bimanual indicates demonstrated bimanual capability. Params refers to the total model size. Visual encoder and training data columns provide additional context on each system's design. Init. indicates whether the visual encoder is fine-tuned (FT) from a pre-trained checkpoint or trained from scratch (Scr.).}
\label{tab:architectures}
\fontsize{6.15}{6.15}\selectfont

\begin{adjustwidth}{-\extralength}{0cm}
%\centering %% If there is a figure in wide page, please release command \centering

\begin{tabularx}{\fulllength}{llcccccccccl}
\toprule
\textbf{Method} & \textbf{Action Type} & \textbf{VLM Backbone} & \textbf{Visual Enc.} & \textbf{Init.} & \textbf{Params} & \textbf{Chunk $\textbf{\emph{H}}$} & \textbf{Pre-Train Data} & \textbf{Bimanual} & \textbf{Open-Source} & \textbf{Year} \\
\midrule
\textbf{RT-1 %
}~\cite{brohan2022rt1} & AR (discrete) & EfficientNet & EfficientNet-B3 & FT (ImageNet) & 35M %
 & 1 & 130K demos & -- & -- & 2022 \\
\textbf{RT-2}~\cite{brohan2023rt2} & AR (discrete) & PaLI-X/PaLM-E & ViT-22B & FT (web) & 55B & 1 & RT-1 data & -- & -- & 2023 \\
\textbf{Octo}~\cite{octo2024} & Diff head & Custom & Custom ViT & Scr. & 93M & 4 & OXE (800K) & \checkmark & \checkmark & 2024 \\
\textbf{OpenVLA}~\cite{kim2024openvla} & AR (discrete) & Prismatic & DINOv2 + %
 SigLIP & FT (web) & 7B & 1 & OXE & -- & \checkmark & 2024 \\
\textbf{$\pi_0$}~\cite{black2024pi0} & FM & PaLIGemma & SigLIP & FT (web) & 3B & 50 & OXE + propri. & \checkmark & -- & 2024 \\
\textbf{$\pi_{0.5}$}~\cite{black2025pi05} & FM (hierarchical) & PaLIGemma & SigLIP & FT (web) & 3B & 50 & OXE + propri. & \checkmark & -- & 2025 \\
\textbf{$\pi_0^*$}~\cite{amin2025pi06} & FM + RL & PaLIGemma & SigLIP & FT (web) & 3B & 50 & OXE + RL data & \checkmark & -- & 2025 \\
\textbf{RDT-1B}~\cite{liu2024rdt1b} & Diff (DiT) & SigLIP + T5 & SigLIP & FT (web) & 1.2B & 64 & 46 datasets (1M+) & \checkmark & \checkmark & 2024 \\
\textbf{CogACT}~\cite{li2025cogact} & Diff head & Prismatic & DINOv2 + SigLIP & FT (web) & 7B & 16 & OXE & -- & \checkmark & 2025 \\
\textbf{HybridVLA}~\cite{chen2025hybridvla} & AR + Diff & Prismatic & DINOv2 + SigLIP & FT (web) & 7B & 50 & OXE + DROID & \checkmark & -- & 2025 \\
\textbf{TinyVLA}~\cite{wen2024tinyvla} & Diff head & LLaVA & CLIP ViT & FT (web) & 1B & 10 & LLaVA + robot & -- & \checkmark & 2024 \\
\textbf{MiniVLA}~\cite{belkhale2024minivla} & Efficient head & Prismatic & DINOv2 + SigLIP & FT (web) & 1B & 8 & OXE subset & -- & \checkmark & 2024 \\
\textbf{FAST}~\cite{pertsch2025fast} & AR (learned tok.) & Prismatic & DINOv2 + SigLIP & FT (web) & 7B & 50 & OXE & -- & \checkmark & 2025 \\
\textbf{Hi Robot}~\cite{shi2025hirobot} & Hierarchical FM & PaLIGemma & SigLIP & FT (web) & 3B+ & 50 & OXE + propri. & \checkmark & -- & 2025 \\
\bottomrule
\end{tabularx}\end{adjustwidth}
\end{table}

Table~\ref{tab:architectures} reveals several structural trade-offs across the four VLA families.

\emph{Autoregressive} VLAs (RT-1, RT-2, OpenVLA, FAST) inherit the simplicity and scalability of language modeling, but discretizing continuous actions into bins introduces quantization artifacts that degrade fine-grained bimanual control; only FAST's learned tokenizer alleviates this without abandoning the autoregressive paradigm.

\emph{Flow-based} VLAs ($\pi_0$, $\pi_{0.5}$, $\pi_0^*$) generate smooth, continuous action chunks and currently achieve the best bimanual performance but require multiple forward passes through the flow head ($K$ {=} 10 denoising steps), coupling model size to inference cost.

\emph{Diffusion-based} VLAs (Octo, RDT-1B, CogACT) excel at modeling multi-modal action distributions---critical when multiple valid trajectories exist---but their 50--100 denoising steps incur 2--4$\times$ higher latency than flow matching.

\emph{Hybrid} approaches (HybridVLA, Hi Robot) combine discrete high-level reasoning with continuous low-level execution, achieving strong benchmark performance at the cost of increased architectural complexity and harder end-to-end training.

A notable trend is parameter efficiency: the best-performing architecture ($\pi_0$, 3B parameters) is an order of magnitude smaller than the earliest VLA (RT-2, 55B%
), suggesting that action head design and training strategy matter more than raw model scale. Looking at the \textbf{Params} column of Table~\ref{tab:architectures} together with the benchmark scores in Table~\ref{tab:benchmark_performance} reinforces this point: efficient architectures such as \textbf{TinyVLA}~\cite{wen2024tinyvla} (1B) and \textbf{MiniVLA}~\cite{belkhale2024minivla} (300M--1B) retain roughly 85--90\% of the success rate of multi-billion-parameter flow-based VLAs on LIBERO and SIMPLER while reducing inference cost by 3--10$\times$. In other words, the \emph{Params} entry in our tables is best read not as a proxy for capability but as one axis of a latency--quality--capacity tradeoff that is critical for real-time bimanual and aerial deployment.

% ============================================================
% FIGURE 3: ARCHITECTURE COMPARISON
% ============================================================

% ============================================================
% TABLE: ARCHITECTURE PERFORMANCE ON BENCHMARKS
% ============================================================
\begin{table}[H]
%\centering
\caption{VLA performance on standard benchmarks. LIBERO scores are normalized success rates averaged across task suites. Bridge V2 reports average success rate across evaluation tasks. SIMPLER reports success rate on the Google Robot evaluation suite. Bold indicates best in column {(values %
 are reported from the original publications under varying evaluation conditions and should be interpreted with caution; not all methods were evaluated on all benchmarks).}}
\label{tab:benchmark_performance}

\begin{adjustwidth}{-\extralength}{0cm}
%\centering %% If there is a figure in wide page, please release command \centering

%
\begin{tabularx}{\fulllength}{lcccccc}
\toprule
\textbf{Method} & \textbf{LIBERO-Spatial} & \textbf{LIBERO-Object} & \textbf{LIBERO-Goal} & \textbf{LIBERO-Long} & \textbf{Bridge V2} & \textbf{SIMPLER} \\
\midrule
\textbf{RT-1 %
}~\cite{brohan2022rt1} & -- & -- & -- & -- & 45.2 & -- \\
\textbf{RT-2}~\cite{brohan2023rt2} & -- & -- & -- & -- & 52.1 & 55.3 \\
\textbf{Octo}~\cite{octo2024} & 78.9 & 85.7 & 72.1 & 46.3 & 54.8 & 48.7 \\
\textbf{OpenVLA}~\cite{kim2024openvla} & 84.7 & 88.4 & 79.2 & 53.7 & 58.3 & 56.2 \\
\textbf{$\pi_0$}~\cite{black2024pi0} & \textbf{96.2} & \textbf{97.1} & \textbf{93.5} & \textbf{78.4} & \textbf{72.6} & \textbf{71.3} \\
\textbf{RDT-1B}~\cite{liu2024rdt1b} & 89.3 & 91.5 & 84.7 & 62.1 & 61.4 & -- \\
\textbf{CogACT}~\cite{li2025cogact} & 87.1 & 89.8 & 81.3 & 58.9 & 59.7 & -- \\
\textbf{FAST}~\cite{pertsch2025fast} & 91.5 & 93.2 & 86.8 & 65.3 & 64.2 & 62.8 \\
\textbf{HybridVLA}~\cite{chen2025hybridvla} & 93.8 & 95.4 & 90.1 & 72.6 & 68.9 & 67.4 \\
\textbf{TinyVLA}~\cite{wen2024tinyvla} & 82.3 & 86.1 & 75.4 & 51.2 & 55.6 & 52.1 \\
\bottomrule
\end{tabularx}
\end{adjustwidth}
\end{table}

Table~\ref{tab:benchmark_performance} compares VLA methods across standard benchmarks. Three findings stand out from this comparison. First, flow-based VLAs (\textbf{$\pi_0$}) dominate across all benchmarks, with notably large margins on long-horizon tasks (LIBERO-Long) where action chunk coherence matters most. Second, hybrid approaches (\textbf{HybridVLA}) rank second, suggesting that combining autoregressive and flow-based generation captures complementary strengths. Third, efficient models (\textbf{TinyVLA}) trade 10--15\% performance for significantly reduced compute, a worthwhile tradeoff for resource-constrained deployments.

Autoregressive VLAs benefit most from VLM pre-training but struggle with high-dimensional continuous actions. Flow-based VLAs offer the best quality--speed balance for bimanual tasks. Diffusion-based VLAs provide strong multi-modal modeling at higher cost. 
Architecture alone does not determine performance; as Section~\ref{sec:training} will show, training strategy---particularly the choice of pre-training data, co-training ratio, and RL fine-tuning---is equally decisive in closing the gap between architectural promise and deployed capability. Two modular extensions apply across all families: \emph{memory modules}~\cite{torne2026mem,shi2025memoryvla,jang2025contextvla} for long-horizon task tracking (Section~\ref{sec:bimanual:longhorizon}) and \emph{world models}~\cite{gigabrain0_2025,rhoda2026dva,worldvla2025} for future state prediction (Section~\ref{sec:crosscutting:worldmodels}).

% ============================================================
% 6. TRAINING RECIPES AND DATA STRATEGIES
% ============================================================
\section{Training Recipes and Data Strategies}
\label{sec:training}

The performance of a VLA depends as much on how it is trained as on its architecture. This section reviews the three-stage training pipeline (pre-training, post-training, and reinforcement learning) as well as data collection strategies for bimanual manipulation. Section~\ref{sec:crosscutting:scalability} %
 illustrates the complete pipeline from VLM pre-training through deployment.

\subsection{Pre-Training}
\label{sec:training:pretrain}

As outlined in the ``Training'' branch of Figure~\ref{fig:taxonomy}, VLA pre-training proceeds in two phases. First, the VLM backbone is pre-trained on internet-scale image-text data, learning general visual and semantic representations. Second, the full VLA (backbone plus action head) is trained on large-scale robot demonstration data.

The VLM pre-training phase builds on existing VLM checkpoints. \textbf{$\pi_0$}~\cite{black2024pi0} initializes from PaLIGemma~\cite{beyer2024paligemma}, a 3B-parameter VLM pre-trained on web-scale image-text data using the SigLIP visual encoder and Gemma~\cite{team2024gemma} language backbone. \textbf{OpenVLA}~\cite{kim2024openvla} initializes from Prismatic, a 7B-parameter VLM combining DINOv2 and SigLIP visual encoders with a Llama-2 language backbone. \textbf{RT-2}~\cite{brohan2023rt2} initializes from PaLI-X, a 55B-parameter VLM. This initialization provides the VLA with broad visual understanding, language comprehension, and spatial reasoning capabilities before it encounters any robot data.

Taken together, the trend is toward smaller, efficient VLM backbones paired with powerful action heads: larger VLMs encode richer representations but are too slow for real-time control, while 3B-parameter models like PaLIGemma allow flow-matching heads with multiple forward passes per chunk.

The robot data pre-training phase then trains the VLA on diverse cross-embodiment data. \textbf{$\pi_0$}~\cite{black2024pi0} pre-trains on a mixture of OXE data and proprietary multi-task data spanning seven robot embodiments and hundreds of tasks, using a co-training objective that mixes action prediction with VLM text generation to preserve language capabilities. The co-training loss is:
\begin{linenomath}
\begin{equation}
    \mathcal{L}_{\text{co-train}} = \lambda_{\text{act}} \mathcal{L}_{\text{FM}} + \lambda_{\text{lang}} \mathcal{L}_{\text{LM}},
    \label{eq:cotrain}
\end{equation}
\end{linenomath}
where $\mathcal{L}_{\text{FM}}$ is the flow matching loss (Equation~(\ref{eq:flow_loss})), $\mathcal{L}_{\text{LM}}$ is the language modeling loss, and $\lambda_{\text{act}}, \lambda_{\text{lang}}$ are balancing weights. \textbf{OpenVLA}~\cite{kim2024openvla} pre-trains exclusively on OXE data, demonstrating that publicly available data suffices for competitive pre-training. The key finding across all approaches is that data diversity (spanning multiple embodiments, environments, and tasks) matters more than dataset size for downstream generalization. This diversity also benefits memory-augmented VLAs~\cite{torne2026mem,shi2025memoryvla}: pre-training on varied multi-step tasks exposes the memory module to diverse temporal patterns, improving its ability to track long-horizon bimanual task state. Additionally, world models can serve as \emph{data engines} during pre-training; \textbf{GigaWorld-0}~\cite{gigaworld0_2025} generates synthetic robot episodes via video generation and sim-to-real transfer, augmenting real demonstration data at scale. However, no published study has rigorously ablated whether pre-training gains stem from data diversity or backbone quality, making it difficult to attribute improvements to either factor~alone.

\subsection{Post-Training and Fine-Tuning}
\label{sec:training:posttrain}

Post-training adapts a pre-trained VLA to specific tasks, embodiments, or environments. Fine-tuning on task-specific demonstrations is the most common approach. \textbf{$\pi_0$}~\cite{black2024pi0} fine-tunes on 50--200 bimanual demonstrations per task, which yields strong performance from a well-initialized model. \textbf{OpenVLA}~\cite{kim2024openvla} demonstrated that fine-tuning on as few as 10 demonstrations can yield significant improvements on in-distribution tasks.

Co-training (mixing target-task data with pre-training data during fine-tuning) prevents catastrophic forgetting and often improves performance on the target task. \textbf{$\pi_0$}~\cite{black2024pi0} mixes cross-embodiment data with single-task data throughout fine-tuning, finding that this consistently outperforms fine-tuning on target data alone. The intuition is that cross-embodiment data provides a regularization effect, preventing the model from overfitting to the small fine-tuning dataset.

The mixing ratio is a key hyperparameter: \textbf{$\pi_0$}~\cite{black2024pi0} uses a 1:1 ratio, while \textbf{Mobile ALOHA}~\cite{fu2024mobilealoha} found even 10\% diverse co-training data improves bimanual success. Tasks dissimilar to pre-training benefit from more target data; visually similar tasks benefit from more diverse co-training.

Parameter-efficient fine-tuning via Low-Rank Adaptation (LoRA)~\cite{hu2022lora} offers an alternative to full fine-tuning: by injecting low-rank weight updates into frozen VLM layers, LoRA reduces GPU memory requirements while preserving pre-trained representations. Preference-based optimization methods such as DPO~\cite{rafailov2023dpo} have also been explored for aligning VLA outputs with human preferences, though their application to bimanual manipulation remains limited.

Other techniques aim to preserve pre-trained knowledge during fine-tuning. \textbf{Knowledge Insulation}~\cite{driess2024ki} freezes certain VLM layers during fine-tuning to preserve language understanding. \textbf{Align-then-Steer}~\cite{xiao2025aligning} uses two-phase alignment and constrained fine-tuning. \textbf{RoboMimic}~\cite{mandlekar2023robomimic} finds that observation representations and data quality have outsized impact on offline IL, and \textbf{RoboAgent}~\cite{mandlekar2023roboagent} improves generalization through semantic augmentations.

\subsection{Reinforcement Learning for VLAs}
\label{sec:training:rl}

Imitation learning alone caps VLA performance at the demonstration quality. Reinforcement learning (RL) offers a path to surpass this ceiling. Early large-scale RL for robotic manipulation (\textbf{QT-Opt}~\cite{kalashnikov2018qtopt} trained grasping policies from over 580,000 real grasps) showed that RL can scale in the real world. Offline RL methods such as \textbf{Conservative Q-Learning (CQL)}~\cite{kumar2020cql} and \textbf{Advantage-Weighted Regression (AWR)}~\cite{peng2019awr} learn from static datasets without further environment interaction, providing a lower-risk alternative (see Levine~\etal~\cite{levine2020offline} for a tutorial). Policy gradient methods such as PPO~\cite{schulman2017ppo} remain the most common choice for online fine-tuning of VLAs.

Online RL for VLAs was first realized by \textbf{$\pi_0^*$ (RECAP)}~\cite{amin2025pi06}. The RECAP pipeline works as follows: (1) the VLA policy autonomously executes tasks in the real world, (2) a VLM-based evaluator labels each episode as success or failure, (3) successful episodes are added to the training set and failures are discarded, and (4) the VLA is fine-tuned on the augmented dataset. This cycle repeats, gradually expanding the VLA's competence beyond the original demonstration distribution.

RECAP's key innovation is using a separate VLM as an autonomous reward function, eliminating the need for hand-designed reward signals. On bimanual tasks, RECAP improved laundry folding success from 60\% (demonstration-only) to over 90\% after several cycles of autonomous practice. This demonstrates that VLAs can self-improve on bimanual tasks, a necessary capability for scaling deployment. Algorithm~\ref{alg:recap} summarizes the RECAP~pipeline.

\begin{algorithm}[H]
\caption{RECAP: Reinforcement Learning from Autonomous Capability.}
\label{alg:recap}
\begin{algorithmic}[1]
\REQUIRE Pre-trained VLA policy $\pi_\theta$, VLM evaluator $\mathcal{V}$, task set $\mathcal{T}$
\REQUIRE Demonstration buffer $\mathcal{D}_{\text{demo}}$, practice buffer $\mathcal{D}_{\text{auto}} \leftarrow \emptyset$
\FOR{cycle $c = 1, 2, \ldots, C$}
    \STATE \textit{// Autonomous data collection}
    \FOR{task $\tau \in \mathcal{T}$}
        \STATE Execute $\pi_\theta$ on $\tau$, collect trajectory $\xi$
        \STATE Evaluate: $r \leftarrow \mathcal{V}(\xi, \tau)$ \hfill \textit{// VLM judges success}
        \IF{$r = \text{success}$}
            \STATE $\mathcal{D}_{\text{auto}} \leftarrow \mathcal{D}_{\text{auto}} \cup \{\xi\}$
        \ENDIF
    \ENDFOR
    \STATE \textit{// Policy improvement}
    \STATE Fine-tune $\pi_\theta$ on $\mathcal{D}_{\text{demo}} \cup \mathcal{D}_{\text{auto}}$
\ENDFOR
\RETURN Improved policy $\pi_\theta$
\end{algorithmic}
\end{algorithm}

While RECAP targets success rate, RL can also optimize for speed: human teleoperation of bimanual systems is typically slow and cautious, so policies that merely imitate inherit this inefficiency. \textbf{SAIL}~\cite{rana2025sail} addresses this by training VLA policies to execute tasks \emph{faster} than the demonstrations. Using time-warped demonstrations and a reward that encourages speed while maintaining success, SAIL produces policies that complete bimanual tasks in less time than human demonstrators, removing the teleoperation speed bottleneck.

Beyond these targeted methods, other RL approaches include \textbf{VLA-RL}~\cite{lu2025vlarl} (scalable RL for general robotic manipulation), \textbf{ConRFT}~\cite{chen2025conrft} (consistency-policy-regularized RL), \textbf{Q-Transformer}~\cite{chebotar2023qlearning} (offline RL via autoregressive Q-functions), \textbf{DPPO}~\cite{ren2023dppo} (policy optimization for diffusion policies), and \textbf{Self-Improving Embodied Foundation Models}~\cite{ghasemipour2025selfimprove} (autonomous data generation and filtering).

Rather than learning purely from collected experience, \textbf{GigaBrain-0.5M}~\cite{gigabrain2025} integrates a world model with VLA training. The world model predicts future visual observations given current actions, allowing the VLA to ``imagine'' the consequences of action sequences without physical execution, a form of model-based RL that generates synthetic training data. Two-arm coordination demands that the world model predict complex multi-body dynamics, including how both arms and the manipulated object evolve over time. While world-model-based RL for bimanual VLAs is still early-stage, it represents a promising path toward sample-efficient training of complex coordination behaviors.

\subsection{Data Collection for Bimanual Manipulation}
\label{sec:training:data}

High-quality bimanual demonstration data is the bottleneck for VLA training. Three data collection strategies have emerged.

\textbf{ALOHA}~\cite{zhao2023aloha} uses bilateral teleoperation: a human operator controls follower arms by physically moving kinematically identical leader arms. This provides intuitive, low-latency control for dexterous bimanual tasks. The ALOHA hardware costs under USD 20,000 and has been replicated at dozens of institutions, creating a growing ecosystem of bimanual~data.

\textbf{UMI}~\cite{chi2024umi} decouples data collection from the robot entirely. Operators use hand-held gripper tools with visual--inertial tracking to demonstrate tasks in any environment. The collected trajectories are retargeted to the robot's action space during training. UMI allows data collection by non-experts in diverse settings, increasing data diversity.

\textbf{Autonomous data collection}, as in RECAP~\cite{amin2025pi06}, uses the VLA itself to collect additional data. Starting from a reasonably capable policy, the robot attempts tasks autonomously, and a success classifier (often another VLM) labels the outcomes. This approach can generate thousands of additional episodes with minimal human effort, and the data naturally covers the policy's distribution, reducing the train--test mismatch that plagues behavioral cloning.

\textbf{Data quality} matters as much as quantity: teleoperation quality varies across operators, language labels are often inconsistent, and demonstrations including recovery from near-failures contribute disproportionately to robustness. Fleet-based data strategies~\cite{levine2018grasping} established the precedent for VLA-scale collection, and \textbf{GigaBrain-0.5M}~\cite{gigabrain2025} addresses quality through automated world-model-based filtering.

\textls[-15]{Data collection throughput varies: ALOHA teleoperation~\cite{zhao2023aloha} yields 50--100 demos/h, %
 UMI~\cite{chi2024umi} achieves $\sim$110 demos/h with non-expert operators ($3\times$ faster than spacemouse), and autonomous RECAP~\cite{amin2025pi06} runs continuously without supervision (\mbox{4--12} episodes/h for complex tasks). These economics favor a hybrid strategy: bootstrap with demonstrations, then scale through autonomous practice.}

% ============================================================
% TABLE 3: TRAINING STRATEGIES
% ============================================================

\subsection{Data Scaling Laws}
\label{sec:training:scaling}

VLA performance scales differently with different data types. Language-guided data generation~\cite{ha2023scaling} has shown that automatically generated language annotations can significantly expand effective training set size. \textbf{Cross-embodiment pre-training data} exhibits log-linear scaling, with dataset diversity (number of embodiments, environments, tasks) at least as important as raw volume~\cite{kim2024openvla}. \textbf{Task-specific fine-tuning data} shows steep initial scaling and diminishing returns: simple tasks require as little as 5 h, while complex bimanual tasks need 100+ hours~\cite{black2024pi0}. \textbf{Autonomous practice data} from RECAP~\cite{amin2025pi06} yields large gains from modest data: $\sim$300 autonomous trajectories per iteration improved laundry folding from $\sim$30\% to over 90\%.

Table~\ref{tab:training} summarizes the training strategies across representative VLA methods. The pipeline is converging on a three-stage recipe: (1) initialize from a strong VLM, (2) pre-train on diverse cross-embodiment data with co-training, and (3) fine-tune on task-specific bimanual demonstrations, optionally followed by RL.

\begin{table}[H]
%\centering
\caption{Comparison %
 of training strategies for VLA models. Pre-train data and fine-tune data indicate the primary datasets used. RL indicates whether reinforcement learning is incorporated.}
\label{tab:training}
\fontsize{9.6}{9.6}\selectfont

\begin{adjustwidth}{-\extralength}{0cm}
%\centering %% If there is a figure in wide page, please release command \centering

\begin{tabularx}{\fulllength}{llllccl}
\toprule
\textbf{Method} & \textbf{VLM Init} & \textbf{Pre-Train Data} & \textbf{Fine-Tune Data} & \textbf{Co-Train} & \textbf{RL} & \textbf{Key Strategy} \\
\midrule
\textbf{RT-2 %
}~\cite{brohan2023rt2} & PaLI-X & Google fleet & -- & -- & -- & VLM co-fine-tuning \\
\textbf{OpenVLA}~\cite{kim2024openvla} & Prismatic & OXE & Task-specific & -- & -- & Open data pre-train \\
\textbf{Octo}~\cite{octo2024} & From scratch & OXE (800K) & Task-specific & -- & -- & Cross-embodiment init \\
\textbf{$\pi_0$}~\cite{black2024pi0} & PaLIGemma & OXE + proprietary & 50--200 demos & \checkmark & -- & Co-training mix \\
\textbf{$\pi_{0.5}$}~\cite{black2025pi05} & PaLIGemma & OXE + fleet & Fleet demos & \checkmark & -- & Hierarchical training \\
\textbf{$\pi_0^*$}~\cite{amin2025pi06} & $\pi_0$ ckpt & Same as $\pi_0$ & Autonomous + demos & \checkmark & \checkmark & RECAP (VLM reward) \\
\textbf{RDT-1B}~\cite{liu2024rdt1b} & SigLIP + %
 T5 & Multi-robot & ALOHA tasks & -- & -- & Scale (1.2B params) \\
\textbf{FAST}~\cite{pertsch2025fast} & VLM & OXE & Task-specific & -- & -- & Learned tokenizer \\
\textbf{GigaBrain}~\cite{gigabrain2025} & VLM & GigaBrain-0.5M & Task-specific & -- & \checkmark & World-model RL \\
\bottomrule
\end{tabularx}\end{adjustwidth}
\end{table}

\textbf{Key takeaways from the training section.} Three points deserve to be carried into the rest of the review. First, \emph{data diversity} (across embodiments, scenes, and tasks) is a more reliable lever for generalization than raw dataset size and is now the de facto requirement for any VLA targeting cross-embodiment deployment. Second, \emph{co-training} that mixes manipulation, mobile, and vision--language data within a single optimization, rather than chaining them sequentially, is what allows VLAs to retain semantic capability while learning low-level control. Third, \emph{reinforcement learning from autonomous practice}---in particular the RECAP-style loop demonstrated by $\pi_0^*$---is the principal mechanism that lifts VLAs above their teleoperated demonstration ceiling and, together with the first two points, accounts for most of the recent progress on long-horizon, contact-rich tasks. How actions are represented and executed within these pipelines is equally consequential for real-time bimanual control, which we address next.

% ============================================================
% 7. ACTION REPRESENTATIONS AND REAL-TIME EXECUTION
% ============================================================
\section{Action Representations and Real-Time Execution}
\label{sec:action}

The choice of action representation directly shapes VLA performance, especially for bimanual tasks where the action space is high-dimensional and temporal coordination is critical. We emphasize that action representation is not merely an output-format choice but the principal mechanism by which a VLA satisfies---or fails to satisfy---real-time control constraints: each candidate representation imposes its own latency profile (per-dimension token generation in autoregressive heads, multi-step denoising in diffusion/flow heads, single-pass continuous prediction in distilled heads), and this latency directly determines whether the policy can close the control loop at the 50\,Hz rates required for bimanual manipulation or the ${\geq}100\,$Hz rates required for stable flight. Building on the architectural foundations in Section~\ref{sec:architectures} and the ``Actions'' branch of the taxonomy in Figure~\ref{fig:taxonomy}, this section covers discrete tokenization, continuous generation, action chunking strategies, and recent advances in real-time execution.

\subsection{Discrete Action Tokenization}
\label{sec:action:discrete}

The simplest approach to interfacing robot actions with a language model is to discretize continuous actions into tokens.

\textbf{Uniform binning}, used by \textbf{RT-2}~\cite{brohan2023rt2} and \textbf{OpenVLA}~\cite{kim2024openvla}, divides each action dimension into $B$ equally spaced bins (typically $B = 256$). A 7-DOF arm action is represented as 7~tokens, each from a vocabulary of size $B$. This approach is simple but introduces quantization error proportional to $1/B$ and requires sequential token generation, with latency scaling linearly with the number of dimensions. For bimanual systems with $d_a = 16$ (see Section~\ref{sec:problem:bimanual}), generating 16 tokens sequentially becomes a latency bottleneck.

The inefficiency of uniform binning motivated a learned alternative. In \textbf{FAST}~\cite{pertsch2025fast}, a VQ-VAE is trained to compress action chunks into a small number of discrete tokens. Because the tokenizer learns the structure of robot actions (temporal smoothness, joint correlations, coordination patterns), it can represent a 50-step, 16-dimensional bimanual action chunk with as few as 32 tokens, compared to $50 \times 16 = 800$ tokens with per-dimension binning. The learned tokenizer also produces a denser token vocabulary where every token corresponds to a meaningful action pattern, unlike uniform binning where most of the 256 bins are rarely used.

Cross-embodiment transfer demands that action tokens generalize across morphologies. \textbf{Universal action tokenization}~\cite{zheng2025universal} achieves this with a shared tokenizer that learns embodiment-agnostic action representations, so that a single autoregressive VLA can generate actions for robots with different morphologies. For bimanual systems, universal tokenization offers the prospect of transferring manipulation knowledge from single-arm datasets to dual-arm systems, since the tokenizer can learn correspondences between single-arm and bimanual action patterns.

Complementary approaches include \textbf{consistency models}~\cite{ke2024ccil} (distilling multi-step diffusion into a single forward pass), \textbf{ACT}~\cite{zhao2023aloha} (improved temporal ensembling for bimanual trajectories), and \textbf{RACER}~\cite{dai2024racer} (language-guided corrective actions for error~recovery).

Discrete tokenization preserves compatibility with VLM text generation but introduces quantization error that compounds across 16 bimanual action dimensions and 50 timesteps per chunk. Continuous generation avoids this issue but requires iterative denoising.

\subsection{Continuous Action Generation}
\label{sec:action:continuous}

Continuous action generation avoids discretization entirely, predicting real-valued action vectors.

Flow matching (\textbf{$\pi_0$}~\cite{black2024pi0}) and diffusion (\textbf{Diffusion Policy}~\cite{chi2023diffusion}, \textbf{RDT-1B}~\cite{liu2024rdt1b}), whose architectural details appear in Section~\ref{sec:architectures}, generate continuous actions through iterative denoising. The primary advantage is fidelity: continuous predictions avoid quantization error, which compounds across the 16 action dimensions of a bimanual system. The primary disadvantage is inference cost: each prediction requires $K$ denoising steps, each involving a forward pass through the action head.

The number of denoising steps $K$ trades off quality against speed. \textbf{$\pi_0$}~\cite{black2024pi0} uses $K = 10$ flow matching steps, which strikes a good balance for bimanual control at $50\,\text{Hz}$. Diffusion policies typically require $K = 50$--$100$ DDPM steps, though DDIM acceleration reduces this to $K = 10$--$20$ with modest quality loss.

\subsection{Action Chunking Strategies}
\label{sec:action:chunking}

Action chunking (Section~\ref{sec:problem:chunking}) is now standard in VLA systems. The chunk horizon $H$ is a critical hyperparameter.

Short chunks ($H = 1$--$4$) provide high reactivity (the policy can respond to environmental changes quickly) but require frequent VLM inference and suffer from myopic behavior. \textbf{Octo}~\cite{octo2024} uses $H = 4$, suitable for its lightweight architecture.

Long chunks ($H = 50$--$100$) amortize inference cost and capture long-range temporal structure but commit the robot to extended open-loop execution. \textbf{$\pi_0$}~\cite{black2024pi0} uses $H = 50$, which at $50\,\text{Hz}$ corresponds to 1 s of motion. When both arms act in concert (\eg, folding), long chunks capture the coordinated motion pattern of both arms within a single~prediction.

When a task has distinct phases (approach, grasp, manipulate, release), the chunk horizon should cover at least one complete phase. Empirically, $H = 50$ at $50\,\text{Hz}$ (1 s) is the sweet spot for most bimanual primitives.

\textbf{Temporal ensembling} blends overlapping action chunks to smooth transitions and improve robustness. Given the current chunk $\mathbf{A}_t$ and the previous chunk $\mathbf{A}_{t-s}$ (shifted by $s$ steps), the executed action is:
\begin{linenomath}
\begin{equation}
    \hat{\mathbf{a}}_t = \lambda \mathbf{a}_t^{\text{new}} + (1 - \lambda) \mathbf{a}_t^{\text{old}},
    \label{eq:temporal_ensemble}
\end{equation}
\end{linenomath}
where $\mathbf{a}_t^{\text{new}}$ is the action at time $t$ from the most recently predicted chunk $\mathbf{A}_t$, $\mathbf{a}_t^{\text{old}}$ is the corresponding action from the previously predicted (overlapping) chunk $\mathbf{A}_{t-s}$, and $\lambda \in [0, 1]$ controls the blending weight, typically decaying exponentially over the chunk.

\subsection{Real-Time Chunking (RTC)}
\label{sec:action:rtc}

Long action chunks and fast reactions seem fundamentally at odds, but \textbf{RTC}~\cite{black2025rtc} resolves this tension by restructuring how chunks are generated and executed. Rather than computing an entire chunk and executing it open-loop, RTC interleaves computation and execution: while the current chunk is being executed, the next chunk is computed in the background. When the next chunk is ready, the policy smoothly transitions to it, regardless of where execution is in the current chunk.

RTC's key contribution is showing that with careful scheduling, a flow-matching VLA can achieve both long-horizon coherence (from large chunks) and sub-100\,ms reactivity (from overlapping computation). In dual-arm tasks, this allows the robot to respond to unexpected perturbations (\eg, an object slipping from one gripper) without waiting for the current chunk to complete.

The effective reaction time is bounded by the computation time for one chunk (typically 50--70\,ms for $\pi_0$), a significant improvement over the full chunk execution time (1000\,ms for $H = 50$ at $50\,\text{Hz}$).

\subsection{Bidirectional Decoding (BID)}
\label{sec:action:bid}

For tasks where both endpoints are well defined but the intermediate trajectory is ambiguous, \textbf{BID}~\cite{liu2024bid} generates action chunks from both ends simultaneously. Given a chunk of horizon $H$, two decoders are initialized: a forward decoder starting from the current state and a backward decoder starting from the goal state. The two decoders produce action sequences that are merged at a midpoint. This bidirectional approach is especially relevant for bimanual handover tasks, where the initial grasp (forward) and the final placement (backward) are well defined but the intermediate transfer motion admits multiple solutions. BID resolves this ambiguity by anchoring both endpoints.

\subsection{Training-Time Action Conditioning}
\label{sec:action:ttac}

A subtle source of performance degradation is the train--test mismatch in action conditioning: during training, the action head conditions on ground-truth features unavailable at test time. \textbf{TTAC}~\cite{black2025ttac} bridges this gap by conditioning the action head on its own predictions during training via a stop-gradient mechanism. On bimanual tasks, this improves success rates by 5--15\% without architectural changes, with gains strongest at high action dimensionality ($d_a = 16$).

% ============================================================
% TABLE 4: ACTION REPRESENTATIONS
% ============================================================

Table~\ref{tab:action_repr} summarizes the action representation landscape. The choice of action representation involves a three-way tradeoff between expressiveness, speed, and simplicity.

\begin{table}[H]
%\centering
\caption{Comparison of action representations in VLA models. Latency is per action-chunk inference on a single GPU. Bimanual $d_a$ indicates the action dimension for bimanual systems.}

\begin{adjustwidth}{-\extralength}{0cm}
\label{tab:action_repr}

\begin{tabularx}{\fulllength}{llcccCl}

\toprule
\textbf{Method} & \textbf{Representation} & \textbf{Chunk $\textbf{\emph{H}}$} & \textbf{Steps $\emph{K}$} & \textbf{Bimanual $\emph{d}_\emph{a}$} & \textbf{Latency} & \textbf{Key Innovation} \\
\midrule
\textbf{RT-2 %
}~\cite{brohan2023rt2} & Uniform bins (256) & 1 & -- & 8 & $\sim$200\,ms & VLM token reuse \\
\textbf{OpenVLA}~\cite{kim2024openvla} & Uniform bins (256) & 1 & -- & 8 & $\sim$150\,ms & Open-source \\
\textbf{FAST}~\cite{pertsch2025fast} & Learned VQ-VAE & 50 & -- & 16 & $\sim$80\,ms & Compressed tokens \\
\textbf{$\pi_0$}~\cite{black2024pi0} & Flow matching & 50 & 10 & 16 & $\sim$70\,ms & VLM-conditioned flow \\
\textbf{Diff.~Policy}~\cite{chi2023diffusion} & DDPM/DDIM & 16 & 50--100 & 16 & $\sim$300\,ms & Multi-modal actions \\
\textbf{RDT-1B}~\cite{liu2024rdt1b} & DiT diffusion & 64 & 20 & 16 & $\sim$150\,ms & Scale (1.2B) \\
\textbf{RTC}~\cite{black2025rtc} & Flow + overlap & 50 & 10 & 16 & $<$50\,ms * & Interleaved exec. \\
\textbf{BID}~\cite{liu2024bid} & Bidirectional & Variable & Variable & 16 & $\sim$100\,ms & Dual-end decode \\
\bottomrule
\end{tabularx}
\end{adjustwidth}
%\centering %% If there is a figure in wide page, please release command \centering
\noindent{\footnotesize{* Effective latency with overlapped computation.}}

\end{table}

Discrete tokenization (RT-2, OpenVLA) is the simplest to implement atop existing VLMs but sacrifices fine-grained control and scales poorly with action dimension. Flow matching ($\pi_0$, RTC) offers the best latency--quality balance for high-dimensional bimanual actions, generating smooth 16-dimensional chunks in ${\sim}70\,\text{ms}$ with only 10 denoising steps. Diffusion models (RDT-1B, Diffusion Policy) provide the strongest multi-modal modeling but at 2--4$\times$ higher latency due to 50--100 denoising steps. Learned tokenizers (FAST) bridge the gap by compressing continuous chunks into compact discrete representations, achieving autoregressive simplicity with chunk-level expressiveness. For bimanual control at 50\,Hz, effective latency below 20\,ms per step is critical, making overlapped execution (RTC) or long-horizon chunking essential regardless of the underlying representation.

% ============================================================
% 8. BIMANUAL MANIPULATION WITH VLAs
% ============================================================
\section{Bimanual Manipulation with VLAs}
\label{sec:bimanual}

\begin{figure}[H]
%\centering
\includegraphics[width=\textwidth]{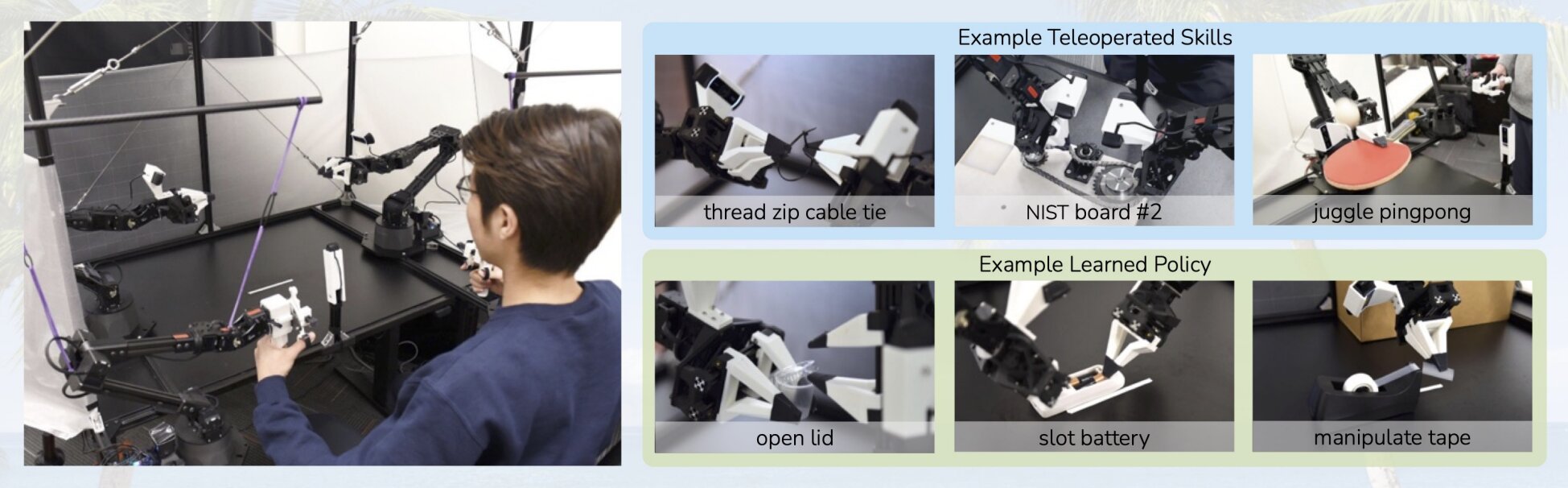}
\caption{The ALOHA bimanual teleoperation platform and representative tasks. A human operator controls two follower arms via leader arms for intuitive demonstration collection. ALOHA and its ACT policy established the standard platform for bimanual VLA research. Reprinted with permission from Ref.~\cite{zhao2023aloha}. Copyright 2023, Zhao~et~al.}
\label{fig:aloha_hardware}
\end{figure}

Bimanual manipulation is the primary application focus of this review. Prior work has studied bimanual coordination from multiple perspectives: Grannen~\etal~\cite{grannen2023stabilize} propose a ``stabilize to act'' framework where one arm stabilizes while the other manipulates, and Chitnis~\etal~\cite{chitnis2020efficient} learn task schemas for efficient bimanual planning. As shown in the ``Bimanual'' branch of Figure~\ref{fig:taxonomy}, this section examines how VLAs address the unique challenges of bimanual coordination, organized by coordination strategy (Table~ %
\ref{tab:coordination}) and task type (Table~\ref{tab:bimanual_tasks}). Table~\ref{tab:hardware} lists the principal hardware platforms. The action representations and chunking strategies discussed in Section~\ref{sec:action} are central to the coordination mechanisms described below. Figure~\ref{fig:aloha_hardware} illustrates the ALOHA platform and its bilateral \mbox{teleoperation~interface.}

% ============================================================
% FIGURE: ALOHA Bimanual Hardware
% ============================================================

% ============================================================
% TABLE: COORDINATION STRATEGIES
% ============================================================
\begin{table}[H]

\caption{Comparison %
 of bimanual coordination strategies in VLA models. $d_a$ indicates the effective action dimensionality per step. Tight coupling indicates whether the strategy supports simultaneous force-coordinated bimanual actions.}
\label{tab:coordination}

\begin{tabularx}{\textwidth}{LCCL}
\toprule
\textbf{Strategy} & \textbf{$\textbf{\emph{d}}_\textbf{\emph{a}}$} & \textbf{Coupling} & \textbf{Methods} \\
\midrule
Joint space & $d_L$ {+} $d_R$ & \checkmark & $\pi_0$, RDT-1B, ACT \\
Independent & $\max(d_L, d_R)$ & -- & Hi Robot \\
Leader--follower & $d_L$ & Partial & Custom setups \\
Hierarchical & Variable & \checkmark & $\pi_{0.5}$, Hi Robot \\
\bottomrule
\end{tabularx}
\end{table}
\vspace{-12pt}

% ============================================================
% TABLE 5: BIMANUAL TASKS
% ============================================================
\begin{table}[H]
\centering
\caption{Approximate %
 success rates (\%) on bimanual tasks, as reported in the original publications under varying evaluation conditions. Values across methods are not directly comparable due to differences in task definitions, object sets, and evaluation protocols. Tasks grouped by category.}
\label{tab:bimanual_tasks}
\begin{tabular}{llccccc}
\toprule
\textbf{Task Category} & \textbf{Specific Task} & \textbf{$\boldsymbol{\pi}_\textbf{0}$} & \textbf{$\boldsymbol{\pi}_\textbf{0}^\textbf{*}$} & \textbf{RDT-1B} & \textbf{ACT} & \textbf{Diff.~Policy} \\
\midrule
\multirow{2}{*}{\textit{Deformable %
}} & Laundry folding & 80 & 92 & -- & 50 & 35 \\
& Towel folding & 85 & 95 & 60 & 55 & 40 \\
\midrule
\multirow{2}{*}{\textit{Contact-rich}} & Box assembly & 75 & 88 & 55 & 40 & 30 \\
& Peg insertion (bimanual) & 70 & 85 & 50 & 45 & 35 \\
\midrule
\multirow{2}{*}{\textit{Long-horizon}} & Table busing & 65 & 80 & -- & 30 & -- \\
& Kitchen cleanup & 60 & 78 & -- & -- & -- \\
\midrule
\multirow{2}{*}{\textit{Coordination}} & Object handover & 90 & 95 & 75 & 70 & 60 \\
& Collaborative lift & 85 & 93 & 70 & 60 & 50 \\
\bottomrule
\end{tabular}
\end{table}
\vspace{-12pt}
% ============================================================
% TABLE 6: BIMANUAL HARDWARE
% ============================================================
\begin{table}[H]
%\centering
\caption{Bimanual %
 hardware platforms used with VLA models. DOF/arm indicates degrees of freedom per arm plus gripper. Cost is approximate at time of introduction.}
\label{tab:hardware}

\begin{tabularx}{\textwidth}{lCCC}
\toprule
\textbf{Platform} & \textbf{DOF/Arm} & \textbf{Mobile} & \textbf{Cost} \\
\midrule
\textbf{ALOHA %
}~\cite{zhao2023aloha} & 6 + 1 & -- & $<$\$20K \\
\textbf{Mobile ALOHA}~\cite{fu2024mobilealoha} & 6 + 1 & \checkmark & $<$\$30K \\
\textbf{Franka Dual} & 7 + 1 & -- & $>$\$60K \\
\textbf{UMI}~\cite{chi2024umi} & N/A * & -- & $<$\$5K \\
\bottomrule\end{tabularx}
\noindent{\footnotesize{* UMI is a data collection interface, not a robot.}}

\end{table}

\subsection{Coordination Strategies}
\label{sec:bimanual:coord}

We first formalize the bimanual coordination objective. The bimanual VLA policy $\pi_\theta^{\text{bi}}$ maps observations and proprioception from both arms to a joint action chunk:
\begin{linenomath}
\begin{equation}
    \pi_\theta^{\text{bi}} : \mathcal{O} \times \mathcal{L} \times \mathcal{Q}^L \times \mathcal{Q}^R \rightarrow \mathbb{R}^{H \times (d_L + d_R)},
    \label{eq:bimanual_policy}
\end{equation}
\end{linenomath}
where $\mathcal{Q}^L$ and $\mathcal{Q}^R$ are the proprioceptive state spaces of the left and right arms. The degree of required coordination can be characterized by the mutual information $I(\mathbf{A}_t^L ; \mathbf{A}_t^R \mid \mathbf{o}_t, \ell)$ between the two arms' action chunks. For independent tasks, this quantity is near zero; for loosely coupled tasks, it reflects timing dependencies; and for tightly coupled tasks such as folding or assembly, it is large, indicating strong inter-arm correlations. The three coordination strategies described below differ in how they handle this coupling.

\subsubsection{Joint Action Space}

The most common approach treats the bimanual system as a single high-dimensional policy. The action chunk $\mathbf{A}_t \in \mathbb{R}^{H \times (d_L + d_R)}$ encodes both arms jointly, allowing the model to learn implicit coordination. \textbf{$\pi_0$}~\cite{black2024pi0} and \textbf{RDT-1B}~\cite{liu2024rdt1b} both use this approach, predicting the left and right arm actions as a concatenated vector. The advantage is simplicity: no explicit coordination mechanism is needed, and the model can learn arbitrary coordination patterns from data. The disadvantage is that the action space is large ($H \times 16$ dimensions for typical settings), requiring expressive generative models to capture the joint distribution.

Flow matching is well suited to this approach because it generates the entire action chunk in a single denoising process, naturally preserving inter-arm correlations. The flow field $\mathbf{v}_\theta$ operates on the full $H \times (d_L + d_R)$-dimensional space, learning the joint velocity field that transports noise to coordinated bimanual trajectories. This global denoising preserves correlations between left and right arm motions at every timestep within the~chunk.

In contrast, autoregressive generation of a joint action vector must predict left and right arm actions in some sequential order, potentially breaking symmetry. If the model generates left arm actions before right arm actions (or vice versa), the second arm's predictions are conditioned on the first arm's, introducing an artificial asymmetry that may not reflect the actual task structure. While this asymmetry can be mitigated through data augmentation (randomly swapping left and right arm labels), it remains a conceptual limitation of autoregressive bimanual action generation.

\subsubsection{Independent Policies}

An alternative is to train separate policies for each arm, coordinated by a high-level planner. \textbf{Hi Robot}~\cite{shi2025hirobot} decomposes bimanual tasks hierarchically: a high-level VLM generates subtask descriptions for each arm, and separate low-level policies execute them. This approach simplifies each policy's action space but requires the high-level planner to handle coordination timing and conflict avoidance.

In practice, independent policies work well for loosely coupled tasks (\eg, one arm holds an object while the other operates on it) but struggle with tightly coupled tasks (\eg, folding, where both arms must move in concert). The coordination information that joint policies learn implicitly must be provided explicitly through the high-level planner's~instructions.

\subsubsection{Leader--Follower}

In the leader--follower strategy, one arm (the leader) executes the primary manipulation while the other (the follower) adapts to maintain a constraint (\eg, holding an object stable). This asymmetric decomposition reduces the effective planning complexity and can be encoded in the VLA by conditioning one arm's actions on the other's predicted trajectory. Several bimanual VLA systems implement soft leader--follower coordination implicitly through the joint action space, where the model learns that one arm typically initiates contact while the other provides support.

\subsection{Contact-Rich Bimanual Tasks}
\label{sec:bimanual:contact}

Contact-rich tasks, where both arms simultaneously exert forces on an object, are among the most challenging for VLAs. Examples include inserting a peg with one arm while the other holds the socket, tightening a cap on a bottle held by the other arm, and assembling parts that require precise force alignment.

Contact-rich bimanual performance has been most thoroughly evaluated on box assembly, where one arm holds a box while the other folds flaps. \textbf{$\pi_0$}~\cite{black2024pi0} achieves smooth force profiles on this task through its flow-matching action head, avoiding the jerkiness of discrete-action policies. An equally important insight came from \textbf{ALOHA (ACT)}~\cite{zhao2023aloha}: action chunking is critical for contact-rich tasks, as single-step predictions produce oscillatory contact forces while chunked predictions maintain stable contact throughout a manipulation primitive.

Most VLAs operate in position or velocity space without explicit force feedback, inferring contact states from visual cues alone. \textbf{$\pi_0$}~\cite{black2024pi0} learns appropriate forces during box assembly purely from visual demonstrations, but vision alone likely breaks down for high-precision force-sensitive operations. \textbf{$\pi_0^*$}~\cite{amin2025pi06} partially addresses this through RL from autonomous practice, where the robot discovers effective force profiles through trial and error. The difficulty scales with contact points: current VLAs handle two-point contact well, but multi-point contact and multi-fingered dexterous manipulation remain open frontiers.

\subsection{Deformable Object Manipulation}
\label{sec:bimanual:deformable}

Deformable objects (fabric, rope, dough, plastic bags) present distinct challenges for bimanual VLAs: the object state is high-dimensional and partially observable, and manipulation requires coordinated bimanual actions that account for material dynamics.

Laundry folding is the canonical bimanual deformable-object task and, until recently, an unsolved problem. Success rates above 80\% on T-shirt folding became possible when \textbf{$\pi_0$}~\cite{black2024pi0} combined flow matching with long action chunks, learning the complex bimanual coordination required to pinch, lift, fold, and smooth fabric. The success relies on long action chunks ($H = 50$, as analyzed in Section~\ref{sec:action:chunking}) that capture the full folding motion as a continuous trajectory and on the VLM backbone's ability to visually parse the garment's~configuration.

Building on this, \textbf{$\pi_{0.5}$}~\cite{black2025pi05} generalized folding to arbitrary garments in novel homes. The hierarchical architecture decomposes folding into subgoals (\eg, ``pick up the left sleeve'', ``fold it toward the center''), with the high-level VLM reasoning about garment topology and the low-level policy handling motor execution. Current success rates on fabric folding are measured on a narrow range of garment types; generalization to thin, slippery, or multi-layered fabrics remains undemonstrated.

Beyond fabric, bimanual deformable-object manipulation encompasses rope knotting, dough shaping, and bag opening. The difficulty increases from 1D deformation (rope) through 2D (fabric with self-occlusion) to 3D (dough, clay). Current VLAs address these through large action chunks that capture entire primitives, which works when deformation is predictable but fails for materials with complex dynamics. Integrating tactile sensing to detect material state is a promising direction.

\subsection{Long-Horizon Bimanual Tasks}
\label{sec:bimanual:longhorizon}

Long-horizon tasks require the robot to execute many bimanual primitives in sequence, with the correct ordering determined by task semantics. Table clearing, for example, requires picking up plates, stacking them, wiping the table, and placing items in a bin: a sequence of 10--20 bimanual actions over several minutes.

\textbf{$\pi_{0.5}$}~\cite{black2025pi05} handles long-horizon tasks through its hierarchical architecture: the high-level VLM maintains a task plan and generates subgoal instructions, while the low-level policy executes each subgoal. The high-level model can re-plan based on visual feedback, recovering from failures or adapting to unexpected states. \textbf{Hi Robot}~\cite{shi2025hirobot} similarly uses hierarchical VLA reasoning for open-ended instructions, decomposing ``tidy the desk'' into a sequence of specific bimanual actions.

Prior work on LLM-based planning provides the conceptual foundation for long-horizon VLA reasoning. \textbf{Code as Policies}~\cite{liang2023code} uses LLMs to generate executable code that sequences manipulation primitives, while \textbf{SayCan}~\cite{ahn2022saycan} grounds LLM proposals in learned affordance scores, ensuring that only feasible actions are selected. Both approaches separate high-level reasoning from low-level execution, a principle adopted by hierarchical~VLAs.

Complementary approaches learn long-horizon skills from unstructured data. \textbf{MimicPlay}~\cite{wang2023mimicplay} decomposes human play videos into plan representations, \textbf{PlayFusion}~\cite{chen2023playfusion} acquires skills via diffusion from play data, and Du~\etal~\cite{du2024learning} learn policies via text-guided video generation. \textbf{Look Before You Leap}~\cite{hu2023look} uses GPT-4V to preview action consequences. These planning strategies predate VLAs but are complementary and could be integrated with VLA execution.

Long-horizon bimanual tasks demand robust \emph{error recovery}. Hierarchical VLAs address this naturally: the high-level VLM detects failures and generates recovery subgoals. \textbf{$\pi_0^*$}~\cite{amin2025pi06} improves recovery through autonomous practice (Section~\ref{sec:training:rl}), learning robust behaviors that pure imitation cannot provide. The temporal extent also poses a computational challenge: a 5-minute task at $50\,\text{Hz}$ involves 15,000 control steps (300 chunk-level decisions with $H = 50$), requiring either hierarchical planning or long-context models.

Recent work addresses this limitation by equipping VLAs with explicit \emph{memory} mechanisms. \textbf{MEM}~\cite{torne2026mem} introduces \emph{Multi-Scale Embodied Memory}, a system that combines two complementary modalities: a video encoder for short-horizon, image-based memory (which supports in-context adaptation and occlusion recovery over seconds) and a language-based memory that maintains compressed text summaries of semantic events over long horizons (up to 15 min). Integrated into the $\pi_{0.6}$ VLA, MEM achieves state-of-the-art results on tasks such as recipe setup, kitchen cleanup, and grilled cheese preparation while matching non-memory VLAs on standard dexterous manipulation. Different time scales require different memory representations: dense visual context for recent events and compressed language for long-term semantic state.

Concurrent approaches explore complementary designs. Context-compression methods include \textbf{ContextVLA}~\cite{jang2025contextvla} (amortizing multi-frame context into a single token), \textbf{CronusVLA}~\cite{li2025cronusvla} (learnable temporal feature chunking), \textbf{BPP}~\cite{mark2026bpp} (conditioning on VLM-detected keyframes), and past-token prediction~\cite{torne2025pasttokenpred} ($3\times$ gains at $10\times$ reduced cost). Retrieval-based methods include \textbf{MemoryVLA}~\cite{shi2025memoryvla} (perceptual-cognitive memory bank; +14.6\% on Bridge), \textbf{SAM2Act}~\cite{fang2025sam2act} (episodic spatial memory; 86.8\% across 18 RLBench tasks), \textbf{MemER}~\cite{sridhar2025memer} (VLM-guided keyframe retrieval), and \textbf{CycleManip}~\cite{wei2025cyclemanip} (cost-aware historical sampling for cyclic tasks).

Memory-augmented VLAs enable long-horizon execution (up to 15 min), in-context adaptation, and partial observability handling. However, they face two intertwined challenges: \emph{causal confusion}, where the policy learns to copy its own past actions rather than reason about the current state, and \emph{train-inference shift}, where self-generated memory summaries at test time may contain compounding errors. MEM mitigates causal confusion via language compression that discards failed attempts, but the general problem remains unsolved. Additional limitations include computational overhead that scales with history length and information loss from memory compression (semantic events are retained, but fine-grained contact forces are not). Current memory systems are episodic, with no mechanism for accumulating knowledge across deployment sessions.

\subsection{Mobile Bimanual Manipulation}
\label{sec:bimanual:mobile}

Mobile bimanual systems add navigation to the manipulation challenge. The robot must move to the task location, position itself appropriately, and then perform bimanual manipulation, requiring coordination between the base and both arms.

Whole-body teleoperation and imitation learning for mobile bimanual tasks were first shown by \textbf{Mobile ALOHA}~\cite{fu2024mobilealoha}, which controls the mobile base and two arms simultaneously for tasks such as cooking and furniture assembly, with action chunks covering all degrees of freedom.

\textbf{$\pi_{0.5}$}~\cite{black2025pi05} deployed VLA-controlled mobile bimanual robots in real homes, achieving high success rates on household tasks. The hierarchical architecture separates navigation decisions (handled by the high-level VLM) from manipulation execution (handled by the low-level policy), simplifying the learning problem.

\begin{figure}[H]
\centering
\includegraphics[width=\textwidth]{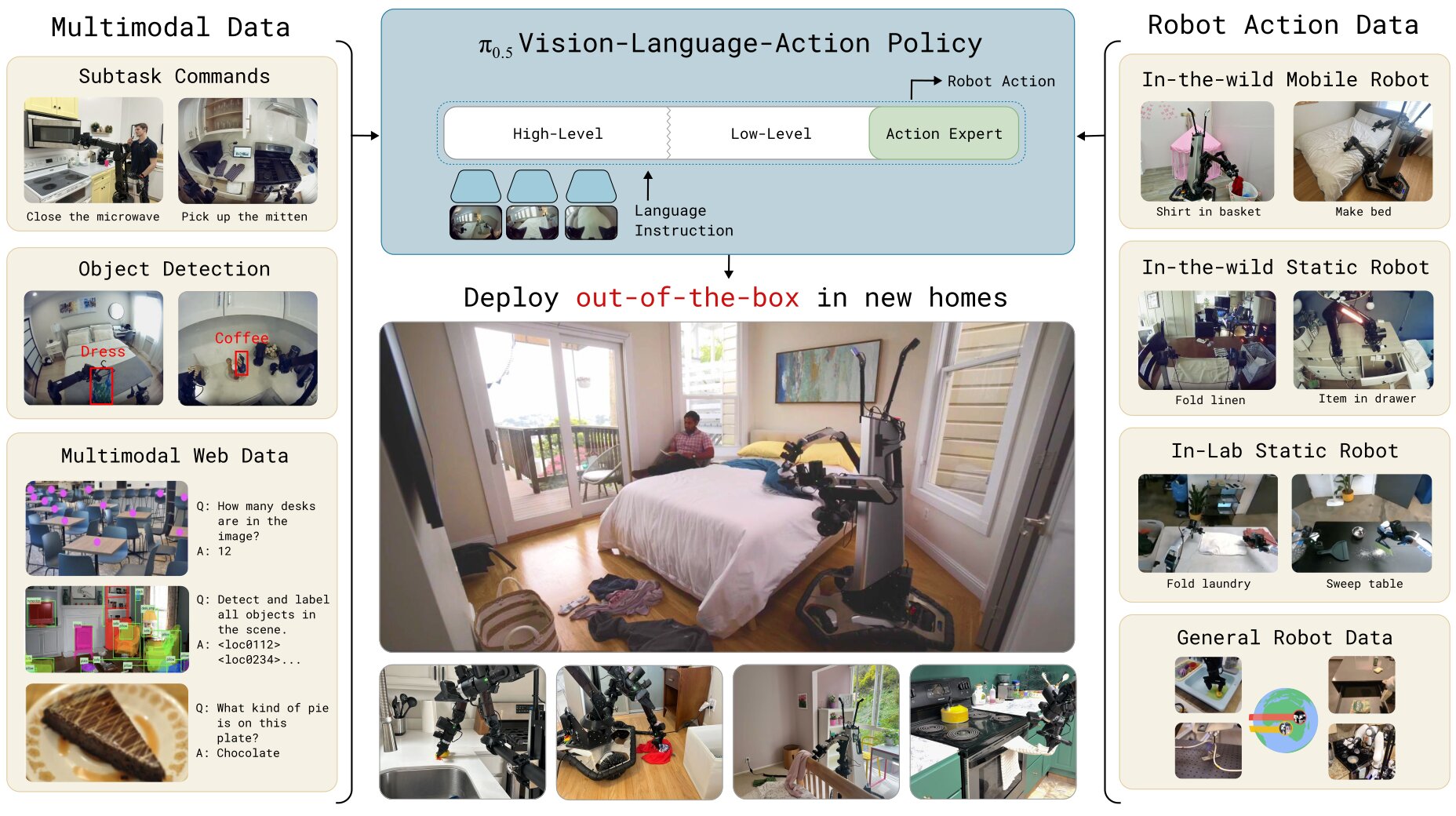}
\caption{Real-world %
 deployment of $\pi_{0.5}$ in homes. A hierarchical VLA decomposes high-level instructions into subgoals, with high success rates on household tasks such as table clearing and laundry folding. Reprinted with permission from Ref.~\cite{black2025pi05}. Copyright 2025, Black~et~al.}
\label{fig:pi05_deployment}
\end{figure}

\emph{Base-arm coordination} is the additional challenge: the base must position itself so both arms reach target objects. \textbf{UMI-on-Legs}~\cite{collaboration2025umi2} extends hardware-agnostic data collection to legged platforms, \textbf{BUMBLE}~\cite{shah2024bumble} addresses building-wide mobile manipulation, and industrial efforts (\textbf{AgiBot World}~\cite{bu2025agibot}, \textbf{NILS}~\cite{intelligence2025physical}) are scaling bimanual fleet data and policy learning. Figure~\ref{fig:pi05_deployment} shows $\pi_{0.5}$ deployed in real homes.

% ============================================================
% FIGURE: Pi0.5 Real-World Deployment
% ============================================================

Flow-based VLAs (\textbf{$\pi_0$}, \textbf{$\pi_0^*$}) currently lead on most bimanual tasks, with the joint action space approach outperforming decoupled approaches on tightly coupled tasks by preserving inter-arm correlations. For loosely coupled tasks, hierarchical approaches offer interpretability through auditable subgoal decompositions.

The coordination strategies analyzed above, joint action spaces for tightly coupled agents, hierarchical decomposition for complex tasks, leader--follower for asymmetric roles, are not specific to two-armed robots. They apply whenever a VLA must coordinate multiple coupled actuators from shared observations. We now examine a domain where exactly the same coordination problem arises in a different physical setting: unmanned aerial robotics.

% ============================================================
% 9. VLA FOR AERIAL ROBOTICS AND DRONES
% ============================================================
\section{VLA for Unmanned Aerial Robotics and Drones}
\label{sec:aerial}

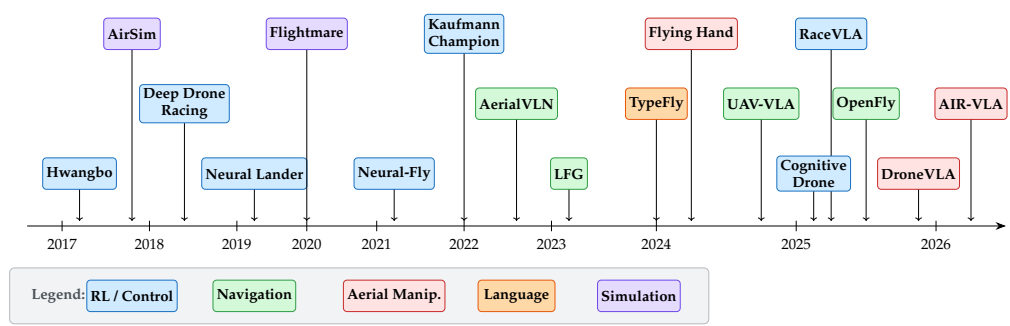
\begin{figure}[H]
\centering
\resizebox{\textwidth}{!}{%
\begin{tikzpicture}[
    every node/.style={font=\large, align=center},
    rl/.style={rectangle, rounded corners=3pt, draw=cBlue, fill=cBlueFill, minimum height=9mm, font=\large\bfseries, inner sep=3pt, line width=0.6pt},
    nav/.style={rectangle, rounded corners=3pt, draw=cGreen, fill=cGreenFill, minimum height=9mm, font=\large\bfseries, inner sep=3pt, line width=0.6pt},
    manip/.style={rectangle, rounded corners=3pt, draw=cRed, fill=cRedFill, minimum height=9mm, font=\large\bfseries, inner sep=3pt, line width=0.6pt},
    lang/.style={rectangle, rounded corners=3pt, draw=cOrange, fill=cOrangeFill, minimum height=9mm, font=\large\bfseries, inner sep=3pt, line width=0.6pt},
    sim/.style={rectangle, rounded corners=3pt, draw=cViolet, fill=cVioletFill, minimum height=9mm, font=\large\bfseries, inner sep=3pt, line width=0.6pt},
]
% Timeline axis
\coordinate (axislevel) at (0, 0.15);
\draw[-{Stealth[length=3mm]}, thick] (0,0) -- (28,0);
\foreach \x/\yr in {1/2017, 3.5/2018, 6/2019, 8/2020, 10/2021, 12.5/2022, 15/2023, 18/2024, 22/2025, 26/2026} {
    \draw[thick] (\x, -0.15) -- (\x, 0.15);
    \node[below, font=\large] at (\x, -0.25) {\yr};
}

% Draw RaceVLA node and its straight arrow FIRST so lower nodes render on top
\node[rl] (nRace) at (23, 5.5) {RaceVLA};
\draw[->, thin] (nRace.south) -- (23, 0.15);

% Row A (y=1.5) — renders on top of RaceVLA arrow
\node[rl] (nHwang) at (1.5, 1.5) {Hwangbo};
\node[rl] (nNeural) at (6.5, 1.5) {Neural Lander};
\node[rl] (nFly) at (10.5, 1.5) {Neural-Fly};
\node[nav] (nLFG) at (15.5, 1.5) {LFG};
\node[rl] (nCog) at (22.5, 1.5) {Cognitive\\Drone};
\node[manip] (nDrone) at (25.5, 1.5) {DroneVLA};

% Row B (y=3.5)
\node[rl] (nDDR) at (4.5, 3.5) {Deep Drone\\Racing};
\node[nav] (nAVLN) at (14, 3.5) {AerialVLN};
\node[lang] (nType) at (18, 3.5) {TypeFly};
\node[nav] (nUAV) at (21, 3.5) {UAV-VLA};
\node[nav] (nOpen) at (24, 3.5) {OpenFly};
\node[manip] (nAIR) at (27, 3.5) {AIR-VLA};

% Row C (y=5.5) — RaceVLA already placed above
\node[sim] (nAir) at (3, 5.5) {AirSim};
\node[sim] (nFlight) at (8, 5.5) {Flightmare};
\node[rl] (nKauf) at (12.5, 5.5) {Kaufmann\\Champion};
\node[manip] (nHand) at (19, 5.5) {Flying Hand};

% Arrows from nodes to timeline axis
\draw[->, thin] (nHwang.south) -- (1.5, 0.15);
\draw[->, thin] (nNeural.south) -- (6.5, 0.15);
\draw[->, thin] (nFly.south) -- (10.5, 0.15);
\draw[->, thin] (nLFG.south) -- (15.5, 0.15);
\draw[->, thin] (nCog.south) -- (22.5, 0.15);
\draw[->, thin] (nDrone.south) -- (25.5, 0.15);
\draw[->, thin] (nDDR.south) -- (4.5, 0.15);
\draw[->, thin] (nAVLN.south) -- (14, 0.15);
\draw[->, thin] (nType.south) -- (18, 0.15);
\draw[->, thin] (nUAV.south) -- (21, 0.15);
\draw[->, thin] (nOpen.south) -- (24, 0.15);
\draw[->, thin] (nAIR.south) -- (27, 0.15);
\draw[->, thin] (nAir.south) -- (3, 0.15);
\draw[->, thin] (nFlight.south) -- (8, 0.15);
\draw[->, thin] (nKauf.south) -- (12.5, 0.15);
\draw[->, thin] (nHand.south) -- (19, 0.15);

% Legend background box (shifted down)
\fill[cGrayFill, rounded corners=4pt] (-0.5, -2.8) rectangle (19.5, -1.2);
\draw[cGray!40, rounded corners=4pt, line width=0.4pt] (-0.5, -2.8) rectangle (19.5, -1.2);
\node[font=\large\bfseries, text=cGray] at (0, -2.0) [right] {Legend:};
\node[rl, minimum width=22mm] at (3, -2.0) {RL / Control};
\node[nav, minimum width=22mm] at (6.5, -2.0) {Navigation};
\node[manip, minimum width=22mm] at (10.5, -2.0) {Aerial Manip.};
\node[lang, minimum width=22mm] at (14, -2.0) {Language};
\node[sim, minimum width=22mm] at (17.5, -2.0) {Simulation};
\path (-1, -3.3) rectangle (29, 6.5);
\end{tikzpicture}%
}% end resizebox
\caption{Timeline of unmanned aerial robotics milestones for learning-based drone control (2017--2026). Colors indicate the research area: RL-based control (blue), vision--language navigation (green), aerial manipulation (red), language-guided planning (orange), and simulation platforms (purple). Early work focused on RL for agile flight and simulators; 2023--2024 saw the emergence of language-guided navigation; 2025--2026 marks the arrival of full VLA systems (CognitiveDrone, RaceVLA, DroneVLA, AIR-VLA) that integrate perception, language, and action generation end-to-end.}
\label{fig:aerial_timeline}
\end{figure}

Section~\ref{sec:bimanual} showed how VLAs coordinate two arms through joint action spaces, hierarchical planning, and action chunking. Unmanned aerial robotics faces the same coordination problem in a different physical setting. A single drone must coordinate thrust, attitude, and (optionally) gripper commands; a multi-drone system must coordinate an entire fleet. The VLA machinery (VLM backbone for language grounding, flow matching or diffusion for smooth trajectory generation, action chunking for temporal coherence) applies directly. What changes is the action space: whereas bimanual VLAs generate joint positions or end-effector poses, aerial VLAs must produce velocity commands, waypoints, or low-level thrust-and-torque signals for underactuated platforms operating in three-dimensional space. Latency constraints are stricter (${\geq}100\,\text{Hz}$ for stable flight versus ${\sim}50\,\text{Hz}$ for manipulation), the observation space often includes GPS, IMU, and depth sensing alongside monocular or stereo vision, and the environment is outdoor, three-dimensional, and wind-affected. Table~\ref{tab:aerial_methods} compares representative methods; Figure~\ref{fig:aerial_timeline} charts the key~milestones.

% ============================================================
% FIGURE: AERIAL ROBOTICS TIMELINE
% ============================================================

\subsection{VLA-Based Drone Navigation and Control}
\label{sec:aerial:nav}

\subsubsection{Vision--Language Navigation for UAVs}

Vision--language navigation (VLN) requires a drone to reach a goal described in natural language (``fly above the red building'', ``turn left at the intersection'') using only visual observations. VLN is related to but narrower than VLA: it addresses the \emph{navigation} subtask---selecting waypoints to reach a described destination---while VLA refers to the full perception-to-action pipeline including low-level motor control (thrust, attitude commands). Although VLN originated in indoor settings, the aerial variant poses distinct challenges: a vastly different visual perspective, a larger action space that includes altitude, and outdoor visual diversity.

The \textbf{AerialVLN} benchmark~\cite{liu2023aerialvln} established this task with over 25,000 instruction--trajectory pairs across urban and rural environments, revealing that indoor VLN methods transfer poorly to aerial scenes. A zero-shot alternative, \textbf{LFG}~\cite{shah2023navigation}, sidesteps task-specific training entirely by having an LLM convert language instructions into spatial cost maps that a standard path planner optimizes over.

The most complete aerial VLA to date is \textbf{UAV-VLA}~\cite{uavvla2025}, which processes satellite imagery through a VLM backbone and generates full mission plans (waypoints, altitudes, sensor configurations) from natural language. On a 100K-mission dataset, it produces plans $6.5\times$ faster than human operators at comparable quality, showing that VLAs can scale to operational aerial planning beyond single-flight control.

More broadly, the field is rapidly standardizing. \textbf{UAV-VLN}~\cite{uavvln2025} parses instructions into structured sub-goals grounded by a vision model; \textbf{OpenFly}~\cite{openfly2025} provides a large-scale benchmark spanning urban, suburban, rural, and industrial settings; and \textbf{CityNavAgent}~\cite{citynavagent2025} adds a persistent semantic map that supports city-scale navigation with hierarchical planning.
Beyond urban environments, \textbf{AgriVLN}~\cite{agrivln2024} extends VLN to agricultural settings, providing 1,560 episodes across six outdoor scene types (farms, greenhouses, forests, mountains, gardens, and villages) and demonstrating that VLN models trained on indoor or urban data transfer poorly to unstructured agricultural landscapes.

\subsubsection{End-to-End Learned Flight Control}

\begin{figure}[H]
%\centering
\includegraphics[width=\textwidth]{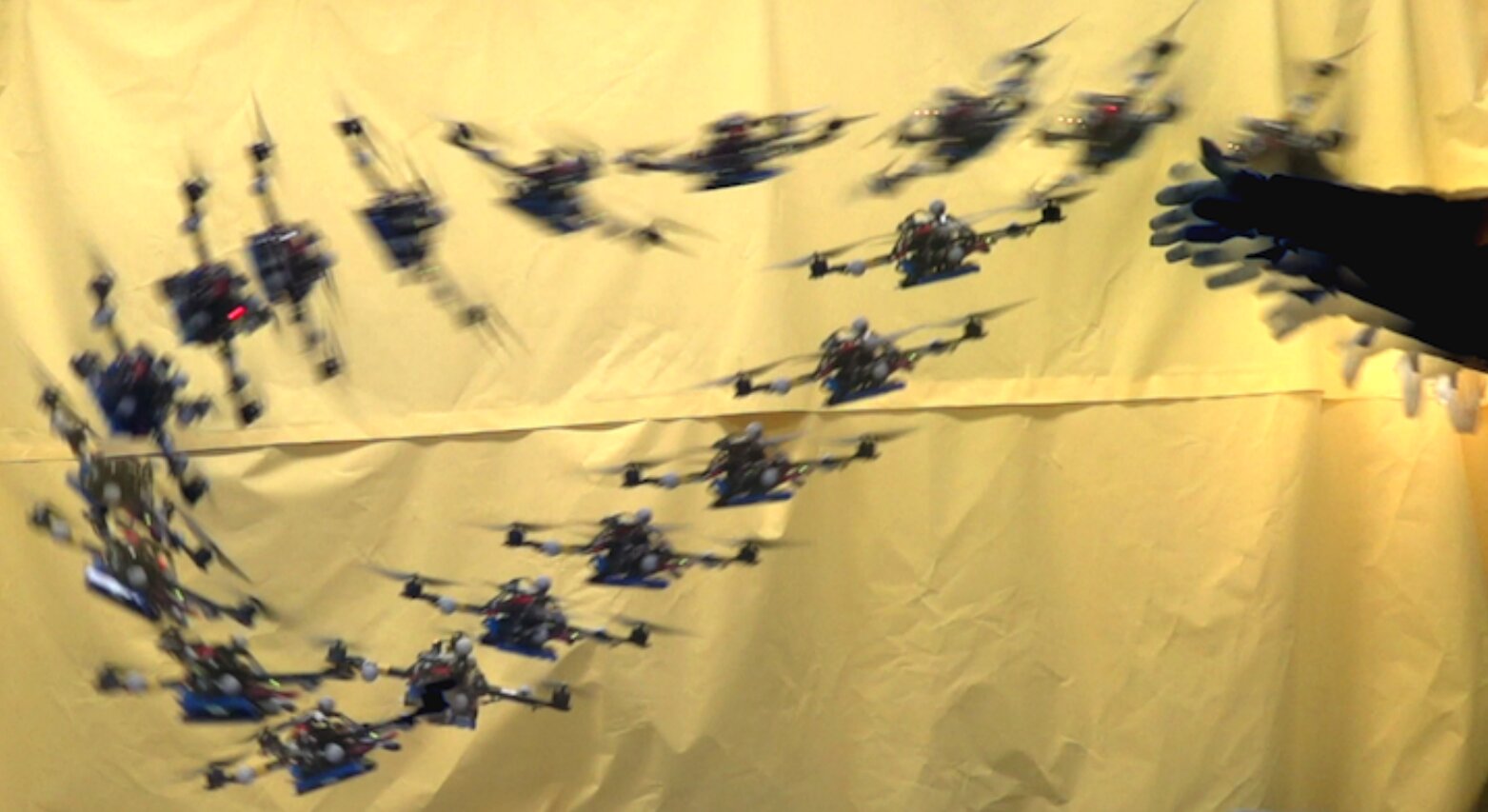}
\caption{An RL-trained quadrotor recovering from an inverted throw at $5\,\text{m/s}$. The policy maps state to motor commands at $7\,\upmu\text{s}$ per step, establishing the viability of learned end-to-end drone control. Reprinted with permission from Ref.~\cite{hwangbo2017quadrotor}. Copyright 2017, Hwangbo~et~al., IEEE.}
\label{fig:hwangbo_quadrotor}
\end{figure}

In contrast to modular pipelines, end-to-end approaches map raw sensor observations directly to flight commands, bypassing the traditional perception--planning--control pipeline. The feasibility of this approach was established early: a single neural network trained with RL~\cite{hwangbo2017quadrotor} can map quadrotor state directly to motor commands, stabilizing the vehicle even when thrown upside-down at $5\,\text{m/s}$, with policy evaluation taking only $7\,\upmu\text{s}$ per step (Figure~\ref{fig:hwangbo_quadrotor}); this proves that learned policies can replace hand-designed cascaded PID controllers for agile flight.

% ============================================================
% FIGURE: Hwangbo Quadrotor Recovery
% ============================================================

Subsequent learned systems pushed the performance frontier to superhuman levels. The landmark result is an RL-trained autonomous racing policy~\cite{kaufmann2023champion} that processes onboard vision and IMU at $100\,\text{Hz}$ and defeated world-champion human pilots. A complementary approach, neural residual dynamics models~\cite{shi2019neural}, uses supervised learning on flight data to learn the gap between the physics model and reality, improving autonomous landing accuracy by compensating for unmodeled aerodynamic effects.

Two recent systems bring the VLA framework directly to drones. \textbf{CognitiveDrone}~\cite{cognitivedrone2025}, trained on 8000+ simulated trajectories, generates real-time 4D actions $(x, y, z, \text{yaw})$ from first-person imagery and text instructions. Its R1 variant adds VLM-based chain-of-thought reasoning before acting, which lifts the success rate to 77.2\%, a 30\% gain that demonstrates the value of deliberation for aerial cognitive tasks. \textbf{RaceVLA}~\cite{racevla2025} trains on expert pilot demonstrations annotated with language (``aggressive apex cutting'', ``conservative trajectory'') and produces stylistically diverse racing trajectories, going beyond time-optimal control to capture human-interpretable flight behavior.

World models are also gaining traction for aerial control. \textbf{Dream to Fly}~\cite{dream2fly2025} learns a latent dynamics model from visual observations and plans by simulating future trajectories in the learned space, reducing real-world data needs by an order of magnitude compared to model-free RL (see Section~\ref{sec:crosscutting}). Robustness to real-world disturbances remains a key gap: \textbf{Neural-Fly}~\cite{neuralfly2022} addresses this through rapid online adaptation, maintaining stable aggressive flight in winds exceeding $12\,\text{m/s}$ by learning a wind-invariant representation from just a few flight segments. Diffusion-based policies, originally developed for manipulation (Section~\ref{sec:architectures}), are also being applied to generate smooth, multi-modal drone trajectories.

\subsection{Aerial Manipulation}
\label{sec:aerial:manipulation}

Aerial manipulation, where drones grasp, transport, and interact with objects, directly inherits the coordination challenges discussed in the bimanual context (Section~\ref{sec:bimanual}). A drone performing aerial grasping must simultaneously stabilize its flight while executing precise gripper motions, a challenge analogous to bimanual base-arm coordination (Section~\ref{sec:bimanual:mobile}).

\subsubsection*{Grasping and Payload Transport} %

Aerial manipulation combines the challenges of flight stabilization and precise object interaction. While RL-based flight control~\cite{hwangbo2017quadrotor} established that learned policies can stabilize quadrotors under extreme conditions, extending this to aerial grasping requires additionally coordinating gripper commands and compensating for payload-induced dynamics shifts. Two recent systems bring the full VLA pipeline to this problem. \textbf{DroneVLA}~\cite{dronevla2026} integrates open-vocabulary object detection (Grounding DINO), gripper pose estimation (MediaPipe), and visual servoing into a language-commanded retrieval system: given ``pick up the red box and deliver it to the table,'' a VLM decomposes the instruction into manipulation sub-goals executed in sequence. \textbf{AIR-VLA}~\cite{airvla2026} provides the first benchmark and simulation testbed dedicated to aerial manipulation VLAs, with 3000 teleoperated demonstrations spanning base control, object understanding, semantic reasoning, and long-horizon planning; it systematically evaluates mainstream VLA and VLM models, revealing current capabilities and limitations for coupled UAV-manipulator control.

\begin{figure}[H]
%\centering
\includegraphics[width=\textwidth]{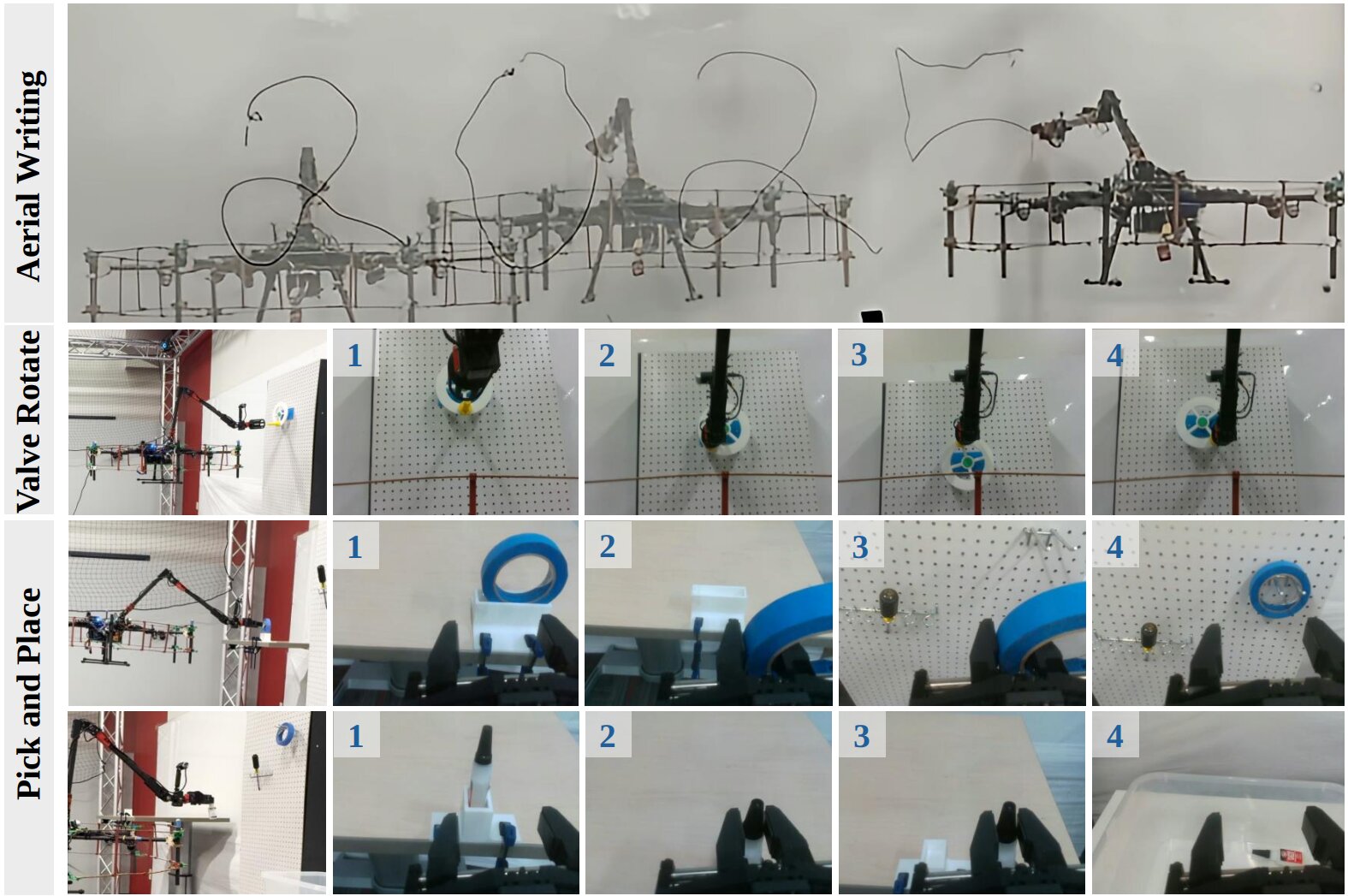}
\caption{Flying %
 Hand: a fully actuated hexarotor with a 4-DOF arm performing writing, peg-in-hole, and pick-and-place via ACT, demonstrating that action chunking transfers from manipulation to aerial systems. The numbers 1--4 along each row index successive video frames of the same task sequence (1: approach, 2: contact, 3: execution, 4: completion). Reprinted with permission from Ref.~\cite{flyinghand2025}. Copyright 2025, He~et~al.}
%AUTHOR REPLY (re: numbers 1-4): They denote consecutive time-step frames of each task; this is now explained in the caption.
\label{fig:flying_hand}
\end{figure}

A particularly relevant platform is \textbf{Flying Hand}~\cite{flyinghand2025}, a fully actuated hexarotor with a 4-DOF arm that formulates control in the end-effector frame, decoupling manipulation precision from flight stabilization (Figure~\ref{fig:flying_hand}). Its imitation learning policy uses ACT (Section~\ref{sec:action}) to perform writing, peg-in-hole insertion, and pick-and-place, directly demonstrating that action chunking transfers from bimanual to aerial manipulation. The connection to bimanual coordination becomes explicit in an aerial harvesting system~\cite{avianharvest2024} where a dual-arm drone picks avocados: one arm stabilizes the branch while the other detaches the fruit by rotation, a leader--follower strategy identical to those analyzed in Section~\ref{sec:bimanual:coord}.

% ============================================================
% FIGURE: Flying Hand Aerial Manipulation
% ============================================================

These platforms combine flight commands with gripper commands, yielding high-dimensional action spaces that benefit from the same chunking strategies used for bimanual VLAs. Aerial inspection tasks (close-proximity structure navigation, contact-based measurement) further build on the contact-rich manipulation insights from Section~\ref{sec:bimanual:contact}. A significant gap remains, however: all current aerial manipulation VLAs have been tested only on simplified pick-and-place tasks with lightweight objects. Precision manipulation under wind disturbances and with heavy or awkward payloads has not been demonstrated.

\subsection{Language-Guided Drone Missions}
\label{sec:aerial:language}

\subsubsection{Natural Language to Flight Plans}

The same Code-as-Policies idea used for manipulation~\cite{liang2023code} (Section~\ref{sec:language:hierarchical}) extends naturally to drones: prompting an LLM with a drone API description lets it convert ``survey the perimeter at 20 m'' into executable waypoint commands with no task-specific training. \textbf{AeroAgent}~\cite{zhang2025aeroagent} adds safety awareness to this approach, decomposing complex missions into atomic flight actions while respecting no-fly zones and altitude limits. A latency bottleneck remains, however: generating free-form code---\ie, having the LLM produce arbitrary Python %
 or similar programs without structural constraints---is slow. \textbf{TypeFly}~\cite{typefly2024} addresses this by constraining the LLM to output programs in MiniSpec, a minimal drone-specific language with primitives like takeoff, move, rotate, and sense. The restricted grammar cuts generation latency below $500\,\text{ms}$, an order of magnitude faster than unconstrained approaches, making real-time mission replanning practical.

\subsubsection{Interactive and Corrective Language Control}

Real-time language correction during flight (``go higher'', ``move left'', ``stop'') requires low-latency VLA inference. This setting parallels the interactive language control studied for manipulation~\cite{lynch2023interactive} but with stricter latency requirements due to flight dynamics. Current approaches use lightweight VLM encoders or pre-computed language embeddings to minimize inference overhead.

\subsection{Multi-Agent Aerial Systems}
\label{sec:aerial:multiagent}

Multi-drone coordination with VLAs exhibits direct structural parallels to bimanual coordination (Section~\ref{sec:bimanual:coord}). The joint action space approach for bimanual VLAs, where a single model generates actions for both arms simultaneously, naturally extends to multi-drone systems where a centralized policy generates waypoints for all drones in the swarm.

Multi-drone coordination mirrors the three bimanual strategies from Section~\ref{sec:bimanual:coord}. Centralized policies generate joint actions for all drones but face the same dimensionality scaling as bimanual joint action spaces. Decentralized approaches reduce per-agent complexity but require explicit coordination. Hierarchical approaches, where a high-level VLM assigns subgoals to individual drones (\textbf{Hi Robot}~\cite{shi2025hirobot}, \textbf{$\pi_{0.5}$}~\cite{black2025pi05}), offer the best scalability. \textbf{Decentralized MARL}~\cite{marl_drones2021} trains swarm policies where each drone outputs velocity commands from local observations, scaling to 10+ drones. Graph neural network architectures that model inter-drone communication provide a natural framework for swarm VLAs. Integrating VLM-based task assignment with MARL execution is a promising direction. However, no multi-drone VLA has been demonstrated on physical hardware; all results remain simulation-only.

\subsection{UAV--UGV Collaborative Systems}
\label{sec:aerial:uavugv}

Heterogeneous UAV--UGV systems exploit complementary capabilities: drones provide aerial survey while ground robots perform manipulation. VLMs provide a natural coordination interface via language-based task allocation. Cross-embodiment VLA pre-training (Section~\ref{sec:language:cross}) is directly applicable; \textbf{Octo}~\cite{octo2024} and \textbf{Octo~2.0}~\cite{team2024octo2} already span manipulation, navigation, and locomotion embodiments. The coordination challenge mirrors bimanual leader--follower strategies (Section~\ref{sec:bimanual:coord}): one agent (typically the drone) provides context while the other (the ground robot) executes manipulation. No published system yet deploys a shared VLA policy across both a UAV and a UGV in a single mission, making this an open research direction (Section~\ref{sec:discussion:directions}).

\subsection{Sim-to-Real Transfer for Aerial VLAs}
\label{sec:aerial:sim2real}

Simulation is especially important for aerial VLAs because real-world drone data collection is expensive, risky, and constrained by regulations.

\subsubsection{Simulation Environments}

Two simulators dominate aerial VLA research. \textbf{AirSim}~\cite{shah2018airsim}, built on Unreal Engine, provides photorealistic rendering with accurate flight dynamics and a rich sensor API (cameras, IMU, GPS, LiDAR). \textbf{Flightmare}~\cite{song2021flightmare} takes a different approach, decoupling rendering from physics to reach ${\sim}200\times$ real-time speeds for large-scale parallel RL training. Additional platforms (\textbf{RotorS}~\cite{furrer2016rotors}, \textbf{Isaac Sim}, \textbf{Gazebo}) serve complementary fidelity and scale requirements.

Large-scale datasets fill the gap between simulated and real-world training data. \textbf{TartanAir}~\cite{tartanair2020} spans hundreds of scenes with diverse weather and lighting; \textbf{Mid-Air}~\cite{midair2019} focuses on low-altitude flights (1--20\,m) with stereo images, depth, and semantic labels.

\subsubsection{Domain Adaptation and Reality Gap}

The sim-to-real gap for aerial systems involves both visual and dynamics discrepancies. \emph{Visual domain randomization} during training improves transfer of vision-based policies by exposing the model to varied textures, lighting, and weather conditions. \emph{Dynamics randomization} varies mass, inertia, drag, and motor characteristics to produce policies robust to the physical reality gap. These techniques parallel the sim-to-real methods used for manipulation VLAs (Section~\ref{sec:crosscutting}), with the additional challenge that aerodynamic effects (ground effect, wind gusts, rotor wash) are difficult to simulate accurately.

% ============================================================
% TABLE: AERIAL METHODS
% ============================================================
\begin{table}[H]
%\centering
\caption{Representative %
 VLA and learning-based methods for unmanned aerial robotics. Action type indicates the output space of the learned policy. Sim indicates whether the method uses simulation for training.}
\label{tab:aerial_methods}
\fontsize{7.5}{7.5}\selectfont

\begin{adjustwidth}{-\extralength}{0cm}
%\centering %% If there is a figure in wide page, please release command \centering
\begin{tabularx}{\fulllength}{llllccl}
\toprule

\textbf{Method} & \textbf{Task} & \textbf{Approach} & \textbf{Action Type} & \textbf{Sim} & \textbf{Year} & \textbf{Highlights} \\
\midrule
\multicolumn{7}{l}{\textit{Navigation and Control %
}} \\
AerialVLN~\cite{liu2023aerialvln} & VL navigation & VLN baseline & Waypoints & \checkmark & 2023 & First outdoor aerial VLN benchmark \\
LFG~\cite{shah2023navigation} & Language nav. & LLM $\rightarrow$ cost map & Waypoints & -- & 2023 & Zero-shot LLM-guided navigation \\
UAV-VLA~\cite{uavvla2025} & Mission gen. & VLA (sat. imagery) & Waypoints & -- & 2025 & $6.5\times$ faster than human; 100K missions \\
UAV-VLN~\cite{uavvln2025} & VL navigation & LLM + vision & Waypoints & \checkmark & 2025 & End-to-end VLN with LLM parsing \\
OpenFly~\cite{openfly2025} & VLN benchmark & Toolchain & Waypoints & \checkmark & 2025 & Large-scale aerial VLN benchmark \\
CityNavAgent~\cite{citynavagent2025} & City-scale nav. & Hierarchical VLN & Waypoints & \checkmark & 2025 & Semantic planning + global memory \\
AgriVLN~\cite{agrivln2024} & Agricultural nav. & VLN benchmark & Waypoints & \checkmark & 2025 & 1560 episodes; 6 outdoor scene types \\
Hwangbo~\etal~\cite{hwangbo2017quadrotor} & Stabilization & RL & Motor cmds & \checkmark & 2017 & $7\,\upmu$s inference; thrown recovery \\
Kaufmann~\etal~\cite{kaufmann2023champion} & Drone racing & RL & Motor cmds & \checkmark & 2023 & Superhuman agile flight \\
CognitiveDrone~\cite{cognitivedrone2025} & Cognitive tasks & VLA & 4D $(x,y,z,\text{yaw})$ & \checkmark & 2025 & 77.2\% success with VLM reasoning \\
RaceVLA~\cite{racevla2025} & Drone racing & VLA & Velocity cmds & \checkmark & 2025 & Human-like racing behavior \\
Neural-Fly~\cite{neuralfly2022} & Agile flight & Adaptive NN & Motor cmds & -- & 2022 & Online adaptation in strong winds \\
Dream to Fly~\cite{dream2fly2025} & Vision flight & Model-based RL & Velocity cmds & \checkmark & 2025 & Learned world model for planning \\
\midrule
\multicolumn{7}{l}{\textit{Aerial Manipulation}} \\
Hwangbo~\etal~\cite{hwangbo2017quadrotor} & Quadrotor ctrl & RL & Thrust cmds & \checkmark & 2017 & RL-based flight control \\
DroneVLA~\cite{dronevla2026} & Object retrieval & VLA + servoing & EE pose + grip & -- & 2026 & Language-commanded aerial manipulation \\
AIR-VLA~\cite{airvla2026} & Aerial manip. & VLA benchmark & Flight + grip & -- & 2026 & Safety-constrained; 20\,Hz testbed \\
Flying Hand~\cite{flyinghand2025} & Dexterous manip. & ACT + MPC & 6-DOF + 4-DOF arm & -- & 2025 & Hexarotor; writing, peg-in-hole \\
Aerial Bimanual~\cite{avianharvest2024} & Harvesting & Dual-arm aerial & Dual-arm cmds & -- & 2024 & Bimanual aerial manipulation \\
\midrule
\multicolumn{7}{l}{\textit{Language-Guided Missions}} \\
AeroAgent~\cite{zhang2025aeroagent} & Mission plan & LLM agent & API calls & -- & 2023 & LLM mission decomposition \\
TypeFly~\cite{typefly2024} & Mission plan & LLM $\rightarrow$ MiniSpec & API calls & -- & 2024 & Low-latency program generation \\
\midrule
\multicolumn{7}{l}{\textit{Multi-Agent}} \\
MARL Swarms~\cite{marl_drones2021} & Formation & Decentralized MARL & Velocity cmds & \checkmark & 2021 & Scalable to 10+ drones \\
\midrule
\multicolumn{7}{l}{\textit{Simulation, Datasets, and Sim-to-Real}} \\
AirSim~\cite{shah2018airsim} & Sim platform & UE4 rendering & Various & \checkmark & 2018 & Photorealistic drone sim \\
Flightmare~\cite{song2021flightmare} & Sim platform & Parallel RL & Various & \checkmark & 2021 & $200\times$ real-time training \\
TartanAir~\cite{tartanair2020} & Dataset & Multi-modal & -- & \checkmark & 2020 & Diverse visual conditions; SLAM focus \\
Mid-Air~\cite{midair2019} & Dataset & Multi-modal & -- & \checkmark & 2019 & Low-altitude flights; depth + semantics \\
\bottomrule
\end{tabularx}\end{adjustwidth}
\end{table}

The aerial landscape shows rapid progress: 2025--2026 has seen a surge of aerial VLA systems and benchmarks (UAV-VLA, CognitiveDrone, DroneVLA, AIR-VLA, Flying Hand) that target mapping observations and language to flight actions. The bridging of bimanual coordination with aerial manipulation~\cite{avianharvest2024,flyinghand2025} suggests bidirectional technical transfer. Research directions for advancing this convergence appear in Section~\ref{sec:discussion}.

% ============================================================
% 10. LANGUAGE GROUNDING, REASONING, AND GENERALIZATION
% ============================================================
\section{Language Grounding, Reasoning, and Generalization}
\label{sec:language}

VLAs condition on natural-language instructions, inheriting the semantic understanding of pre-trained VLMs. This section examines language grounding, hierarchical reasoning, open-ended instruction following, and cross-embodiment generalization, mechanisms applicable to both manipulation and aerial domains.

\subsection{Language-Conditioned Policies}
\label{sec:language:conditioned}

The roots of language-conditioned policy learning trace back to \textbf{Language-conditioned IL}~\cite{stepputtis2020lcil}, which established multi-task learning from language instructions, and \textbf{BC-Z}~\cite{jang2022bcz}, which scaled it to zero-shot generalization. VLAs extend this lineage by processing language and image tokens through shared Transformer layers, supporting deep cross-modal reasoning. Emergent grounding (following novel phrasings and generalizing to unseen objects) was first observed in \textbf{RT-2}~\cite{brohan2023rt2}. In contrast, \textbf{$\pi_0$}~\cite{black2024pi0} conditions the flow-matching head on VLM hidden representations, grounding abstract instructions in continuous motor behaviors rather than discrete tokens. A persistent weakness is brittle instruction parsing: minor rephrasing or typographical errors can cause large performance drops.

\subsection{Hierarchical Reasoning}
\label{sec:language:hierarchical}

Complex instructions require decomposition into executable subgoals. \textbf{Hi Robot}~\cite{shi2025hirobot} introduces a hierarchical VLA architecture where a high-level ``reasoner'' VLM processes the user's open-ended instruction and the current visual observation to generate a specific subgoal instruction. A low-level ``executor'' VLA then carries out the subgoal. This decomposition allows Hi Robot to follow complex instructions such as ``make me a sandwich'' by generating subgoals such as ``open the bread bag,'' ``pick up two slices,'' ``place cheese between them.''

\textbf{$\pi_{0.5}$}~\cite{black2025pi05} implements a similar hierarchy, with the high-level model additionally maintaining a task state representation that tracks progress through multi-step tasks. The high-level model can detect when a subgoal has failed and re-plan, providing robustness to execution errors.

Prior work established hierarchical planning principles adopted by VLAs: \textbf{SayCan}~\cite{ahn2022saycan} grounds LLM plans in affordance scores, \textbf{Code as Policies}~\cite{liang2023code} generates executable code, and \textbf{VoxPoser}~\cite{huang2023voxposer} produces 3D value maps. These affordance-grounding principles are incorporated in hierarchical VLAs such as Hi Robot and $\pi_{0.5}$.

The language channel between high-level planner and low-level executor determines control granularity. Natural language subgoals (\textbf{Hi Robot}~\cite{shi2025hirobot}) are flexible but may be ambiguous; alternatives include code-based specifications~\cite{liang2023code} (precise but brittle) and goal images (rich but expensive to generate). Natural language is currently favored for its compatibility with VLM backbones.

\subsection{Open-Ended Instruction Following}
\label{sec:language:openended}

Open-ended instruction following tests generalization beyond the training distribution. \textbf{RT-2}~\cite{brohan2023rt2} showed VLM pre-training enables following instructions involving novel objects absent from robot data. Related work includes \textbf{Manipulate-Anything}~\cite{duan2024manipulateanything} (detailed instruction following), \textbf{Interactive Language}~\cite{lynch2023interactive} (real-time streaming corrections), and \textbf{Chain-of-Thought Predictive Control}~\cite{xian2023chainoil} (reasoning-guided action generation).

VLM backbones bridge the gap between simple human commands (``fold the towel'') and the motor detail required for execution; \textbf{$\pi_0$}~\cite{black2024pi0} and \textbf{$\pi_{0.5}$}~\cite{black2025pi05} operate effectively with natural instructions. A persistent limitation is the lack of systematic evaluation: most results use hand-picked instruction sets---a small, manually curated collection of test commands rather than open-vocabulary or user-generated prompts---so reported success rates may overestimate real-world flexibility. Although VLMs can parse arbitrary natural language, current VLAs have only been validated on narrow instruction distributions and cannot reliably detect ambiguous or contradictory commands.

\subsection{Cross-Embodiment Transfer}
\label{sec:language:cross}

A key promise of VLAs is cross-embodiment generalization: a policy trained on data from multiple robots can be deployed on a new robot with minimal fine-tuning. The evidence is now substantial. \textbf{Octo}~\cite{octo2024} pre-trains on OXE data spanning 22 embodiments and transfers to unseen robots with a few hundred fine-tuning demonstrations, while \textbf{OpenVLA}~\cite{kim2024openvla} confirms that OXE pre-training improves performance even on embodiments absent from the pre-training set.

Cross-embodiment transfer is critical for dual-arm systems because bimanual demonstration data is scarce (as discussed in Section~\ref{sec:training:data}). A VLA pre-trained on diverse single-arm data can transfer visual and semantic representations to a bimanual system, even though the action space differs. The strongest evidence comes from \textbf{$\pi_0$}~\cite{black2024pi0}, which pre-trains on a mixture of single-arm and bimanual data and finds that single-arm data improves bimanual performance through shared visual representations.

The transition of VLA technology from research to industrial deployment is exemplified by \textbf{Xiaomi-Robotics-0}~\cite{xiaomi2025}, which trains on data from multiple Xiaomi robot platforms with real-time execution optimizations for consumer hardware.

Cross-embodiment transfer for bimanual systems requires handling different arm configurations. Approaches include action space normalization (\textbf{$\pi_0$}~\cite{black2024pi0} maps to a common end-effector format), embodiment-specific projection layers (\textbf{Octo}~\cite{octo2024}, extended to navigation and locomotion in \textbf{Octo~2.0}~\cite{team2024octo2}), and language-based action hierarchies (\textbf{RT-H}~\cite{rt_h2024}).

\subsection{Zero-Shot and Few-Shot Generalization}
\label{sec:language:zeroshot}

Zero-shot generalization (performing tasks with no task-specific training data) remains challenging for VLAs but is an active research frontier. \textbf{OK-Robot}~\cite{liu2024okrobot} combines a VLM for object detection with a pre-trained manipulation primitive to achieve zero-shot pick-and-place in novel environments. While not a full VLA, OK-Robot demonstrates the potential of VLM-based perception for zero-shot manipulation.

\textbf{Robot Utility Models}~\cite{etukuru2024robot} train VLAs as general-purpose ``utilities'' that can perform a broad range of manipulation tasks from language instructions, approaching zero-shot capability for common manipulation primitives. The remaining gap to true zero-shot bimanual manipulation is substantial, as bimanual coordination patterns are difficult to infer from language alone without motor experience.

A limitation of current cross-embodiment claims is that transfer is evaluated after fine-tuning without controlling for data quantity; rigorous ablations separating pre-training benefit from fine-tuning benefit are needed.

Table~\ref{tab:generalization} summarizes the generalization capabilities observed across VLA methods, distinguishing between environment, object, instruction, and embodiment generalization.

Tables~\ref{tab:generalization} and~\ref{tab:language} summarize the generalization and language grounding landscape across VLA methods. Language and cross-embodiment capabilities do not exist in isolation; they interact with visual representations, safety requirements, and deployment constraints, which we address in Section~\ref{sec:crosscutting}.

% ============================================================
% TABLE: GENERALIZATION
% ============================================================
\begin{table}[H]
%\centering
\caption{Generalization capabilities of VLA models across four dimensions. Strong/Partial/Weak indicate the degree of demonstrated generalization.}
\label{tab:generalization}

\begin{tabularx}{\textwidth}{LCCCC}
\toprule
\textbf{Method} & \textbf{Env.} & \textbf{Obj.} & \textbf{Instr.} & \textbf{Embod.} \\
\midrule
\textbf{RT-1 %
}~\cite{brohan2022rt1} & Weak & Weak & Weak & -- \\
\textbf{RT-2}~\cite{brohan2023rt2} & Partial & Strong & Strong & -- \\
\textbf{OpenVLA}~\cite{kim2024openvla} & Partial & Partial & Partial & Partial \\
\textbf{Octo}~\cite{octo2024} & Partial & Partial & Partial & Strong \\
\textbf{$\pi_0$}~\cite{black2024pi0} & Partial & Strong & Strong & Partial \\
\textbf{$\pi_{0.5}$}~\cite{black2025pi05} & Strong & Strong & Strong & Partial \\
\textbf{Hi Robot}~\cite{shi2025hirobot} & Partial & Partial & Strong & -- \\
\bottomrule
\end{tabularx}
\end{table}

% ============================================================
% TABLE 7: LANGUAGE GROUNDING
% ============================================================

\vspace{-12pt}

\begin{table}[H]
%\centering
\caption{Comparison %
 of language grounding and reasoning capabilities in VLA models. Novel instr. indicates generalization to unseen instruction phrasings. Novel obj. indicates generalization to unseen object categories.}
\label{tab:language}
\resizebox{\textwidth}{!}{%
\begin{tabular}{lcccccc}
\toprule
\textbf{Method} & \textbf{Hierarchical} & \textbf{Novel Instr.} & \textbf{Novel Obj.} & \textbf{Open-Ended} & \textbf{Cross-Embod.} & \textbf{Zero-Shot} \\
\midrule
\textbf{RT-1 %
}~\cite{brohan2022rt1} & -- & Limited & Limited & -- & -- & -- \\
\textbf{RT-2}~\cite{brohan2023rt2} & -- & \checkmark & \checkmark & Partial & -- & Partial \\
\textbf{OpenVLA}~\cite{kim2024openvla} & -- & \checkmark & \checkmark & -- & \checkmark & -- \\
\textbf{$\pi_0$}~\cite{black2024pi0} & -- & \checkmark & \checkmark & -- & \checkmark & -- \\
\textbf{$\pi_{0.5}$}~\cite{black2025pi05} & \checkmark & \checkmark & \checkmark & \checkmark & \checkmark & Partial \\
\textbf{Hi Robot}~\cite{shi2025hirobot} & \checkmark & \checkmark & \checkmark & \checkmark & -- & -- \\
\textbf{SayCan}~\cite{ahn2022saycan} & \checkmark & \checkmark & -- & \checkmark & -- & -- \\
\bottomrule
\end{tabular}}
\end{table}

% ============================================================
% 10. CROSS-CUTTING CONCERNS
% ============================================================
\section{Cross-Cutting Concerns}
\label{sec:crosscutting}

Several concerns cut across all VLA architectures and application domains. This section addresses visual representation learning, world models and future state prediction, safety, sim-to-real transfer, and human--robot interaction. Table~\ref{tab:crosscutting} compares these capabilities across representative VLA methods.

\begin{table}[H]
%\centering
\caption{Cross-cutting %
 capabilities of VLA models. Multi-view indicates support for multiple camera inputs. Safety indicates explicit safety mechanisms. Sim-to-Real indicates simulation-to-real transfer capability.}
\label{tab:crosscutting}
\resizebox{\textwidth}{!}{%
\begin{tabular}{lcccccc}
\toprule
\textbf{Method} & \textbf{Visual Encoder} & \textbf{Multi-View} & \textbf{Safety} & \textbf{Sim-to-Real} & \textbf{HRI} & \textbf{Proprioception} \\
\midrule
\textbf{RT-1 %
}~\cite{brohan2022rt1} & EfficientNet & -- & Basic & -- & -- & -- \\
\textbf{RT-2}~\cite{brohan2023rt2} & ViT (PaLI-X) & -- & Basic & -- & Partial & -- \\
\textbf{OpenVLA}~\cite{kim2024openvla} & DINOv2 + %
 SigLIP & -- & -- & -- & -- & -- \\
\textbf{$\pi_0$}~\cite{black2024pi0} & SigLIP & \checkmark & Rate limit & -- & -- & \checkmark \\
\textbf{$\pi_{0.5}$}~\cite{black2025pi05} & SigLIP & \checkmark & Multi-layer & -- & \checkmark & \checkmark \\
\textbf{$\pi_0^*$}~\cite{amin2025pi06} & SigLIP & \checkmark & Rate limit & -- & -- & \checkmark \\
\textbf{RDT-1B}~\cite{liu2024rdt1b} & SigLIP & \checkmark & Basic & -- & -- & \checkmark \\
\textbf{Octo}~\cite{octo2024} & Custom ViT & \checkmark & -- & \checkmark & -- & -- \\
\bottomrule
\end{tabular}}
\end{table}

\subsection{Visual Representation Learning}
\label{sec:crosscutting:visual}

The choice of visual encoder significantly impacts VLA performance. Three approaches dominate.

\textbf{Pre-trained VLM encoders} (\eg, SigLIP in PaLIGemma~\cite{beyer2024paligemma}, ViT in CLIP~\cite{radford2021clip}) provide rich semantic features pre-trained on web-scale data. These encoders excel at object recognition and scene understanding but may lack fine-grained spatial information needed for precise manipulation. Spatial structure may be as important as semantic richness for manipulation-oriented visual encoders: \textbf{Transporter Networks}~\cite{zeng2021transporter} achieve strong rearrangement performance using equivariant spatial representations learned without large-scale pre-training.

\textbf{Robot-specific visual representations} offer an alternative to generic VLM encoders. \textbf{R3M}~\cite{nair2022r3m}, pre-trained on robot video data using time-contrastive and language-aligned objectives, captures temporal dynamics and manipulation-relevant features that generic encoders miss. Offline data paired with crowd-sourced annotation also yields transferable representations~\cite{nair2022learning}. A large-scale comparison by \textbf{Cortex}~\cite{majumdar2024cortex} found that representations trained on diverse egocentric video outperform those from static image classification, while \textbf{SPA}~\cite{cheng2024spa} adds explicit 3D spatial-awareness to improve embodied policy learning. Several VLAs use R3M or similar robot-specific encoders alongside VLM encoders, processing images through both pathways.

\textbf{Multi-view fusion} is critical for bimanual manipulation, where a single camera may not capture both arms and the workspace simultaneously. Most bimanual VLAs (\textbf{$\pi_0$}~\cite{black2024pi0}, \textbf{RDT-1B}~\cite{liu2024rdt1b}) use multiple camera views (typically a wrist camera on each arm plus one or more third-person cameras) and fuse the resulting tokens within the Transformer backbone.

Multi-view fusion approaches range from early concatenation (\textbf{$\pi_0$}~\cite{black2024pi0}, \textbf{Octo}~\cite{octo2024}) to late fusion and learned view selection. Wrist cameras are indispensable for bimanual setups, capturing fine-grained contact information that third-person cameras miss. Multi-frame visual context~\cite{torne2026mem,jang2025contextvla,li2025cronusvla} extends temporal scope beyond the current observation. A limitation across all three approaches is that no principled method exists for selecting which visual encoder or fusion strategy best suits a given task; current practice relies on empirical trial-and-error, and the relative contribution of semantic versus spatial features to bimanual coordination remains poorly understood.

\textbf{Fine-tuning versus training from scratch.} As Table~\ref{tab:architectures} (Init.\ column) shows, nearly all modern VLAs fine-tune pre-trained visual encoders rather than training from scratch, because web-scale pre-training produces representations impractical to learn from robot data alone. Most systems freeze or lightly fine-tune the encoder and allocate training budget to the action head; for example, \textbf{$\pi_0$}~\cite{black2024pi0} fine-tunes SigLIP weights with a reduced learning rate, while \textbf{RT-1}~\cite{brohan2022rt1} adapts an ImageNet-pretrained EfficientNet-B3 via FiLM conditioning on 130K demonstrations. A notable exception is \textbf{Octo}~\cite{octo2024}, which trains from scratch on 800K episodes without any pre-trained encoder. The optimal balance between frozen and updated parameters remains an open question that depends on the available robot data and the domain gap.

\subsection{Safety}
\label{sec:crosscutting:safety}

Safety is paramount for bimanual systems operating near humans. VLA safety concerns include:

\textbf{Action bounds and rate limiting}: VLA outputs are typically clipped to safe action ranges and rate-limited to prevent high-velocity motions. Two-arm coordination introduces an additional collision-avoidance constraint between the arms that is not inherently captured by the VLA.

\textbf{Out-of-distribution detection}: Some VLAs use confidence-based filtering, halting when action head uncertainty (estimated from denoising variance or velocity field norms) exceeds a threshold.

\textbf{Collision avoidance}: Bimanual systems face self-collision risk between arms. Post hoc safety layers that project actions to collision-free trajectories add latency but provide guarantees; learning collision avoidance from demonstrations is an alternative that may not generalize to novel configurations.

\textbf{Human-in-the-loop correction}: For deployment in homes (\eg, \textbf{$\pi_{0.5}$}~\cite{black2025pi05}), the ability for humans to intervene and correct the robot is essential. VLAs that accept real-time language feedback can be redirected mid-task, providing a natural correction mechanism.

\subsection{Sim-to-Real Transfer}
\label{sec:crosscutting:simtoreal}

Simulation provides scalable, safe data generation, but the reality gap (differences between simulated and real visual appearances, physics, and dynamics) limits direct~transfer.

\textbf{SIMPLER}~\cite{li2024simpler} provides simulation environments calibrated to match real-world VLA evaluation setups, so that VLAs can be evaluated without physical hardware. The correlation between simulated and real-world success rates validates simulation as a development tool for VLAs.

The reality gap is acute for bimanual manipulation because contact dynamics (friction, deformation, compliance) are difficult to simulate accurately. Current bimanual VLAs (\textbf{$\pi_0$}~\cite{black2024pi0}, \textbf{$\pi_0^*$}~\cite{amin2025pi06}) rely primarily on real-world demonstrations and practice, with simulation playing a secondary role.

Strategies for closing this gap include \textbf{domain randomization} (varying visual and physical parameters), \textbf{system identification} (calibrating simulation to match the real robot), and generative approaches such as \textbf{Gen2Act}~\cite{bharadhwaj2024gen2act} (human video demonstrations) and \textbf{Track2Act}~\cite{bharadhwaj2024track2act} (point tracks from internet videos). \textbf{Hybrid training} (simulation pre-training plus real fine-tuning) works well for single-arm VLAs but remains underexplored for bimanual systems, where higher action dimensionality makes sim-to-real alignment~harder.

\subsection{World Models and Future State Prediction}
\label{sec:crosscutting:worldmodels}

A complementary approach to reactive VLA policies is to equip robots with \emph{world models} that predict future states, whether as visual frames, latent representations, or explicit physical quantities, before committing to actions. This ``predict-then-act'' approach offers several advantages for bimanual manipulation: it allows look-ahead planning for multi-step coordination, provides a mechanism for evaluating action consequences before execution, and can generate synthetic training data to alleviate the data scarcity problem.

The GigaBrain family illustrates the rapid maturation of world model-powered VLAs. \textbf{GigaBrain-0}~\cite{gigabrain0_2025} first used generative models to produce synthetic robot data. Its successor, \textbf{GigaBrain-0.5M$^*$}~\cite{gigabrain2025}, added RAMP (RL via World Model-conditioned Policy), yielding $\sim$30\% improvement on bimanual tasks. \textbf{GigaWorld-0}~\cite{gigaworld0_2025} completes the picture with a unified framework combining video generation and 3D modeling (Gaussian Splatting) as a scalable data engine.

A more radical approach formulates control as video generation. \textbf{Rhoda AI (DVA)}~\cite{rhoda2026dva} uses a causal video model pre-trained on web-scale video, with an inverse dynamics model translating predicted frames to actions (10--20 h of robot data). \textbf{VPP}~\cite{hu2025vpp} learns implicit inverse dynamics via video diffusion (+18.6\% on Calvin ABC-D). \textbf{ViPRA}~\cite{vipra2025} learns from actionless videos at 22\,Hz, and \textbf{Mimic-Video}~\cite{mimicvideo2025} achieves $10\times$ sample efficiency over standard VLAs.

Other approaches operate on optical flow and latent representations. \textbf{FOFPred}~\cite{fofpred2026} achieves 68.6\% on bimanual tasks via language-driven flow prediction. \textbf{V-JEPA 2}~\cite{vjepa2_2025} provides a self-supervised world model (1M+ hours of video; 65--80\% zero-shot success). \textbf{WorldVLA}~\cite{worldvla2025} jointly generates actions and future frames, \textbf{UP-VLA}~\cite{upvla2025} uses next-frame prediction for implicit physics, and \textbf{NVIDIA Cosmos}~\cite{cosmos2025} provides open world foundation models trained on 20M+ h %
 of data.

The predict-then-act approach offers four advantages: \emph{data efficiency} (DVA requires only 10--20 h of robot data; Mimic-Video achieves $10\times$ sample efficiency), \emph{interpretability} (predicted frames can be visualized), \emph{planning} (look-ahead evaluation before committing), and \emph{synthetic data generation} (GigaWorld-0 and Cosmos produce unlimited training data).

However, fundamental limitations remain. \emph{Prediction accuracy degrades} as compounding errors in autoregressive video generation make long-horizon forecasts unreliable, especially for contact-rich bimanual tasks with rapid state changes. \emph{Inverse dynamics accuracy} suffers from the additional error of translating predicted video back to precise actions. \emph{Computational cost} conflicts with real-time control budgets (ViPRA runs at only 22\,Hz). Video models can \emph{hallucinate} plausible but incorrect states after occlusions, and current world models lack \emph{haptic grounding} for force and tactile signals.

\subsection{Human--Robot Interaction}
\label{sec:crosscutting:hri}

VLAs facilitate more natural human--robot interaction through language. A user can instruct a bimanual robot in natural language, observe its behavior, and provide corrections or new instructions in real time.

The most complete HRI demonstration to date comes from \textbf{$\pi_{0.5}$}~\cite{black2025pi05}, where users gave verbal instructions to a bimanual robot in home environments, the robot executed them, and users could redirect it as needed. The hierarchical architecture allows the robot to ask clarifying questions through the high-level VLM when instructions are ambiguous.

\textbf{PaLM-E}~\cite{driess2023palme} and \textbf{PIVOT}~\cite{nasiriany2024pivot} show that VLMs can engage in dialogue about the physical world and elicit actionable knowledge through visual prompting. However, HRI evaluation for VLAs remains qualitative; no standardized metrics exist for interaction quality or correction latency in VLA-based bimanual systems.

\subsection{Scalability and Deployment}
\label{sec:crosscutting:scalability}

Deploying VLAs on bimanual systems in real-world settings introduces engineering challenges beyond model performance. \textbf{Compute requirements} are substantial: a 3B-parameter VLA running flow matching with $K = 10$ steps requires a high-end GPU (A100 or equivalent) for real-time bimanual control. Edge deployment on embedded GPUs is not yet practical for full-size VLAs, motivating the efficient architectures discussed in Section~\ref{sec:architectures:hybrid}.

\textbf{Communication latency} between the VLA compute server and the robot controller adds to the end-to-end control delay. With dual-arm systems running at $50\,\text{Hz}$, the total loop delay (image capture, network transfer, VLA inference, action transfer, motor execution) must remain below $20\,\text{ms}$ per step. Action chunking mitigates this by amortizing the VLA inference over $H$ steps but introduces a minimum reaction latency of one chunk period.

\textls[-15]{A gap remains between research demonstrations and reliable deployment: most results use controlled laboratory conditions, and long-term reliability metrics are absent. Reproducibility is limited by leading systems' reliance on proprietary data. \textbf{Xiaomi-Robotics-0}~\cite{xiaomi2025} represents industrial VLA deployment with custom hardware accelerators for low-latency bimanual control. \textbf{TidyBot}~\cite{wu2023tidybot} demonstrates LLM-powered household robotics, and \textbf{ManiWAV}~\cite{liu2025maniwav} shows that auditory feedback complements vision for contact-rich~tasks.}

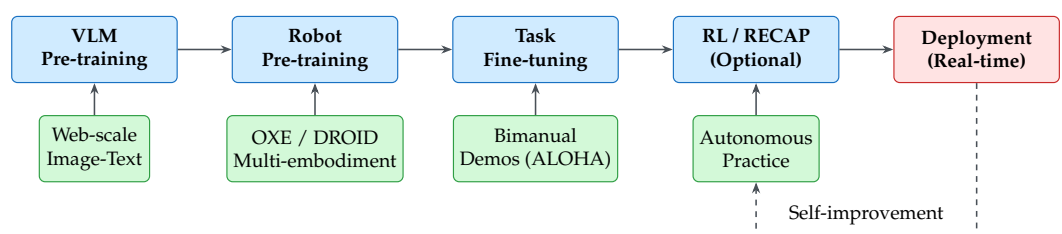
\begin{figure}[H]
\centering
\resizebox{\textwidth}{!}{%
\begin{tikzpicture}[
    every node/.style={font=\scriptsize, align=center},
    block/.style={rectangle, draw, rounded corners=3pt, minimum width=20mm, minimum height=12mm, font=\scriptsize, line width=0.6pt},
    phase/.style={block, draw=cBlue, fill=cBlueFill, minimum width=30mm, font=\normalsize\bfseries},
    data/.style={block, draw=cGreen, fill=cGreenFill, font=\normalsize},
    arr/.style={-{Stealth[length=2mm]}, thick, draw=cGray},
]
% Phase 1: VLM Pre-training
\node[phase] (vlm) at (0, 0) {VLM\\Pre-training};
\node[data] (web) at (0, -1.8) {Web-scale\\Image-Text};
\draw[arr] (web) -- (vlm);

% Phase 2: Robot Pre-training
\node[phase] (robot_pt) at (4, 0) {Robot\\Pre-training};
\node[data] (oxe) at (4, -1.8) {OXE / DROID\\Multi-embodiment};
\draw[arr] (vlm) -- (robot_pt);
\draw[arr] (oxe) -- (robot_pt);

% Phase 3: Fine-tuning
\node[phase] (ft) at (8, 0) {Task\\Fine-tuning};
\node[data] (demos) at (8, -1.8) {Bimanual\\Demos (ALOHA)};
\draw[arr] (robot_pt) -- (ft);
\draw[arr] (demos) -- (ft);

% Phase 4: RL (optional)
\node[phase] (rl) at (12, 0) {RL / RECAP\\(Optional)};
\node[data] (auto) at (12, -1.8) {Autonomous\\Practice};
\draw[arr] (ft) -- (rl);
\draw[arr] (auto) -- (rl);

% Phase 5: Deployment
\node[phase, draw=cRed, fill=cRedFill] (deploy) at (16, 0) {Deployment\\(Real-time)};
\draw[arr] (rl) -- (deploy);

% Feedback loop
\draw[arr, dashed] (deploy.south) -- ++(0, -2.8) -| (auto.south);
\node[font=\normalsize, fill=white] at (14, -3.0) {Self-improvement};

\end{tikzpicture}%
}% end resizebox
\caption{The %
 VLA training and deployment pipeline for bimanual manipulation. Training proceeds through four phases: VLM pre-training on web data, robot pre-training on cross-embodiment datasets, task fine-tuning on bimanual demonstrations, and optional reinforcement learning from autonomous practice. The RL phase creates a self-improvement loop where the deployed policy generates additional training data.}
\label{fig:pipeline}
\end{figure}

Figure~\ref{fig:pipeline} summarizes the complete VLA training and deployment pipeline that ties together the architectural choices (Section~\ref{sec:architectures}), training recipes (Section~\ref{sec:training}), and deployment considerations discussed above. The interplay among these architectural, training, and deployment considerations shapes the current state of the art, which we synthesize next.

% ============================================================
% TABLE 8: CROSS-CUTTING
% ============================================================

% ============================================================
% FIGURE 4: VLA PIPELINE
% ============================================================

% ============================================================
% FIGURE 5: PERFORMANCE EVOLUTION
% ============================================================

% ============================================================
% 11. DISCUSSION AND CONCLUSION
% ============================================================
\section{Discussion and Conclusions}
\label{sec:discussion}

\subsection{State-of-the-Art Performance}
\label{sec:discussion:sota}

\begin{figure}[H]
\centering
\begin{tikzpicture}
\begin{axis}[
    width=\textwidth,
    height=6cm,
    xlabel={Year},
    ylabel={Success Rate (\%)},
    xmin=2022.5, xmax=2025.5,
    ymin=0, ymax=100,
    xtick={2023, 2024, 2025},
    xticklabel style={/pgf/number format/1000 sep={}},
    ytick={0, 20, 40, 60, 80, 100},
    legend style={at={(0.47,0.32)}, anchor=north west, font=\small, row sep=-2pt},
    grid=major,
    grid style={dashed, gray!30},
    every axis plot/.append style={thick, mark size=2.5pt},
]
% Bimanual folding task
\addplot[color=cRed, mark=*] coordinates {
    (2023, 30)
    (2024, 80)
    (2025, 95)
};
\addlegendentry{Bimanual folding}

% Bimanual assembly
\addplot[color=cBlue, mark=square*] coordinates {
    (2023, 20)
    (2024, 75)
    (2025, 90)
};
\addlegendentry{Bimanual assembly}

% Single-arm (reference)
\addplot[color=cGreen, mark=triangle*] coordinates {
    (2023, 70)
    (2024, 85)
    (2025, 95)
};
\addlegendentry{Single-arm manip.}

% Annotations
\node[font=\tiny, anchor=south] at (axis cs:2023, 30) {ACT};
\node[font=\tiny, anchor=south] at (axis cs:2024, 84) {$\pi_0$};
\node[font=\tiny, anchor=south west] at (axis cs:2025, 84) {$\pi_0^*$};

\end{axis}
\end{tikzpicture}
\caption{Approximate %
 evolution of VLA performance on bimanual manipulation tasks (2023--2025). Values are approximate trend values synthesized by the authors from reported results across different evaluation setups and task definitions; they illustrate general trends rather than exact comparable benchmarks. Bimanual task success rates have improved dramatically, from $\sim$30\% with early methods such as ACT~\cite{zhao2023aloha} to $>$90\% with $\pi_0^*$~\cite{amin2025pi06}. The gap between bimanual and single-arm performance has narrowed but persists for the most dexterous tasks.}
\label{fig:performance_evolution}
\end{figure}
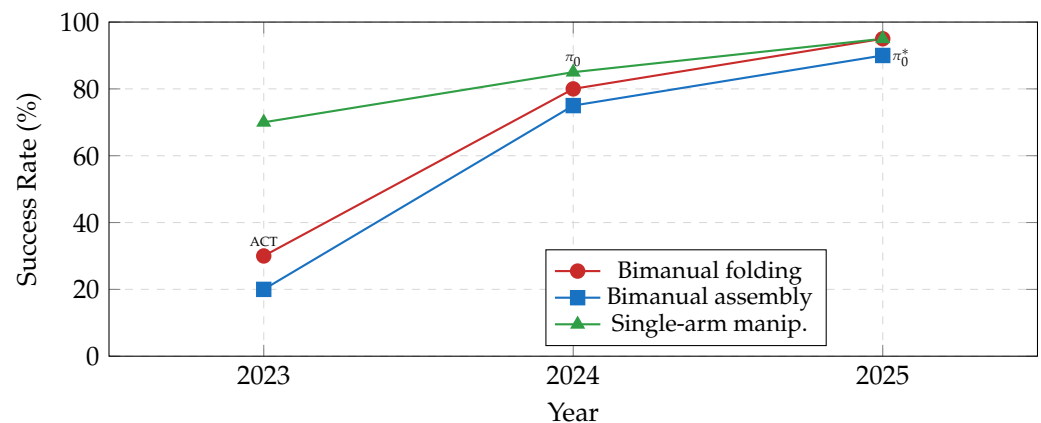

Synthesizing the results presented across Sections~\ref{sec:architectures}--\ref{sec:crosscutting}, in our assessment, the state of the art in VLA-based bimanual manipulation and unmanned aerial robotics can be characterized along several dimensions. Figure~\ref{fig:performance_evolution} charts the rapid progress from 2023 to 2025: bimanual folding success rose from $\sim$30\% (ACT) to over 90\% ($\pi_0^*$), narrowing the gap with single-arm performance.

\textbf{Architecture.} As reflected in Tables~\ref{tab:benchmark_performance} and~\ref{tab:bimanual_tasks}, flow-based VLAs, led by \textbf{$\pi_0$}~\cite{black2024pi0} and its successors \textbf{$\pi_{0.5}$}~\cite{black2025pi05} and \textbf{$\pi_0^*$}~\cite{amin2025pi06}, currently achieve the strongest bimanual manipulation performance. The flow-matching action head generates smooth, high-dimensional action chunks without quantization, and the iterative denoising process captures the multi-modal coordination patterns inherent in bimanual tasks. Diffusion-based models, particularly \textbf{RDT-1B}~\cite{liu2024rdt1b}, demonstrate that scale improves performance but trail flow-based approaches in inference efficiency. Autoregressive VLAs (\textbf{OpenVLA}~\cite{kim2024openvla}, \textbf{RT-2}~\cite{brohan2023rt2}) provide the simplest integration with VLM pre-training but are limited by discretization for bimanual action~spaces.

\textbf{Training.} The three-stage recipe (VLM initialization, cross-embodiment pre-training, task-specific fine-tuning) is now standard. The addition of reinforcement learning from autonomous practice (\textbf{RECAP}~\cite{amin2025pi06}) represents the most impactful recent advance; it allows VLAs to surpass demonstration quality by 10--40\% on bimanual tasks. Co-training with diverse data during fine-tuning consistently improves performance and robustness.

\textbf{Action representation.} Action chunking with $H = 50$ steps is the dominant choice for bimanual VLAs, providing the temporal coherence needed for coordinated two-arm motions. Learned action tokenization (\textbf{FAST}~\cite{pertsch2025fast}) narrows the gap between autoregressive and continuous approaches. Real-time execution techniques (\textbf{RTC}~\cite{black2025rtc}) enable reactive bimanual control despite the computational cost of large VLA models.

\textbf{Generalization.} Hierarchical VLAs (\textbf{$\pi_{0.5}$}~\cite{black2025pi05}, \textbf{Hi Robot}~\cite{shi2025hirobot}) demonstrate the strongest generalization to novel environments and open-ended instructions. Cross-embodiment pre-training on OXE data provides a foundation for transfer, though bimanual-specific skills require task-specific fine-tuning.

\textbf{Efficiency.} The computational cost of VLA inference remains a concern for bimanual real-time control. Table~ %
\ref{tab:efficiency} summarizes the efficiency characteristics of representative methods. Flow-based models with $K = 10$ denoising steps achieve the best latency--quality tradeoff, while efficient architectures (\textbf{TinyVLA}~\cite{wen2024tinyvla}, \textbf{MiniVLA}~\cite{belkhale2024minivla}) sacrifice some capability for deployment on resource-constrained hardware.

\textbf{Memory.} Memory-augmented VLAs represent the most significant recent advance for long-horizon bimanual tasks. \textbf{MEM}~\cite{torne2026mem}, integrated into the $\pi_{0.6}$ VLA, enables tasks spanning up to 15 min by combining video-based short-horizon memory with language-based long-horizon memory. Concurrent approaches~\cite{shi2025memoryvla,fang2025sam2act,jang2025contextvla} explore complementary designs, from perceptual--cognitive memory banks to amortized multi-frame context. The common finding is that different time scales demand different memory representations.

\textbf{World Models.} World model-powered VLAs~\cite{gigabrain0_2025,gigabrain2025} and direct video-action models~\cite{rhoda2026dva,hu2025vpp} represent a shift from purely reactive policies to predictive ones. The GigaBrain family demonstrated that world model-generated data improves bimanual task performance by $\sim$30\%, and web-scale video pre-training transfers to robot control with 10--20 h of task-specific data. However, none of these approaches have been evaluated on standardized bimanual benchmarks, limiting direct comparison.

\textbf{Unmanned Aerial Robotics.} VLA adoption for drones lags manipulation by $\sim$2~years. Strong individual components exist (RL flight policies~\cite{kaufmann2023champion}, VLN benchmarks~\cite{liu2023aerialvln}, LLM-based planning~\cite{shah2023navigation}), but integrated end-to-end VLA systems for physical drones remain rare. The architectural innovations proven for bimanual manipulation are directly applicable (see directions 11--13 below).

\textbf{Industrial VLA Deployment.} The transition from research to product-level systems has accelerated in 2025--2026, with several companies deploying VLA-based robots commercially (Table~\ref{tab:industrial_vla}). Three architectural patterns have emerged from the industry that differ from the research approach.

First, \emph{dual-system (S1/S2) architectures} separate slow reasoning from fast action. \textbf{Gemini Robotics}~\cite{gemini_robotics2025} runs a distilled VLM in the cloud ($<$160\,ms query latency) paired with a local action decoder achieving 50\,Hz control and more than doubles performance on a broad generalization benchmark compared to other VLAs. \textbf{GR00T N1}~\cite{groot_n1_2025} pairs a 1.34B-parameter VLM (System~2, 10\,Hz) with a diffusion Transformer action head (System~1, 120\,Hz). \textbf{Helix}~\cite{figure_helix2025} from Figure~AI uses a 7B VLM at 7\,Hz and an 80M-parameter action model at 200\,Hz, controlling 35 degrees of freedom on embedded GPUs without cloud dependency. This pattern resolves the latency--capability tradeoff that limits monolithic VLAs (Direction~3).

Second, \emph{video-as-action models} bypass direct action regression entirely. \textbf{Rhoda AI}'s DVA~\cite{rhoda2026dva} pre-trains a causal video model on web-scale video, predicts future frames conditioned on the current scene, and extracts actions via a learned inverse dynamics model. \textbf{1X Technologies}~\cite{1x_worldmodel2025} takes a similar approach with a 14B-parameter text-conditioned diffusion world model trained on 900 h of egocentric human video plus 70 h of robot data. Both systems achieve one-shot or few-shot task adaptation from human demonstrations injected into the context window, without retraining.

Third, \emph{continuous autonomous improvement} has reached production scale. \textbf{Dyna Robotics}~\cite{dyna2025} deploys a dual-arm foundation model (DYNA-1) with a proprietary reward model that enables self-supervised error recovery. DYNA-1 reports a 99.4\% success rate over 24+ h %
 of continuous autonomous operation in commercial settings. This validates the RECAP paradigm (Section~\ref{sec:training:rl}) at industrial scale. We note that industrial performance figures are self-reported under company-defined conditions and await \mbox{independent~replication.}

Novel data collection strategies are also emerging. \textbf{Sunday Robotics}~\cite{sunday_act1_2026}, founded by the creators of ALOHA~\cite{zhao2023aloha} and Diffusion Policy~\cite{chi2023diffusion}, trains its ACT-1 foundation model on \emph{zero robot data}, using low-cost Skill Capture Gloves to collect human demonstration episodes across diverse homes, with a learned Skill Transform layer adapting human kinematics to robot morphology. Specific scale figures (glove cost, episode counts, number of homes) are based on company announcements subsequent to the initial publication and should be treated as projected targets. \textbf{Covariant}'s RFM-1~\cite{covariant_rfm1_2024}, an 8B-parameter autoregressive world model trained on millions of real deployment interactions, was designed to extend Covariant's legacy fleet (which achieved 99\%+ warehouse picking precision across hundreds of sites) with generalized reasoning capabilities; RFM-1 itself has not been independently deployed at that scale. \textbf{AgiBot World}~\cite{bu2025agibot} contributes 1M+ real robot trajectories across 217 tasks, with its GO-1 generalist policy outperforming Open X-Embodiment baselines by 30\%.

Hardware-focused companies are also integrating VLA-class models. \textbf{Boston Dynamics} and Toyota Research Institute jointly developed a Large Behavior Model (LBM)~\cite{bostondynamics2025lbm} for the Atlas humanoid that controls the entire robot (hands and feet) through a single whole-body policy for packing, sorting, and organizing tasks. Atlas fleets are scheduled to ship to Hyundai and Google DeepMind in 2026, with Google's Gemini Robotics~\cite{gemini_robotics2025} being integrated for enhanced cognitive capabilities.

\begin{figure}[H]
\centering
\includegraphics[width=\textwidth]{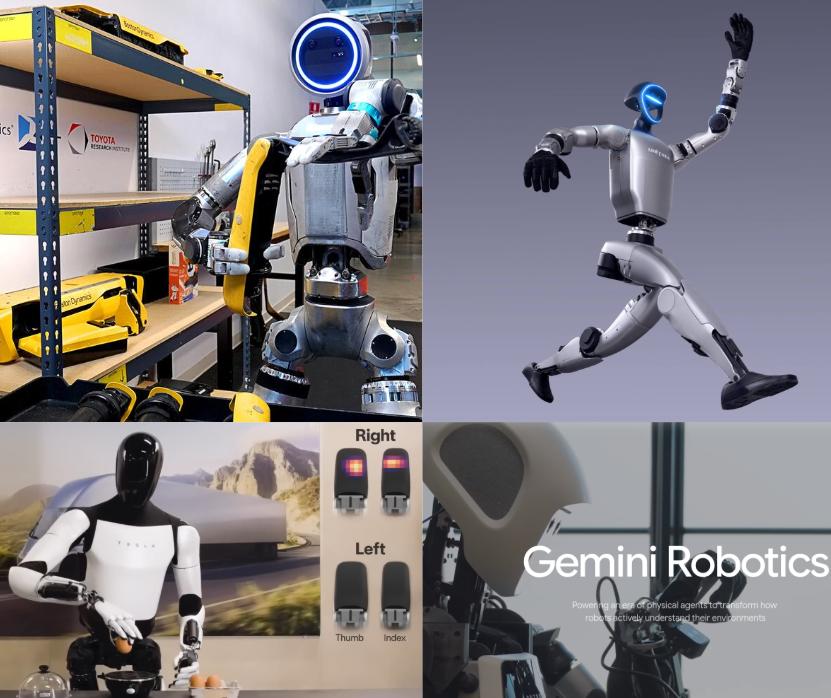}
\caption{Industrial %
 VLA-powered humanoid robot systems. (\textbf{a}, %
 top left)~Boston Dynamics Atlas with TRI Large Behavior Model performing warehouse manipulation. Reprinted with permission from Ref.~\cite{bostondynamics2025lbm}. Copyright 2025, Boston Dynamics and Toyota Research Institute. (\textbf{b}, top right)~Unitree humanoid executing dynamic whole-body control (image courtesy of Unitree Robotics). (\textbf{c}, bottom left)~Tesla Optimus humanoid with dexterous hands for general-purpose manipulation. Reprinted with permission from Ref.~\cite{tesla_optimus2025}. Copyright 2025, Tesla AI. (\textbf{d}, bottom right)~Google DeepMind Gemini Robotics, which integrates actions as a native Gemini modality for dexterous manipulation and general-purpose robot control. Reprinted with permission from Ref.~\cite{gemini_robotics2025}. Copyright 2025, Google DeepMind.}
\label{fig:industrial_robots}
\end{figure}

Open-source efforts from \textbf{NVIDIA}~\cite{groot_n1_2025} (GR00T~N1 weights and training data), \textbf{Xiaomi}~\cite{xiaomi2025} (Xiaomi-Robotics-0, LIBERO SOTA at 98.7\%), and \textbf{AgiBot}~\cite{bu2025agibot} are accelerating community progress. Figure~\ref{fig:industrial_robots} shows representative systems from this industrial wave.

% ============================================================
% FIGURE: INDUSTRIAL ROBOT SYSTEMS
% ============================================================

% ============================================================
% TABLE: EFFICIENCY
% ============================================================
\begin{table}[H]
\centering
\caption{Efficiency comparison of VLA models for bimanual deployment. GPU indicates the minimum GPU for real-time control.}
\label{tab:efficiency}

\begin{tabularx}{\textwidth}{LCCC}
\toprule
\textbf{Method} & \textbf{Params} & \textbf{Latency} & \textbf{Min. GPU} \\
\midrule
\textbf{RT-2 %
}~\cite{brohan2023rt2} & 55B & $\sim$1\,s & TPU v4 \\
\textbf{OpenVLA}~\cite{kim2024openvla} & 7B & $\sim$150\,ms & A100 \\
\textbf{$\pi_0$}~\cite{black2024pi0} & 3B & $\sim$70\,ms & A100 \\
\textbf{RDT-1B}~\cite{liu2024rdt1b} & 1.2B & $\sim$150\,ms & A6000 \\
\textbf{TinyVLA}~\cite{wen2024tinyvla} & 1B & $\sim$40\,ms & RTX 4090 \\
\textbf{MiniVLA}~\cite{belkhale2024minivla} & 300M & $\sim$25\,ms & RTX 3090 \\
\textbf{FAST}~\cite{pertsch2025fast} & 7B & $\sim$80\,ms & A100 \\
\bottomrule
\end{tabularx}
\end{table}
\vspace{-12pt}

% ============================================================
% TABLE: INDUSTRIAL VLA SYSTEMS
% ============================================================
\begin{table}[H]
%\centering
\caption{Industrial VLA systems (2024--2026). Architecture indicates the action generation approach. Deployment column indicates the current operational status. Success rates are self-reported under company-defined conditions.}

\label{tab:industrial_vla}
\begin{adjustwidth}{-\extralength}{0cm}
\fontsize{7}{7}\selectfont

\begin{tabularx}{\fulllength}{llllll}
\toprule
\textbf{System} & \textbf{Organization} & \textbf{Architecture} & \textbf{Key Innovation} & \textbf{Deployment} & \textbf{Year} \\
\midrule
\multicolumn{6}{l}{\textit{Dual-System (S1/S2) Architectures %
}} \\
Gemini Robotics~\cite{gemini_robotics2025} & Google DeepMind & VLM + action decoder & Actions as native Gemini modality & Partner testing & 2025 \\
GR00T N1~\cite{groot_n1_2025} & NVIDIA & VLM (10\,Hz) + DiT (120\,Hz) & Open-source; neural trajectory augment. & Research & 2025 \\
Helix~\cite{figure_helix2025} & Figure AI & VLM (7\,Hz) + action (200\,Hz) & 35-DOF on embedded GPU & BMW partnership %
 & 2025 \\
\midrule
\multicolumn{6}{l}{\textit{Video-as-Action / World Models}} \\
DVA~\cite{rhoda2026dva} & Rhoda AI & Causal video $\rightarrow$ inv.~dynamics & Web-scale video pre-training & Industrial pilots & 2026 \\
1XWM~\cite{1x_worldmodel2025} & 1X Technologies & Diffusion WM $\rightarrow$ IDM & 900 h human video + 70 h robot & Development & 2026 \\
RFM-1~\cite{covariant_rfm1_2024} & Covariant & 8B AR world model & Generalized reasoning from fleet data & Development & 2024 \\
\midrule
\multicolumn{6}{l}{\textit{Continuous Autonomous Improvement}} \\
DYNA-1~\cite{dyna2025} & Dyna Robotics & FM + proprietary RM & 99.4\% success, 24 h autonomy & Commercial sites & 2025 \\
$\pi_0^*$~\cite{amin2025pi06} & Physical Intelligence & FM + RECAP (RL) & 10--40\% over demo baseline & Research & 2025 \\
\midrule
\multicolumn{6}{l}{\textit{Novel Data Collection}} \\
ACT-1~\cite{sunday_act1_2026} & Sunday Robotics & Zero robot data; glove demos & Human demos via gloves (scale TBD) & Beta 2026 & 2026 \\
AgiBot World~\cite{bu2025agibot} & AgiBot & Latent action repr. & 1M+ trajectories; 30\% over OXE & Shipping at scale & 2025 \\
\midrule
\multicolumn{6}{l}{\textit{Open-Source VLAs}} \\
Xiaomi-Robotics-0~\cite{xiaomi2025} & Xiaomi & MoT + DiT (4.7B) & LIBERO 98.7\% SOTA & Open-source & 2025 \\
GR00T N1~\cite{groot_n1_2025} & NVIDIA & VLM + DiT (2.2B) & Weights + data released & Open-source & 2025 \\
\bottomrule
\end{tabularx}

%\centering %% If there is a figure in wide page, please release command \centering

\end{adjustwidth}
\end{table}

The field has also converged on several \textbf{long-standing challenges} that remain open despite recent progress:
\begin{itemize}
    \item \textbf{Distribution shift}: VLA policies still degrade when encountering out-of-distribution observations, especially for bimanual tasks where object configurations have high variability.
    \item \textbf{Contact modeling}: precise force control during bimanual contact is not addressed by current position-space VLAs.
    \item \textbf{Evaluation standardization}: the lack of common bimanual benchmarks prevents fair comparison across methods.
    \item \textbf{Data scarcity}: high-quality bimanual demonstrations remain expensive to collect, limiting the scale of bimanual VLA training.
    \item \textbf{Temporal credit assignment}: for long-horizon bimanual tasks, determining which actions contributed to success or failure is difficult, hindering RL-based improvement.
\end{itemize}

\subsection{Summary and Discussion}
\label{sec:discussion:summary}

We organize our conclusions into fourteen findings, grouped thematically: (1)--(3) cover core VLA design choices, (4)--(6) cover training and data strategies, (7)--(8) identify current limitations, (9)--(10) describe emergent capabilities, and (11)--(14) address cross-domain transfer and deployment. Each finding states a conclusion from the evidence reviewed, followed by its implications.

In our view, the analysis yields the following key findings:

\textbf{(1) Flow matching is the current best action generation mechanism for bimanual VLAs.} We find that the combination of continuous action generation, efficient sampling ($K = 10$ steps), and long action chunks makes flow matching uniquely suited to the high-dimensional, temporally correlated action spaces of bimanual manipulation. Diffusion models offer similar expressiveness but at higher computational cost.

\textbf{(2) VLM pre-training provides critical semantic grounding for bimanual tasks.} In our analysis, VLAs that inherit web-scale knowledge from VLM backbones consistently outperform architectures trained from scratch on robot data alone. The VLM's understanding of objects, spatial relationships, and task semantics transfers directly to manipulation, reducing the amount of robot-specific data needed.

\textbf{(3) Action chunking is essential for bimanual coordination.} We observe that single-step action prediction cannot capture the coordinated motion patterns of two arms working in concert. Chunks of $H = 50$ steps at $50\,\text{Hz}$ (1 s of motion) provide sufficient temporal context for most bimanual primitives, including folding, handovers, and assembly.

\textbf{(4) Reinforcement learning from autonomous practice is, in our assessment, the single most impactful recent advance for bimanual VLAs.} RECAP~\cite{amin2025pi06} showed that VLAs can self-improve by practicing autonomously and learning from success/failure signals. This matters especially for bimanual tasks where demonstration data is expensive to collect and expert performance is difficult to achieve via teleoperation.

\textbf{(5) Hierarchical architectures enable long-horizon bimanual tasks.} We find that flat VLA policies struggle with tasks requiring more than a few steps of bimanual coordination. Hierarchical decomposition (high-level VLM reasoning plus low-level VLA execution) extends the effective planning horizon from seconds to minutes.

\textbf{(6) Data diversity matters more than data quantity for generalization, but at the cost of per-task precision.} VLAs pre-trained on diverse cross-embodiment data generalize better than those trained on larger quantities of homogeneous data. However, \textbf{$\pi_0$}~\cite{black2024pi0} pre-trains on hundreds of tasks yet still requires 50--200 task-specific demonstrations for strong bimanual performance, and \textbf{RT-1}~\cite{brohan2022rt1} achieved 97\% on seen tasks but only 76\% on unseen ones. A two-stage recipe has thus become standard: broad pre-training for coverage, then narrow fine-tuning for precision.

\textbf{(7) The latency--reactivity tradeoff remains a fundamental challenge.} We note that large VLA models incur significant inference latency, conflicting with the need for reactive bimanual control. A 3B-parameter flow-based VLA requires ${\sim}12\,\text{GB}$ of GPU VRAM and an A100-class GPU for real-time inference; edge deployment on embedded GPUs (Jetson-class) is not yet practical at this scale. Techniques such as RTC~\cite{black2025rtc}, TinyVLA~\cite{wen2024tinyvla}, and FAST~\cite{pertsch2025fast} mitigate inference latency but do not fully resolve it.

\textbf{(8) Bimanual benchmarks are insufficient.} We consider this the most pressing infrastructure gap in the field. Most VLA evaluation occurs on single-arm tasks or bespoke bimanual setups that vary across papers. Without standardized bimanual benchmarks, comparing methods fairly and tracking progress systematically is not possible.

\textbf{(9) Pre-training on web data transfers to bimanual tasks.} We observe that the semantic knowledge encoded in VLM backbones (object affordances, material properties, spatial reasoning) directly benefits bimanual manipulation, even though web data contains no robot actions. This transfer is most evident in language grounding (understanding what ``fold'' or ``stack'' means) and visual scene understanding (identifying object parts and configurations).

\textbf{(10) Bimanual coordination emerges from joint prediction.} We find this result surprising: VLAs that predict both arms' actions jointly in a single action chunk learn coordination patterns implicitly from data, without explicit coordination mechanisms. This emergent coordination is strongest with flow-based and diffusion-based models that generate the full bimanual action in a single denoising process.

\textbf{(11) VLA architectures are cross-embodiment, with aerial applications lagging by $\sim$2 years.} We observe that the same VLM backbones, action generation mechanisms, and training recipes that power bimanual manipulation are being adapted for unmanned aerial robotics. As of early 2026, the aerial VLA field is at the stage manipulation reached in 2022--2023: strong individual components exist but integrated end-to-end systems remain nascent. High-fidelity simulators, accessible hardware, and cross-embodiment pre-training provide the ingredients for rapid convergence.

\textbf{(12) Dual-system architectures are the industry consensus for product-level VLAs.} We find that Google (Gemini Robotics~\cite{gemini_robotics2025}), NVIDIA (GR00T N1~\cite{groot_n1_2025}), and Figure AI (Helix~\cite{figure_helix2025}) all independently converged on separating a slow reasoning module (${\leq}10\,\text{Hz}$) from a fast action module (${\geq}100\,\text{Hz}$). This pattern resolves the latency--capability tradeoff that monolithic VLAs face: the reasoning module provides semantic understanding and task decomposition, while the action module generates smooth, high-frequency motor commands. Research VLAs that adopt this pattern will be better positioned for deployment.

\textbf{(13) Production reliability requires continuous self-improvement, not just better demonstrations.} In our view, the most reliable deployed systems, Dyna's DYNA-1~\cite{dyna2025} (99.4\% over 24 h) and Covariant's RFM-1~\cite{covariant_rfm1_2024} (99\%+ precision), achieve their performance through continuous RL loops where every deployment interaction feeds back into training, not through larger demonstration datasets alone. This validates the RECAP approach~\cite{amin2025pi06} at industrial scale and suggests that the path to product-level VLAs runs through autonomous improvement infrastructure.

\textbf{(14) Video prediction is gaining traction as an alternative to direct action regression.} We note that Rhoda AI's FutureVision~\cite{rhoda2026dva} and 1X Technologies' world model~\cite{1x_worldmodel2025} generate future video frames first and extract actions via inverse dynamics, exploiting web-scale video pre-training that contains orders of magnitude more data than robot demonstration datasets. This approach allows one-shot task adaptation from human demonstrations without retraining, though inference latency and physics fidelity remain open challenges.

These findings point to a converging design pattern: a pre-trained VLM backbone for semantic grounding, paired with a continuous (preferably flow-based) action head that generates multi-dimensional action chunks. Training follows a two-stage recipe---diverse pre-training then task-specific fine-tuning---increasingly augmented by RL from autonomous practice. The same pattern appears in both bimanual manipulation and aerial robotics, supporting the view that VLAs generalize across embodiments.

Translating this pattern into deployed systems, however, requires addressing practical constraints that cut across all findings: environmental variability degrades laboratory-trained policies, communication latency between GPU servers and robot controllers adds to the control loop, long-term reliability over hours of operation remains undemonstrated, and the energy cost of continuous GPU inference presents barriers to fleet-level scaling. The most pressing open questions are scalable evaluation (Finding~8), the diversity--precision tradeoff (Finding~6), and the gap between research prototypes and production systems with sustained reliability (Findings~12--13).

\subsection{Research Directions}
\label{sec:discussion:directions}

Despite rapid progress, several fundamental challenges remain. We consider the following research directions most pressing:

\textbf{(1) Standardized Bimanual Benchmarks:} The field urgently needs standardized simulation and real-world benchmarks for bimanual manipulation, analogous to LIBERO for single-arm tasks. Such benchmarks should cover the full spectrum of coordination types (independent, loosely coupled, tightly coupled), object categories (rigid, articulated, deformable), and task horizons (single-step to multi-minute). Without standardized evaluation, comparing bimanual VLA methods remains unreliable.

\textbf{(2) Dexterous, Force-Aware, and Multi-Modal Manipulation:} Current VLA systems use parallel-jaw grippers and rely solely on visual observations, limiting both dexterity and contact awareness. Extending VLAs to multi-fingered hands would unlock tasks such as in-hand reorientation, but the action space (two 16-DOF hands plus two 7-DOF arms) exceeds 40 dimensions per step, posing extreme challenges for action generation. Simultaneously, incorporating force/torque feedback and tactile sensing (\eg, GelSight) into VLA observations is essential for contact-rich tasks such as tightening screws, snapping parts, and kneading dough. Auditory signals can further complement vision for detecting task-relevant events such as clicks and snaps. Jointly addressing dexterity, force awareness, and multi-modal sensing is necessary to move bimanual VLAs beyond the current pick-and-place regime.

\textbf{(3) Real-Time Reactive Control:} Despite advances in RTC~\cite{black2025rtc} and efficient architectures~\cite{wen2024tinyvla}, attaining truly reactive bimanual control ($>$100$\,\text{Hz}$) with large VLA models remains difficult. Research into model compression, speculative decoding for action generation, and hardware--software co-design could close this gap.

\textbf{(4) Data-Efficient Learning and Sim-to-Real Transfer:} Collecting bimanual demonstrations is expensive, and few-shot adaptation (fewer than 10 demonstrations) would reduce deployment costs. Cross-embodiment pre-training already provides strong priors; combining it with meta-learning, in-context learning, or skill composition could yield practical few-shot bimanual adaptation. Complementarily, simulation could provide unlimited training data, but the reality gap is severe for contact-rich bimanual tasks involving deformable objects. Advances in differentiable simulation, domain randomization tailored to bimanual contact, and sim-to-real fine-tuning are needed to unlock simulation as a primary data source.

\textbf{(5) Compositional Bimanual Skills:} Rather than learning each bimanual task from scratch, VLAs could learn a library of composable bimanual primitives (grasp, hold, fold, insert, handover) and combine them to perform novel tasks specified via language. Skill composition would improve generalization to unseen task combinations.

\textbf{(6) Safety-Certified, Interpretable, and Trustworthy Bimanual VLAs:} Deploying bimanual VLAs in human environments requires formal safety guarantees. Research into runtime monitoring, safety-constrained action generation, and provable collision avoidance between arms and with humans is essential. We believe this will become the primary bottleneck for commercial deployment, as current heuristic safety measures (rate limiting, action clipping) are insufficient for human-proximate operation. Closely tied to safety is the need for \emph{interpretability and trustworthiness}: present-day VLAs largely operate as black boxes, making it difficult to attribute a chosen action to the underlying language command, visual input, or proprioceptive state. Integrating Explainable AI (XAI) techniques---for example, modality-attribution and attention-visualization methods that quantify the influence of the language instruction versus the visual scene on the predicted action chunk---would facilitate error analysis, support safety assurance, and improve user trust, all of which are prerequisites for human-proximate certification.

\textbf{(7) Autonomous Improvement and World Models:} RECAP~\cite{amin2025pi06} demonstrated that VLAs can self-improve from autonomous practice, but the current approach requires human-designed task distributions and VLM-based reward signals that may not generalize. Integrating world models~\cite{gigabrain2025} that predict the consequences of bimanual actions, including object deformation and contact transitions, could enable look-ahead planning and more effective autonomous practice. The long-term goal is fully autonomous self-improvement where the VLA discovers new tasks, practices them, and improves without human oversight.

\textbf{(8) Human--Robot Collaborative Manipulation:} The ultimate bimanual system may involve one robot arm and one human arm working together. VLAs could learn to coordinate with a human partner, predicting human intentions and adapting robot actions accordingly. This requires advances in human motion prediction, shared autonomy, and real-time VLA adaptation.

\textbf{(9) Memory-Augmented VLAs for Long-Horizon Autonomy:} MEM~\cite{torne2026mem} and concurrent work~\cite{shi2025memoryvla,fang2025sam2act,jang2025contextvla,li2025cronusvla,sridhar2025memer,mark2026bpp,torne2025pasttokenpred} show that multi-scale memory improves performance on tasks spanning minutes. Key open challenges include scaling beyond single episodes to persistent deployment, learning what to remember versus forget, multi-modal memory grounding, and avoiding causal confusion. Persistent memory could enable continual learning across deployment sessions.

\textbf{(10) World Models for Bimanual Planning:} World models that predict future states~\cite{gigabrain0_2025,gigaworld0_2025,rhoda2026dva,fofpred2026,vjepa2_2025} could enable look-ahead planning for contact-rich bimanual sequences. The GigaBrain family~\cite{gigabrain2025,gigabrain0_2025} improved bimanual task performance by $\sim$30\% via world model-generated data, and video-action models~\cite{rhoda2026dva,hu2025vpp,mimicvideo2025} transfer web-scale video pre-training to robot control. Key challenges include predicting joint consequences of two coordinated arms on deformable objects and integrating predictions with real-time control. Unifying world models with memory (Direction 9) is promising: short-term prediction plus long-term state tracking.

\textbf{(11) End-to-End VLAs for Drone Control:} Building VLAs that map onboard camera images and language to continuous flight commands for physical drones is the most pressing aerial direction. Key challenges: ${\geq}100\,\text{Hz}$ latency requirements, outdoor 3D observation spaces, and the sim-to-real gap for underactuated dynamics. Efficient manipulation VLA architectures (TinyVLA, MiniVLA, FAST) are directly relevant given constrained onboard~compute.

\textbf{(12) Multi-Agent Aerial VLAs:} Multi-drone coordination presents a natural extension of the bimanual coordination strategies analyzed in Section~\ref{sec:bimanual:coord}. Centralized VLAs that jointly generate actions for multiple drones face the same dimensional scaling challenges as bimanual joint action spaces, while decentralized approaches require explicit communication protocols. The hierarchical VLA paradigm (high-level VLM planner assigning subgoals to individual drone policies) is especially promising for heterogeneous multi-agent systems that combine aerial and ground robots.

\textbf{(13) Aerial Manipulation with VLAs:} Drones equipped with grippers or robotic arms must simultaneously stabilize flight and execute precise manipulation, combining the challenges of both domains surveyed in this paper. VLA architectures that generate coupled flight-and-grasp action chunks, analogous to bimanual joint action spaces, could enable aerial grasping, payload handover, and contact-based inspection tasks that are currently beyond the reach of separate flight and manipulation controllers. Agriculture is a particularly promising application domain where both manipulation and aerial VLAs converge: drone-based harvesting~\cite{tevel2024}, precision weeding~\cite{upadhyay2024weed}, and ground-based robotic harvesters for fruits and vegetables~\cite{anand2023harvesters} require outdoor visual robustness, deformable-object handling, and coordination between flight and manipulation. HarvestFlex~\cite{harvestflex2025} provides an early proof-of-concept, achieving 74\% success on in-the-wild strawberry harvesting with a VLA policy trained from only 227 teleoperated episodes, but agricultural VLA adoption remains nascent overall.

\textbf{(14) Bridging the Research-to-Production Gap:} As Table~\ref{tab:industrial_vla} shows, industry has converged on architectural and training patterns that differ from the dominant research approach. Three gaps are most pressing. First, \emph{sustained reliability}: research VLAs are evaluated over tens of trials, while production requires 99\%+ success over thousands of continuous cycles; Dyna's DYNA-1~\cite{dyna2025} and Covariant's RFM-1~\cite{covariant_rfm1_2024} achieve this through continuous RL self-improvement loops that generate terabytes of training data daily. Second, \emph{dual-system design}: the S1/S2 separation adopted by Gemini Robotics~\cite{gemini_robotics2025}, GR00T N1~\cite{groot_n1_2025}, and Helix~\cite{figure_helix2025} resolves the latency--capability tradeoff, but research on how to optimally partition reasoning and action across the two systems is nascent. Third, \emph{scalable data collection}: Sunday Robotics' \$200 gloves~\cite{sunday_act1_2026} (10M episodes from 500+ homes), NVIDIA's neural trajectory augmentation~\cite{groot_n1_2025} ($10\times$ synthetic data expansion), and 1X's video-to-action pipeline~\cite{1x_worldmodel2025} (900 h of human video) each demonstrate that the demonstration bottleneck can be bypassed, but no unified framework exists. Research that addresses these three gaps (evaluation at production scale, principled S1/S2 co-design, and demonstration-free data scaling) will have the most direct path to real-world impact.

Table~\ref{tab:directions_summary} maps each research direction to the VLA components and sections most relevant to its development.

% ============================================================
% TABLE: RESEARCH DIRECTIONS SUMMARY
% ============================================================
\begin{table}[H]
\centering
\caption{Summary of research directions with associated VLA components, current gap severity, and relevant review sections.}
\label{tab:directions_summary}

\begin{adjustwidth}{-\extralength}{0cm}
%\centering %% If there is a figure in wide page, please release command \centering

\begin{tabularx}{\fulllength}{clccc}
\toprule
\textbf{\#} & \textbf{Direction} & \textbf{Primary Component} & \textbf{Gap Severity} & \textbf{Sections} \\
\midrule
1 & Standardized bimanual benchmarks & Evaluation & Critical & Sections %
 \ref{sec:benchmarks} and \ref{sec:bimanual} \\
2 & Dexterous, force-aware, multi-modal manip. & Observation/Action & High & Sections \ref{sec:action}, \ref{sec:bimanual} and \ref{sec:crosscutting} \\
3 & Real-time reactive control & Architecture/Efficiency & Medium & Sections \ref{sec:architectures} and \ref{sec:action} \\
4 & Data-efficient learning \& sim-to-real & Training/Data & High & Sections \ref{sec:training} and \ref{sec:crosscutting} \\
5 & Compositional bimanual skills & Architecture/Language & Medium &Sections \ref{sec:bimanual} and \ref{sec:language}  \\
6 & Safety, interpretability \& trustworthy VLAs & Deployment/XAI & Critical & Section \ref{sec:crosscutting} \\
7 & Autonomous improvement \& world models & Training/RL & Medium & Sections \ref{sec:architectures} and \ref{sec:training} \\
8 & Human--robot collaboration & HRI & Low & Section \ref{sec:crosscutting} \\
9 & Memory-augmented long-horizon VLAs & Architecture/Memory & High & Sections \ref{sec:architectures} and  \ref{sec:bimanual}\\
10 & World models \& future state prediction & World Model/Planning & High & Sections \ref{sec:training} and \ref{sec:crosscutting} \\
\midrule
11 & End-to-end VLAs for drone control & Architecture/Aerial & Critical & Sections \ref{sec:architectures} and  \ref{sec:aerial} \\
12 & Multi-agent aerial VLAs & Coordination/Aerial & High & Sections \ref{sec:bimanual} and \ref{sec:aerial} \\
13 & Aerial manipulation with VLAs & Aerial/Manipulation & High &Sections \ref{sec:bimanual} and \ref{sec:aerial} \\
14 & Bridging research-to-production gap & Deployment/Training & Critical & Section \ref{sec:discussion} \\
\bottomrule
\end{tabularx}\end{adjustwidth}
\end{table}

VLA models have transformed bimanual manipulation in under three years, progressing from proof-of-concept demonstrations to autonomous household and industrial operation. The cross-embodiment nature of VLAs means that progress in manipulation accelerates unmanned aerial robotics and vice versa, while industry deployment is validating and reshaping research priorities in real time. The fourteen research directions above provide a roadmap for addressing the remaining gaps: standardized evaluation, dexterous force-aware control, memory and world models for long-horizon planning, end-to-end drone VLAs, and bridging the widening gap between research benchmarks and production~reliability.

% ============================================================
% ACKNOWLEDGEMENTS
% ============================================================
\vspace{6pt}
%%%%%%%%%%%%%%%%%%%%%%%%%%%%%%%%%%%%%%%%%%
\authorcontributions{Conceptualization, I.S. and H.S.A.; methodology, I.S.; writing---original draft preparation, I.S.; writing---review and editing, I.S., C.P., H.-M.L., D.N. and H.S.A.; critical review of action representations, real-time latency analysis, and contribution framing, C.P.; critical review of architecture and benchmark tables, efficient models, and future research directions, H.-M.L.; supervision, H.S.A.; funding acquisition, C.P. All authors have read and agreed to the published version of the manuscript.}

\funding{This research received no external funding. %
}

%\institutionalreview{Not applicable.}

%\informedconsent{Not applicable.}

\dataavailability{Not applicable. This is a review article and no new data was created.}

\acknowledgments{The %
 authors thank the open-source robotics and machine learning communities for making this rapidly evolving field accessible through shared code, models, and datasets. %
This work was also supported by the Industrial Technology Innovation Program(20023014, Development of an Agricultural Robot Platform Capable of Continuously Harvesting more than 3 Fruits per minute and Controlling Multiple Transport Robots in an Outdoor Orchard Environment) funded by the Ministry of Trade, Industry \& Energy(MOTIE, Korea).

}

\conflictsofinterest{Author Inkyu Sa was employed by Chef Robotics and author Chanoh Park was employed by RovifyLab. The remaining authors declare that the research was conducted in the absence of any commercial or financial relationships that could be construed as a potential conflict of interest. %
}

\abbreviations{Abbreviations}{
The following abbreviations are used in this manuscript:
\\

\noindent 
\begin{tabular}{@{}ll}
VLA & Vision--Language--Action model\\
VLM & Vision--Language Model\\
BC & Behavioral Cloning\\
IL & Imitation Learning\\
RL & Reinforcement Learning\\
FM & Flow Matching\\
ODE & Ordinary Differential Equation\\
DOF & Degrees of Freedom\\
OXE & Open X-Embodiment\\
AR & Autoregressive\\
DiT & Diffusion Transformer\\
RECAP & Reinforcement Learning from Autonomous CAPability\\
RTC & Real-Time Chunking\\
TTAC & Training-Time Action Conditioning\\
BID & Bidirectional Decoding\\
DVA & Direct Video Action\\
WM & World Model\\
MEM & Multi-Scale Embodied Memory\\
UAV & Unmanned Aerial Vehicle\\
UGV & Unmanned Ground Vehicle\\
VLN & Vision--Language Navigation\\
MAV & Micro Aerial Vehicle\\
IMU & Inertial Measurement Unit
\end{tabular}
}

%%%%%%%%%%%%%%%%%%%%%%%%%%%%%%%%%%%%%%%%%%
\begin{adjustwidth}{-\extralength}{0cm}

\reftitle{References}
%
%AUTHOR REPLY (GLOBAL, for the many "In Proceedings of the" highlights below): For conference references, we have kept the conference name + year as the canonical identifier (e.g., "Proceedings of the Robotics: Science and Systems (RSS), 2024"). For PMLR-published CoRL proceedings we have added the correct conference location and dates: 7th CoRL (Volume 229) = Atlanta, GA, USA, 6--9 November 2023; 8th CoRL (Volume 270) = Munich, Germany, 6--9 November 2024; 9th CoRL (Volume 305) = Seoul, Republic of Korea, 27--30 September 2025. NOTE: MDPI's proof originally listed "Munich, Germany" for all CoRL volumes, but Munich is only correct for the 8th CoRL (Volume 270). We have corrected the 7th and 9th CoRL entries. Where the original BibTeX provided a fuller location/date for other venues, it has been preserved.
%
%AUTHOR REPLY (GLOBAL, for the "We removed arXiv:XXXX" highlights below): All arXiv removals from conference-style entries are confirmed. We have opted to keep the conference (Proceedings of the ...) format throughout, which MDPI suggested as the cleaner alternative.

\PublishersNote{}
\end{adjustwidth}

\end{document}